\documentclass[10pt,journal,compsoc]{IEEEtran}
\ifCLASSOPTIONcompsoc
  \usepackage[nocompress]{cite}
\else
  \usepackage{cite}
\fi
\ifCLASSINFOpdf
\else
\fi

\usepackage{xspace,url}
\usepackage{float}
\usepackage{graphicx}
\graphicspath{{images/}}

\usepackage{amsmath,amssymb}
\usepackage{subcaption}
\usepackage{url}

\usepackage{amssymb,amsmath,graphics,graphicx,latexsym,amssymb,url}
\usepackage{balance}

\usepackage{times}
\usepackage{caption}

\usepackage{breakurl}
\usepackage[breaklinks]{hyperref}

\usepackage{lipsum,algorithm,multicol}

\usepackage{amssymb}
\usepackage{graphicx,float}
\usepackage{algpseudocode}
\usepackage{mdframed}
\usepackage{xcolor}

\newcommand{\revise}[1]{\textcolor{black}{#1}}
\newcommand{\newrevise}[1]{\textcolor{blue}{#1}}

\algnewcommand{\LineComment}[1]{\State\(\triangleright\) #1} 
\let\oldReturn\Return
\renewcommand{\Return}{\State\oldReturn}

\newmdenv[innerlinewidth=0.5pt, roundcorner=4pt,linecolor=gray,innerleftmargin=8pt,
innerrightmargin=8pt,innertopmargin=8pt,innerbottommargin=8pt]{note}

\usepackage{fp}
\usepackage{color}
\usepackage{colortbl}
\usepackage{amsmath}
\usepackage{algorithm}
\usepackage{algpseudocode}
\usepackage{tikz} 
\usetikzlibrary{shapes,arrows}
\usepackage{pgfplots}
\pgfplotsset{compat=1.14}
\usepackage{float}
\usepackage{booktabs}
\usepackage{multirow}
\usepackage{graphicx}
\usepackage{makecell}
\usepackage{tcolorbox,xspace,amssymb,amsmath,amsthm}
\usepackage{balance}
\usepackage{bm}
\usepackage{cancel}
\usepackage{listings}

\definecolor{LightGreen}{rgb}{0.7, 1.0, 0.7}
\definecolor{LightRed}{rgb}{1.0, 0.7, 0.7}
\definecolor{LightBlue}{rgb}{0.7, 0.7, 1.0}
\definecolor{TableHeader}{rgb}{0.9, 0.9, 0.9}

\algdef{SE}[VARIABLES]{Variables}{EndVariables}
   {\algorithmicvariables}
   {\algorithmicend\ \algorithmicvariables}
\algnewcommand{\algorithmicvariables}{\textbf{global variables}}

\usepackage{pifont}

\newcommand{\FT}{\textsc{Ogma}\xspace}

\newtheorem{definition}{Definition}

\begin{document}
\renewcommand{\newrevise}[1]{\textcolor{black}{#1}}
%
\title{Grammar Based Directed Testing of Machine Learning Systems
}



\author{Sakshi Udeshi,~\IEEEmembership{Member,~IEEE,}
        Sudipta~Chattopadhyay,~\IEEEmembership{Member,~IEEE,}
\IEEEcompsocitemizethanks{\IEEEcompsocthanksitem S. Udeshi and 
S. Chattopadhyay are with Singapore University of Technology and Design.}
}

\clubpenalty=10000
\widowpenalty=10000

\IEEEtitleabstractindextext{%


\begin{abstract}
The massive progress of machine learning has seen its application over a variety 
of 
 domains in the past decade. But how do we develop a systematic, 
scalable and modular strategy to validate machine-learning systems? We present, 
to the best of our knowledge, the first approach, which provides a systematic test 
framework for machine-learning systems that accepts grammar-based inputs. Our 
\FT approach automatically discovers erroneous behaviours in classifiers and leverages 
these erroneous behaviours to improve the respective models. 
\FT leverages inherent robustness properties present in any well trained 
machine-learning model to direct test generation and thus, implementing a scalable 
test generation methodology. 
To evaluate our \FT approach, we have tested it on three real world natural language 
processing (NLP) classifiers. We have found thousands of erroneous behaviours in these 
systems. We also compare \FT with a random test generation approach and observe that 
\FT is more effective than such random test generation by up to 489\%. 
%
\end{abstract}

}

\maketitle

\IEEEdisplaynontitleabstractindextext

%
\IEEEpeerreviewmaketitle

\IEEEraisesectionheading{\section{Introduction}
\label{sec:intro}}

In recent years, the application of machine-learning models has escalated 
to several application domains, including sensitive and safety-critical 
application domains such as the automotive industry~\cite{Automotive-AI}, human 
resources\cite{HR-AI} and education\cite{education-AI}. 
One of the key insight behind the usage of such models is to automate 
mundane and typically error-prone tasks of decision making. On the flip side, 
these machine-learning models are susceptible to erroneous behaviour, which 
may induce unpredictable scenarios, even costing human lives and causing 
financial damage. As an example, consider the following sentence that might 
be processed by an automated emergency response service: 
\begin{verse}
{\em``My house is on fire. Please send help in Sebastopol, CA. There is a huge 
forest fire approaching the town."}
\end{verse}
While processing this text using a well trained text classifier model~\cite{rosette-url}, 
it provides the following classification classes for the text: 
\begin{verbatim}
'Hobbies and Interests', 'Science', 
'Arts and Entertainment', 'Home and Garden', 
'Religion and Spirituality'
\end{verbatim}
It is needless to mention that the respective text classifier is unsuitable  
for categorising the emergency aspect underneath the text and therefore, is 
broken for the usage in emergency text classification. In short, systematic 
validation of machine-learning models is of critical importance before 
deploying them in any sensitive application domain.

\begin{figure}[t]
\begin{center}
\includegraphics[scale=0.8]{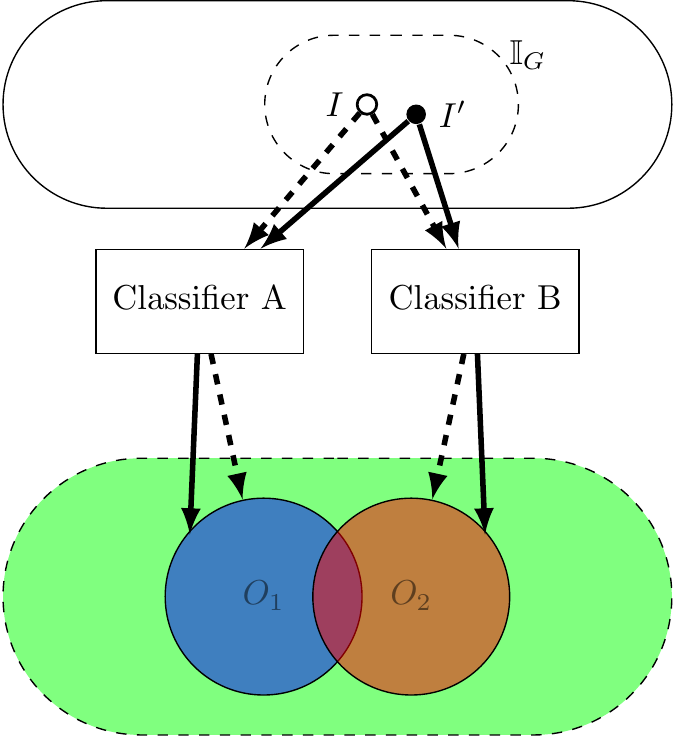}
\end{center}
\caption{Erroneous behaviour of Classifier A and/or Classifier B}
\label{fig:overview-approach}
\end{figure}

In this paper, we broadly consider the problem of systematically testing the 
erroneous behaviours of arbitrary \revise{machine learning based natural language 
processing models}. Moreover, we consider 
these models are amenable only to text inputs conforming to certain 
grammars -- a common feature across a variety of systems including models used  
in text classification. While the nature of erroneous behaviours in a machine-learning 
model depends on its input features, it is often challenging to formally characterise 
such behaviours. This is due to the inherent complexity of real world machine-learning 
models.  
%
%
%
To deal with such complexity, we leverage differential testing. Thus, instead 
of checking whether the output of a classifier is correct for a given input, 
we compare the output with the respective output of a different classifier 
realising the same problem. If the outputs from two classifiers are vastly 
dissimilar, then we call the respective input to be {\em erroneous}. The 
primary objective of this paper is to facilitate discovery of erroneous 
inputs. Specifically, {\em given a pair of machine-learning models and a grammar 
encoding their inputs, our \FT \footnote{God of language from Irish mythology and 
Scottish mythology} approach systematically searches the input space 
of the models and discovers inputs that highlight the erroneous behaviours.}


As an example, consider the behaviours of $\mathit{Classifier}\ A$ and 
$\mathit{Classifier}\ B$, which are targeted for the same classification 
job over an input domain conforming to grammar $G$ (i.e. $\mathbb{I}_G$), in \autoref{fig:overview-approach}. 
Despite being targeted for the 
same classification task, $\mathit{Classifier}\ A$ and 
$\mathit{Classifier}\ B$ generate largely dissimilar classification 
classes $O_1$ and $O_2$ for the same input $I$. The dissimilarity in outputs 
is indicative of one or both of the outputs being incorrect. Such erroneous behaviours 
in the classifiers might appear due to the outdated or inappropriate 
training data. We use our \FT approach to automatically discover 
erroneous inputs such as $I$. Moreover, we can use these inputs to 
retrain and reduce the erroneous behaviours of the classifiers. 



The directed strategy embodied within \FT forms the crux of its scalability 
and effectiveness. Concretely, \FT leverages the robustness property of 
common machine-learning models. According to the robustness 
property~\cite{robustness_classifier}, the classification classes of two 
similar inputs do not vary substantially for well trained machine-learning 
models. As an example, consider the input $I'$ in \autoref{fig:overview-approach}, 
to be similar to input $I$. The classification classes for input $I'$ will 
be similar to $O_1$ and $O_2$ for $\mathit{Classifier}\ A$ and 
$\mathit{Classifier}\ B$, respectively. In other words, if input $I$ is 
an erroneous input, then input $I'$ is likely to be erroneous too. To 
realise this {\em robustness property} for test generation, \FT designs 
a perturbation function to continuously derive similar inputs to $I$ and 
$I'$ and thus, exploring the neighbourhood of erroneous inputs for a given 
classifier. Such a perturbation cannot simply be obtained by mutating a 
raw input, as the mutated input may not conform to the grammar. To this end, 
\FT perturbs the derivation tree to explore the erroneous input subspace.


The grammar-based test input generation and the directed strategy make our 
\FT approach generic in terms of testing arbitrary erroneous behaviours 
of \revise{machine learning based natural language 
processing classifiers}. In contrast to existing works that use   
concrete inputs from the training dataset to test machine-learning 
models~\cite{wicker2017feature,DBLP:conf/sosp/PeiCYJ17}, our \FT approach 
does not require training data for testing the models. Instead, we abstract 
the input space of the model via a grammar, which 
is a common strategy to encode an arbitrarily-large space of structured inputs. 
Thus, the tests generated by \FT can explore these large input space, 
potentially discovering more errors when compared to limiting the test 
generation via the training data.    
%
%
In contrast to previous works, \FT is not limited to test specific 
applications~\cite{DBLP:journals/corr/abs-1801-05950} or 
properties~\cite{DBLP:conf/sigsoft/GalhotraBM17,aequitas}.  
\FT works completely blackbox and can be easily adapted to test real-world 
classification systems for a variety of different applications. Finally, 
we show that the erroneous inputs generated by \FT are useful and 
can be used for retraining the model under test and reducing erroneous behaviours. 


The remainder of the paper is organised as follows. After providing the 
relevant background and an overview of \FT approach in \autoref{sec:overview}, 
we make the following contributions: 

\begin{enumerate}
\item We present \FT, a novel approach for systematically testing erroneous 
behaviours of arbitrary \revise{machine learning based natural language 
processing models}. The \FT approach is based 
on a directed strategy to discover and explore the erroneous input subspace. 
Since the directed strategy embodied in \FT is based on the fundamental 
robustness property of well-trained machine-learning models, we believe that 
\FT can be adopted for testing arbitrary machine-learning models exhibiting 
robustness (\autoref{sec:method}).

\item We provide an implementation of \FT in python. Our implementation and 
all experimental data are publicly available (\autoref{sec:results}). 

\item We evaluate \FT on three real-world text classifier service providers, 
namely Rosette~\cite{rosette-url}, uClassify~\cite{uclassify-url} and 
Aylien~\cite{aylien-url}. We show that our \FT approach discovers up to 
90\% error inducing inputs (with respect to the total number of inputs 
generated) across a variety of grammars. We also show that the directed 
strategy in \FT substantially outperforms (up to 489\%) a strategy that 
randomly generates inputs conforming to the given grammar (\autoref{sec:results}). 

\item We design and evaluate an experiment to show how the error inducing 
inputs generated by \FT can be  utilised to repair the test classifiers. 
We show that by retraining the test classifiers with the generated error 
inducing inputs, the erroneous behaviour can be reduced as much as 24\%.  

\end{enumerate}

After discussing the related work (\autoref{sec:relatedWork}) and threats 
to validity (\autoref{sec:threatsToValidity}), we conclude and reflect in 
\autoref{sec:conclusion}.






\section{Background}
\label{sec:background}

In this section, we introduce the relevant background and the key 
concepts based on which we design our \FT approach. 

\paragraph*{\textbf{Systems based on machine learning}}
In this paper, we are concerned about a \revise{machine learning based natural 
language processing model} that accepts  
an input $I$ and classifies it into one of the $n$ classes from the set 
$\{C_1, C_2, C_3, \cdots, C_n\}$. Moreover, such an input $I$ conforms to 
a grammar $G$, which encodes the set of all valid inputs for the model. 
Some classic examples of such models include deep-learning-based systems 
to categorise news items and systems that analyse the sentiments from Twitter
feeds, among others. As of today, most machine-learning models are tested 
on their accuracy for well-defined sets of data. Such a strategy only 
validates a model on the available datasets. However, it lacks capability 
to systematically and automatically explore the input space accepted by 
the model and not captured by the available datasets. This is crucial, 
as inputs not captured by the available datasets may be presented to the 
model in a production setting and lead to catastrophic error, 
potentially costing human lives~\cite{tesla-blog},~\cite{tesla-crash-url}. 
%
%
%
In summary, a systematic validation of machine-learning model demands the 
machinery of automated software testing, a field that is largely unexplored 
in the light of testing machine-learning models. 


\paragraph*{\textbf{Challenges in validating machine-learning-based systems}} 
There exist multitudes of challenges in systematically validating 
machine-learning models. Consider an arbitrary machine-learning model $M$ that 
accepts input $I$ conforming to grammar $G$ and classifies $I$ in one of the 
category $\{C_1, C_2, C_3, \cdots, C_n\}$. Firstly, without precisely knowing 
the {\em correct} categorisation of input $I$, it is not possible to validate 
the model $M$. In other words, validation of machine-learning models faces the 
{\em oracle problem}~\cite{oracle-paper} in software testing. Secondly, there 
has been significant effort in the software engineering research community 
to design directed test input generation strategies. The insight behind such 
directed strategies is to uncover bugs faster. For instance, to check the 
presence of crashes in {\tt C} programs, a directed strategy may steer the 
test execution towards statements accessing pointers. Such directed 
strategies are well studied for deterministic software and their correctness 
properties. However, systematically steering the execution of a machine-learning 
model, in order to make its prediction dramatically wrong, is still immature. 
Finally, the error inducing inputs for a machine-learning model may not 
necessarily highlight a bug in the respective code (unlike classic software 
debugging process). Instead error inputs may highlight flaws in the data on 
which the respective machine-learning algorithm was trained to obtain the model 
under test. Therefore, the systematic usage of the error inducing inputs, 
to debug the machine-learning model, is also of critical importance. 

%

\paragraph*{\textbf{Differential testing}}
To solve the oracle problem in testing machine-learning models, we leverage 
differential testing. Specifically, consider two models $M_1$ and $M_2$ 
that expect valid inputs conforming to the same grammar $G$ and classifies 
each input from the same set of categories $\{C_1, C_2, C_3, \cdots, C_n\}$. 
For an input $I$ conforming to $G$, if the prediction of $M_1$ and $M_2$ 
are drastically different, then we conclude that $I$ is an error inducing 
input for at least one of $M_1$ and $M_2$. In \autoref{sec:method}, we 
formally define the criteria for identifying such an error inducing input. 
Although our testing strategy requires two models from the same problem 
domain, we believe this is practical, given the presence of a large class 
of machine-learning models targeting real-world problems. Moreover, 
our proposed strategy can also be useful to discover regression bugs via 
comparing the outputs from two different versions (e.g. a stable 
version and a developing version) of the same machine-learning model. 
\newrevise{It is worthwhile to note that differential testing has also been 
successfully used for testing ML models in other domains
\cite{DBLP:conf/sosp/PeiCYJ17} (e.g. computer vision).}

\paragraph*{\textbf{Robustness in machine learning}}
The insight behind the directed testing in \FT is based on the {\em robustness} 
of common machine-learning models. Conceptually, robustness in machine-learning 
captures a phenomenon stating that a slight change in the input does not change 
the output dramatically in well-trained machine-learning 
models~\cite{robustness_classifier}. This means that error inducing inputs are likely 
to be clustered together in the input space of well-trained models. 
Technically, assume a model $f$, and let $I$ be an input to $f$ and $\delta$ be a 
small value. If $f$ is robust, then $f(I) \approx f(I \boxtimes \delta)$, 
where $I \boxtimes \delta$ captures an input obtained via small $\delta$ 
perturbation of input $I$. In such case, we say that input $I \boxtimes \delta$ 
is in the {\em neighbourhood} of input $I$. Since $f(I) \approx f(I \boxtimes \delta)$, 
we hypothesise that if an input $I$ causes an error, then it is likely that input 
$I \boxtimes \delta$ will cause an error too.
This hypothesis forms the crux of our directed testing methodology.

\paragraph*{\textbf{State-of-the-art in testing machine-learning-based systems}}
Adversarial testing~\cite{DBLP:conf/sp/Carlini017,textbugger} 
techniques have the objective 
to fool a machine-learning model with minute perturbation on inputs and guiding 
the model towards a dramatically wrong prediction. However, such testing strategies 
are only limited to minimal and unobservable input perturbations and require a 
set of seed inputs.  Therefore, adversarial techniques are neither sufficient nor 
general enough to check the erroneous behaviour of machine-learning models. Besides, 
adversarial testing does not solve the test design problem in the broadest sense due to 
their dependency on a set of seed inputs and due to their incapability to discover
faults that may only appear with observable differences across inputs. 
Finally, if the processed data by the machine-learning model requires security clearance 
(e.g. healthcare data, finance data), then we need a 
systematic process to generate these inputs during the automated validation 
stage of the model.


In recent years, the software engineering research community have stepped up to 
develop testing methodologies for deep-learning systems~\cite{DBLP:journals/corr/abs-1801-05950,DBLP:conf/sosp/PeiCYJ17,tian2017deeptest,DBLP:conf/sigsoft/GalhotraBM17}. These 
works, however are, limited either to specific applications
~\cite{DBLP:journals/corr/abs-1801-05950} 
or rely on the presence of sample inputs~\cite{wicker2017feature,DBLP:conf/sosp/PeiCYJ17}. 
Moreover, none of the prior works are applicable to generate grammar-based inputs in 
a fashion that such inputs steer the execution of machine-learning models to 
erroneous behaviour. In the subsequent sections, we will discuss the key ingredients 
of our \FT approach that accomplishes this objective.

\begin{figure*}[h]
\begin{center}
\begin{tabular}{ccc}
\includegraphics[scale=1.2]{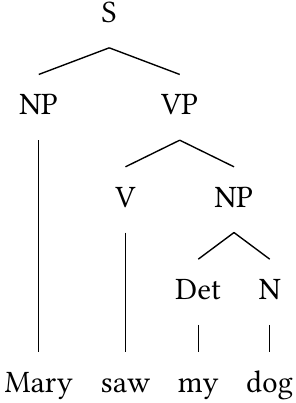} & 
\includegraphics[scale=1.2]{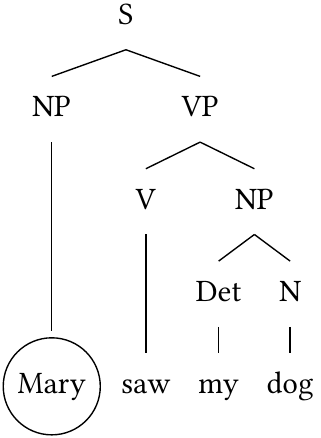} & 
\includegraphics[scale=1.2]{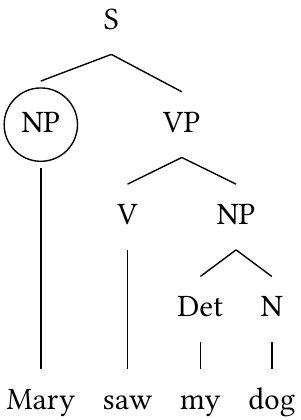}\\
{\bf (a)} & {\bf (b)} & {\bf (c)}\\
\\
\includegraphics[scale=1.2]{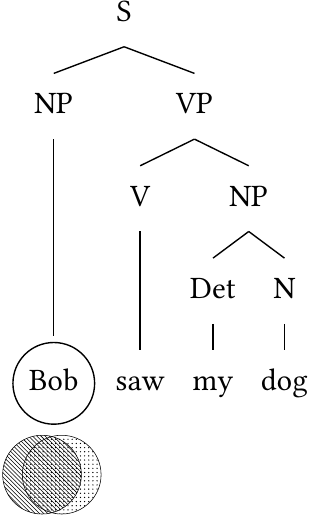} & 
\includegraphics[scale=1.2]{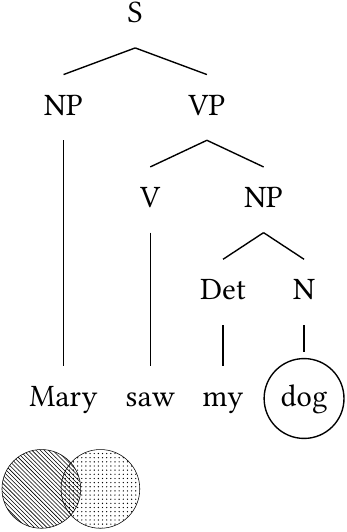} & 
\includegraphics[scale=1.2]{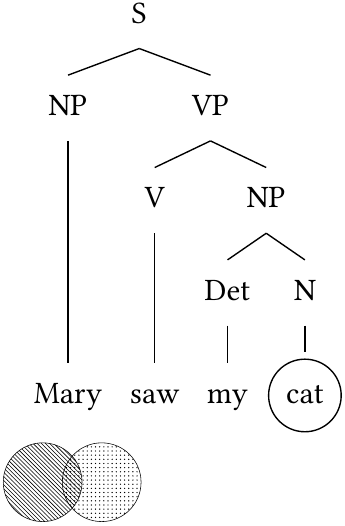}\\
{\bf (d)} & {\bf (e)} & {\bf (f)} \\
\end{tabular}
\end{center}
\caption{Different stages of \FT: (a) Derivation tree of the sentence ``{\em Mary saw my dog}", 
(b) to perturb the sentence slightly, \FT first chooses a terminal symbol at random, in this 
case, the word ``{\em Mary}", (c) \FT discovers the production rule generating the word 
``{\em Mary}", in this case, the rule: {\em NP $\rightarrow$ "John" $|$ "Mary" $|$ "Bob" $|$ Det N $|$ Det N PP}, (d) \FT 
perturbs the initial sentence with a different terminal symbol as per the production rules 
of the non-terminal $NP$, in this case, \FT chooses to replace ``{\em Mary}" with ``{\em Bob}" 
and gets a new sentence ``{\em Bob saw my dog}". This new sentence, however, does not lead to 
any error as the set of classification classes from two models (as shown via the two circles) 
are largely similar. (e) \FT backtracks and randomly chooses another terminal 
symbol from  ``{\em Mary saw my dog}" to perturb, in this case, the terminal ``{\em dog}", 
(f) \FT generates a perturbed sentence ``{\em Mary saw my cat}" that also leads to an error 
(i.e. little overlap between the set of classification classes from two models).}
\label{fig:steps-overview}
\end{figure*}

\section{Overview of \FT}
\label{sec:overview}

%
%
%
%


In this section, we will outline the working principle of \FT via simple 
examples. 


\begin{figure}[h]
\begin{center}
\includegraphics[scale=0.8]{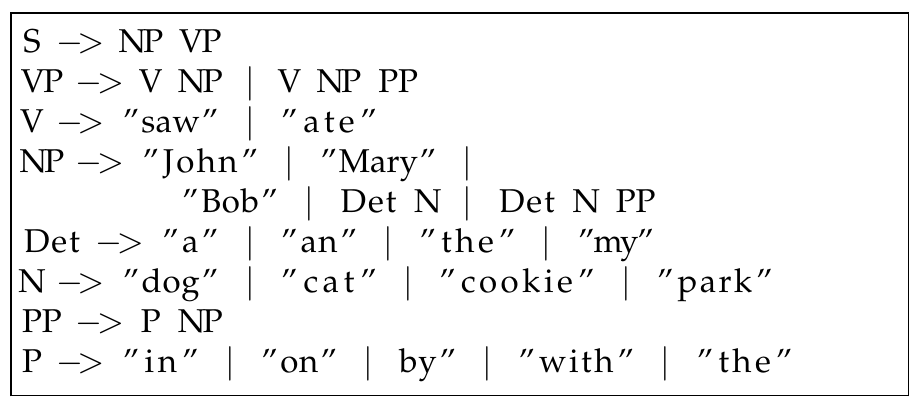}
\end{center}
\caption{Example Grammar}
\label{fig:toy-gram-1}
\end{figure}

Consider a context free grammar as shown in \autoref{fig:toy-gram-1}. We assume that the 
sentences generated from this grammar are used as inputs to machine-learning-based 
systems, such as classifiers. These classifiers may be used to identify a 
sentence into a specific category, such as {\em hobby}, {\em sports} and 
so on. 

\paragraph*{\textbf{Differential Testing}}
Our \FT approach starts with an initial input $I$. Such an initial input is 
randomly generated from the grammar. Let us assume that the initial input is 
``{\em Mary saw my dog}". To check whether this input leads to any 
classification error, we feed it into two text classifiers. Usually the 
real-world text classifiers, as used in our evaluation, return a set of 
classification classes. Let us assume $\mathbb{C}_1$ and $\mathbb{C}_2$ 
are the set of classification classes returned by the two classifiers 
$M_1$ and $M_2$, respectively. To check whether the initial input lead 
to a classification error, we evaluate the Jaccard Index 
$\frac{\left | \mathbb{C}_1  \cap \mathbb{C}_2 \right |}{\left | \mathbb{C}_1 \cup \mathbb{C}_2 \right |}$. 
If the Jaccard Index between $\mathbb{C}_1$ and $\mathbb{C}_2$ is below 
a user-defined threshold $J$, then we conclude that either $M_1$ or $M_2$ 
exhibits a classification error. We note that the threshold $J$ controls 
the error condition. For instance, a very low value for $J$ will enforce 
a strong condition on identifying a classification error.

\paragraph*{\textbf{Directed Testing}}
One of the key challenge for testing machine-learning models is to 
systematically direct the test generation process. This is to discover 
erroneous behaviours as fast as possible. While directed test generation 
is well studied for deterministic software systems, a similar development 
is limited in the case of machine-learning systems. In this paper, we 
leverage the robustness of well trained machine learning models to design 
a directed test generation method. According to robustness, similar inputs 
and similar outputs are clustered together for such machine-learning 
models~\cite{robustness_classifier}. Thus, we hypothesise that error 
inducing inputs are also clustered together. However, as our \FT approach 
targets inputs conforming to certain grammars, it is not straight-forward 
to define the neighbourhood of an input that are likely to be classified 
in a similar fashion. Specifically, we need to explore the following for 
a directed test generation:
\begin{enumerate}
\item {\em The grammar under test:} We should be able to generate a substantial 
number of inputs that are derived similarly (e.g. by applying similar sequence 
of production rules) from the grammar. This is to facilitate exploring the 
neighbourhood of an input conforming to the grammar and thus, exploiting the 
robustness property of the machine-learning model under test. As observed 
from the grammar introduced earlier in this section, it does encode several 
inputs that are derived via similar sequence of production rules. 

\item {\em The distance between inputs conforming to a given grammar:} We 
need an artifact to formally define and explore the neighbourhood of an 
arbitrary input conforming to a grammar. To this end, we chose the 
{\em derivation tree} of an input generated from the grammar. We consider 
two different inputs $I$ and $I'$ (both conforming to the grammar) in the 
same neighbourhood if $I$ and $I'$ have the same derivation tree, but the 
exception of a terminal symbol. For instance, the input sentences 
{\em Mary saw my dog} and {\em Bob saw my dog} are in the same neighbourhood, 
as they differ only in the production rules 
$\mathit{NP} \rightarrow \mathit{Mary}$ and  $\mathit{NP} \rightarrow \mathit{Bob}$ 
(see \autoref{fig:steps-overview}(a) and \autoref{fig:steps-overview}(d)), 
respectively. 

\end{enumerate}

\paragraph*{\textbf{\FT in Action}}
\autoref{fig:steps-overview} captures an excerpt of \FT actions when initiated 
with an input sentence $I \equiv$ {\em Mary saw my dog}. For the sake of illustration, 
let us assume that the initial sentence led to a classification error. Thus, 
\FT aims to explore the neighbourhood of input $I$ and targets to discover more 
error inducing inputs. \autoref{fig:steps-overview}(a) captures the derivation 
tree of the input $I$. We wish to find an input $I'$ that has the same derivation 
tree as $I$ except for one lead node. To this end, we randomly chose a terminal 
symbol appearing in $I$. As shown in \autoref{fig:steps-overview}(b), \FT randomly 
chooses the terminal symbol {\em Mary}. Subsequently, we discover the production 
rule generating the randomly chosen terminal symbol. In \autoref{fig:steps-overview}, 
\FT identifies this production rule to be $\mathit{NP} \rightarrow \mathit{Mary}$. 
Finally, \FT generates $I'$ by randomly choosing a production rule other than 
$\mathit{NP} \rightarrow \mathit{Mary}$. As observed in \autoref{fig:steps-overview}(d), 
\FT identifies the production rule $\mathit{NP} \rightarrow \mathit{Bob}$, leading 
to the new test input $I' \equiv$ {\em Bob saw my dog}.

The test input $I' \equiv$ {\em Bob saw my dog} might not lead to a classification 
error, as reflected in \autoref{fig:steps-overview}(d). Intuitively, this can be 
viewed as \FT moving outside the neighbourhood of error inducing inputs with 
$I' \equiv$ {\em Bob saw my dog}. Thus, \FT stops performing any more modifications 
to input $I'$ and backtracks. To realise this backtracking, \FT sets the input $I'$ 
to the original input {\em Mary saw my dog}. \FT, then chooses another terminal
symbol randomly, as observed in \autoref{fig:steps-overview}(e). A terminal
symbol {\em``dog"} (cf.  \autoref{fig:steps-overview}(e)) is chosen. Subsequently, \FT 
finds the production rule resulting the terminal {\em``dog"} in a similar manner. 
Once \FT finds this rule, a random terminal other than {\em``dog"} is chosen from this rule. This new terminal 
symbol will replace {\em``dog"} in $I'$. As seen in \autoref{fig:steps-overview}(f), 
$I' \equiv $ {\em Mary saw my cat}. Now, input $I'$ leads to a classification error, as 
indicated by \checkmark. Thus, \FT follows the same steps, as explained in the preceding 
paragraphs, to perturb $I'$ and generate more error inducing inputs.



In the case where the initial input $I \equiv$ {\em Mary saw my dog} was 
{\em not} error inducing, we continue to perturb the input until an error 
inducing input is discovered. In our experiments, we 
observed that such a strategy discovers an error inducing input quickly even 
though the initial input is not error inducing. Once an error inducing input 
is discovered, the test effectiveness of \FT accelerates due to the presence 
of more error inducing inputs in the neighbourhood. 
Thus, the initial input 
could be randomly generated and it has negligible impact on the effectiveness 
of \FT. 

\paragraph*{\textbf{Choice of grammar-based equivalence}}
\revise{\FT abstracts the input space via a grammar and explores the input space 
with the objective of generating erroneous inputs. While generating the 
inputs, \FT does not aim to preserve the semantic similarity, instead it 
guarantees that all generated inputs conform to the grammar and thus valid. 
We choose this approach for multiple reasons. Firstly, \FT aims to explore 
a larger input space abstracted by the grammar, instead of restricting the 
exploration to the semantically equivalent inputs. Thus, as long 
as the semantically equivalent inputs conform to the grammar, they can 
potentially be explored by \FT. To this end, our approach is unaffected 
even if there exists antonym in the production rule. 
Secondly, if the semantic similarity is the key factor affecting the 
classifier results and a perturbation involves an antonym, then \FT 
backtracks to the previous (erroneous) sentence. Then, it chooses 
another terminal symbol for perturbation that may preserve the semantic 
similarity. As shown by our empirical evaluation, that the backtracking 
is crucial in the design of \FT and it leads to $\approx 85\%$ erroneous 
inputs.  
%
%
Finally, although we evaluated \FT for natural language 
processing tools, the central idea behind \FT is applicable to any 
machine-learning model whose valid inputs can be captured by a grammar. 
Such ML models may span across 
a wide range of applications including detection of malicious http and 
javascript traffic~\cite{malware-http} and malicious powershell 
commands~\cite{malware-powershell}.
However, the notion of semantic equivalence varies across various application domains. 
For example, the notion of program semantic equivalence (e.g. for an ML-based 
malware detector) is completely different as compared to the notion 
of semantic equivalence in natural language (e.g. for an ML-based 
natural language processing tool). Thus, to keep the idea behind \FT 
applicable to a variety of ML application domains, we focus on grammar-based 
equivalence instead of semantic equivalence.}

\paragraph*{\textbf{Handling non-robust input subspace}}
\revise{
It is well known in existing~\cite{textbugger}
literature that there are certain inputs 
that violate the robustness property of ML systems. 
However, such adversarial inputs generally cover only a small fraction 
of the entire input space. This is evident by the fact that adversarial 
inputs need to be crafted using very specialized techniques. Additionally, 
\FT is designed to avoid these adversarial input regions by systematically 
directing the test generator (e.g. via backtracking). Intuitively, \FT  
achieves this by backtracking from a non-error-inducing input subspace 
(see Algorithm~1 for details). Consequently, if adversarial or non-robust 
input regions do not exhibit erroneous inputs, such regions will eventually
be explored less frequently by \FT.}

\section{Detailed Methodology}
\label{sec:method}
In this section we discuss our \FT approach in detail. Our approach revolves 
around discovering erroneous behaviours by systematically \em{perturbing} the
\em{derivation tree} of an input that conforms to a grammar $G$. First, we 
introduce the notion of Jaccard index, erroneous inputs, tree similarities 
and input perturbation before delving into the algorithmic details of our 
approach. We capture the notations used henceforth in \autoref{table:notation}.

\begin{table}[t]
	\centering
	\caption{Notations used in \FT approach}
	\vspace*{-0.1in}
	\label{table:notation}
	\begin{tabular}{| c | p{6cm} | }
	\hline
	$G$ & A grammar used to generate test inputs \\ \hline
	
	$\mathbb{I}_{G}$ & All inputs described by a grammar $G$  \\ \hline
	
	$\mathbb{T}_{G}$ & The derivation trees of any input 
	$I \in \mathbb{I}_{G}$ \\ \hline
	
	$f_1, f_2$ & Classifiers under test. \\ \hline
	
	$J$ & A pre-determined Jaccard Threshold \\ \hline
	
	$\tau_G$ & A function $\mathbb{I}_{G} \rightarrow \mathbb{T}_{G}$ which 
	outputs the derivation tree of an input $I \in 
	\mathbb{I}_{G}$ \\ \hline
	
	$S$ & The initial input to the directed search. $S$ conforms to 
	grammar $G$ \\ \hline
\end{tabular}
	\vspace*{-0.15in}
\end{table}

\theoremstyle{definition}
\begin{definition}{\textbf{(Jaccard Index)}}
\label{def:jaccard-index}
{
For any two sets $A$ and $B$ the Jaccard Index $JI$ is defined as follows~\cite{jaccard-index}

\[
JI(A, B) = \frac{|A \cap B|}{|A \cup B|} 
\]

\[
0 \leq JI(A,B) \leq 1
\]
If A and B are both empty, we define $JI(A,B) = 1$.
}
\end{definition}
Within our \FT approach, $JI$ is used to compare the output classification 
classes of two test classifiers. 
%
It is worthwhile to mention that we choose the Jaccard Index due to the choice 
of subject classifiers in our empirical evaluation. The choice of such a 
metric is modular and can be fine tuned. This means that \FT is extensible for 
not only other set similarity metrics, but also for regressors. 

\revise {In our experiments, the output of the classifiers are finite sets.
The Jaccard index satisfies the properties of metric
\cite{jaccard-distance-between-sets}. Other common set similarity indices such as 
S\o{}rensen-Dice coefficient and the Tversky index are related to the Jaccard Index 
and may not be metric \cite{dice-not-metric} \cite{tversky-not-metric}. As a 
result, for finite sets, the Jaccard Index is our preferred set similarity index.}

\theoremstyle{definition}
\begin{definition}{\textbf{(Erroneous input)}}
\label{def:error}
{
We say that input $I \in \mathbb{I}_{G}$ is an erroneous input if the output 
sets of the classifiers $f_1, f_2$ satisfy the following condition 
\[
JI(f_1(I), f_2(I)) < J
\]

}
\end{definition}
%
The threshold $J$ is a user-defined threshold. A lower value of $J$ indicates 
a stricter condition for finding erroneous inputs. 

\theoremstyle{definition}
\begin{definition}{\textbf{(Tree Similarity)}}
\label{def:similar-trees}
{
We say two trees $T_1$ and $T_2$ are similar if we can construct a tree 
$T$ by replacing exactly one leaf node in $T_1$ (respectively, $T_2$) such 
that $T$ is identical to $T_2$ (respectively, $T_1$). 
}
\end{definition}

\theoremstyle{definition}
\begin{definition}{\textbf{(Input perturbation)}}
\label{def:perturbation}
{
Let $\tau_G: \mathbb{I}_{G} \rightarrow \mathbb{T}_{G}$ be a function such that 
for an arbitrary input $I \in \mathbb{I}_{G}$, $\tau_G(I)$ is the derivation tree 
for $I$. We define $\mathit{Perturb}$ as a function 
$\mathit{Perturb}: \mathbb{I}_{G} \rightarrow \mathbb{I}_{G}$ such that for an 
input $I \in \mathbb{I}_{G}$, if $I' = \mathit{Perturb}(I)$, then $\tau_G(I)$ 
and $\tau_G(I')$ are similar trees.
}
\end{definition}

\begin{figure}[h]
\begin{center}
\includegraphics[scale=0.6]{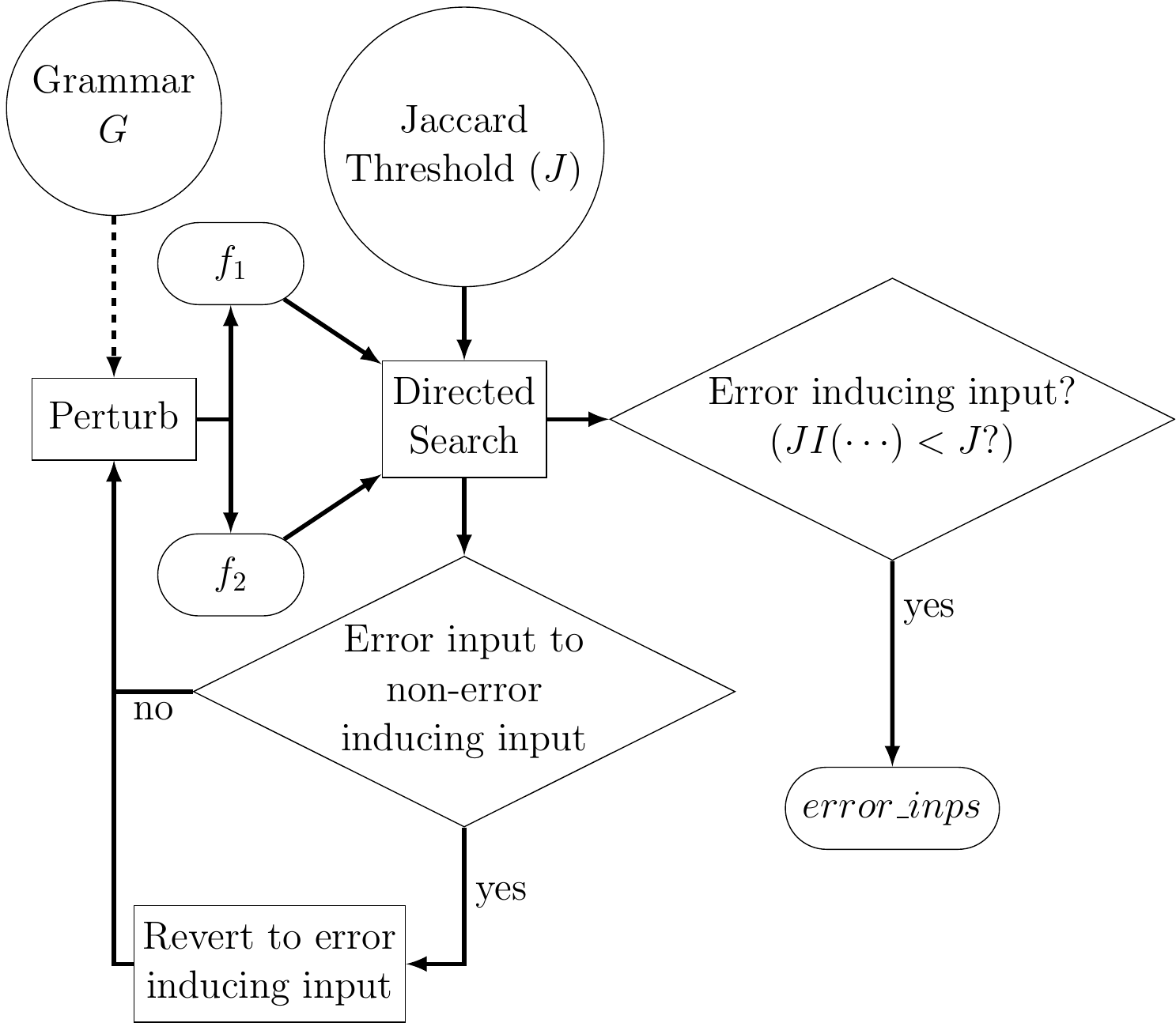}
\end{center}
\caption{Workflow of \FT
}
\label{fig:block-diagram}
\end{figure}

\begin{algorithm}[h]

    \caption{Directed Search}
    {\small
    \begin{algorithmic}[1] 
        \Procedure{directed\_search}{$f_1$, $f_2$, $S$, $J$, $G$} 
            \State $error\_inps \gets$  $\phi$
            \LineComment {\textsf{$N$ is the number of iterations in the
search}}
			\State $I_{cur} \gets S$
			\State $Eval_{S} \gets $ {\textsf{Evaluate}}($f_1(S),
			f_2(S), J$)
			\If{$Eval_{S}$ is True}
				\State $error\_inps \gets error\_inps \cup \{S\}$ 
			\EndIf
			
            \For{$i $ in (0, N)}
            	\LineComment {\textsf{See Algorithm~\ref{alg:perturbation}}}
            	\State $I_{cand} \gets $ {\textsf{Perturb}}($I_{cur}$, $G$)
            	
            	\LineComment {\textsf{Evaluate if $Eval_{cand}$ and $Eval_{cur}$ are error inducing}}
            	\State $Eval_{cand} \gets $ {\textsf{Evaluate}}($f_1(I_{cand}),
				f_2(I_{cand}), J$)
				
				\State $Eval_{cur} \gets $ {\textsf{Evaluate}}($f_1(I_{cur}),
				f_2(I_{cur}), J$)

				\If {$Eval_{cand}$ is True}
            		\State $error\_inps \gets error\_inps \cup \{I_{cand}\}$
            	\EndIf
           		\State $I' \gets I_{cand}$
           		\LineComment {\textsf{This condition prevents the process from
           		going to}}
           		\LineComment{\textsf{a non-error inducing input from an error 
           		inducing}}
           		\If {\label{ln:backtrack} $Eval_{cand}$ is False and $Eval_{cur}$ is True}
            		\State $I' \gets I_{cur}$
            	\EndIf
            	\State $I_{cur} \gets I'$
            \EndFor
           \Return $error\_inps$
            
        \EndProcedure
    \end{algorithmic}}
    \label{alg:directedSearch}
 \end{algorithm}
 
 \smallskip\noindent

 \begin{algorithm}[h]

    \caption{Perturbation}
    {\small
    \begin{algorithmic}[1] 
        \Procedure{perturb}{$I$, $G$} 
            \State Let $i$ be a random terminal symbol in $I$
			\State $T \gets \tau_G(I)$
			\State Let $n$ be the leaf node in $T$ that contains $i$
			\LineComment \textsf{Parent node of $n$, i.e., production rule which 
			creates $i$}
			\State Let $P \gets $ Parent($n$)
			\State Let $\sigma$ be a set of all terminal symbols in 
			production rule $P$ 
			\State Let $K \gets \{k\ |\ k \in \sigma \setminus \{i\} \}$
			\If {$K = \emptyset$}
				\Return print ("Cannot Perturb Terminal")
			\EndIf
			\State Let $i'$ be a randomly chosen terminal symbol in $K$
		   \LineComment \textsf{Replace $i$ with $i'$ in $I$}
           \State $I' \gets I[\![i \rightarrow i']\!]$
           \Return $I'$
            
        \EndProcedure
    \end{algorithmic}}
    \label{alg:perturbation}
 \end{algorithm}
 
\begin{algorithm}[h]
    \caption{Evaluate}
    {\small
    \begin{algorithmic}[1]  
        \Procedure{Evaluate}{$A$, $B$, $J$} 
        	\If {$JI(A, B) < J$}
				\Return True		
			\Else
				\Return False
			\EndIf
            
        \EndProcedure
    \end{algorithmic}}
    \label{alg:evaluate}
\end{algorithm}

An overview of our overall approach can be found in \autoref{fig:block-diagram}. 
The main contribution of this paper is an automated directed test generator for
grammar-based inputs. Our \revise{applications under test (AUTs)} 
are machine-learning models that accept inputs 
conforming to certain grammars. \revise{The initial input to \FT 
(\autoref{fig:block-diagram}) is randomly generated from the grammar.}
Subsequently, \FT involves two major steps:  1) Directed Search 
(\textsc{directed\_search}) in the input domain $\mathbb{I}_{G}$ and 
2) Input perturbation (\textsc{Perturb}). In the following, we describe these 
two procedures in detail. 

%

\subsection{{Directed Search in \FT}}
\label{sec:dir-search}
The motivation behind our directed search (cf. procedure \textsc
{directed\_search}) is to contain the search in the subset of the input space 
$\mathbb{I}_{G}$ where the errors are localised. Conceptually, robustness in
machine-learning  captures a phenomenon stating that a slight change in the input
does not change the output dramatically in well-trained machine-learning 
models~\cite{robustness_classifier}. This means that error inducing inputs are
likely to cluster together in certain input subspace of well-trained models.
The goal of \FT is to discover these subspace(s) and the instances of erroneous 
behaviours that are present in these subspace(s). 

The directed search requires the two classifiers under test ($f_1, f_2$), a
grammar ($G$), a randomly generated initial input
conforming to the grammar $G$ and a Jaccard Threshold ($J$). 
%
The search algorithm evaluates the input $S$ initially. It finds the
Jaccard Index (cf. Definition \autoref{def:jaccard-index}) of the output 
sets $f_1(S)$ and $f_2(S)$. If $\mathit{JI} \left ( f_1(S), f_2(S) \right )$ 
is lower than the threshold $J$, then the input $S$ is added to the set 
$error\_inps$ and $S$ is assigned to $I_{cur}$ for the first iteration. 
Intuitively, this means $S$ falls in the region of error inducing input 
subspace and thus, it is likely to lead to more error inducing inputs 
via perturbation. 

At any point, the directed search process keeps track of two crucial inputs, 
namely $I_{cur}$ (Current input) and $I_{cand}$ (Candidate input), respectively. 
$I_{cur}$ is the input that was discovered in the latest iteration of the 
directed search. $I_{cur}$ can be an error or non error input 
(cf. Definition \autoref{def:error}). $I_{cand}$ is the perturbed input resulting 
from $I_{cur}$ (cf. procedure \textsc {Perturb}), i.e., 
$I_{cand} = \mathit{Perturb}(I_{cur})$ according to Definition~\ref{def:perturbation}. 
%
%
The goal of \FT with the perturbation is to either discover more error inputs 
(if $I_{cur}$ is already an error input) or to discover a subspace of 
$\mathbb{I}_{G}$ which contains error inputs (if $I_{cur}$ is a non-error
input).


It is crucial for \FT to keep track of the transition sequence between error 
and non-error inducing inputs during the test generation process. Specifically, 
\FT prevents the test generation process from entering an input subspace containing 
non-error inducing inputs from the subspace containing an error inducing inputs. 
The rationale behind such a strategy is backed by the robustness property of 
machine-learning models, as perturbing error inducing inputs is certainly more 
effective than perturbing non-error inducing inputs. 
As an example, let $I_{cur}$ be ``{\em Mary saw my dog}", which is an input that
causes erroneous behaviour (cf. Definition \autoref{def:error}) in the classifiers
$f_1$ and/or $f_2$. It is part of a subset of $\mathbb{I}_{G}$ in these classifiers
which causes these classifiers to exhibit erroneous behaviours. Let the
perturbation of $I_{cur}$ result in  ``{\em Bob saw my dog}", which is assigned 
to $I_{cand}$. Let us assume $I_{cand}$ does not show erroneous behaviours and thus, 
is located in an input subspace that is unlikely to exhibit erroneous behaviours. 
In this case, therefore, we discard $I_{cand}$ (line~\ref{ln:backtrack} in 
Algorithm~\ref{alg:directedSearch}) and backtracks the test generation process 
to induce a different perturbation to $I_{cur}$.

In the case where $I_{cand}$ does induce an erroneous behaviour, the test 
generation process is focused to search in the vicinity of $I_{cand}$ to find 
more such inputs. In this case, we update $I_{cur}$ to the value in $I_{cand}$ 
and proceed to the subsequent iterations to repeat the perturbation steps. 

It is worthwhile to note that there are four possible transitions between inputs 
in each iteration. These are namely, {\em Non-error inducing input $\rightarrow$ 
Error inducing input}, {\em Error inducing input $\rightarrow$ Error inducing input}, 
{\em Non-error inducing input $\rightarrow$ Non-error inducing input} and {\em Error
inducing input $\rightarrow$ Non-error inducing input}. We only move out of a subspace 
of interest
of $\mathbb{I}_{G}$ in the last case. This is to avoid getting stuck in a region that does 
not contain error inducing inputs. 
%

 
\subsection{{Perturbation in \FT}}
\label{sec:perturbation}

%
The perturbation function has two responsibilities. The first responsibility 
of this function is to discover the input subspace $\mathbb{I}_{G}$ that 
contains erroneous inputs. The second responsibility is to explore this 
subspace to find instances of error inducing inputs in the same. 
The first responsibility is captured when the initial seed input $S$ is 
non-error inducing. As we can see in \autoref{fig:RQ1}, in some cases 
\FT produces several non-error inputs in the initial stages of the test 
generation process. This is because the initial input to \FT is in some 
part of the input subspace of $\mathbb{I}_{G}$ where the inputs 
show non erroneous behaviour. Thus, we need to perturb the input to get 
the process out of this subspace and find a subspace which 
shows erroneous behaviour. \FT continuously perturbs the input to find 
such a subspace. As we can see in \autoref{fig:RQ1}(a), eventually after 
$\approx$ 200 iterations, \FT finds the subspace of $\mathbb{I}_{G}$ 
where the inputs do indeed show erroneous behaviours.

The function \textsc{Perturb} chooses a random terminal $i$ from an input 
$I \in \mathbb{I}_{G}$. Then, we obtain the derivation tree $T = \tau_G(I)$. 
In this derivation tree, we find the leaf node that contains $i$ and discover
the parent of this node. This, in turn, gives the production rule $P$ that 
produced the terminal symbol $i$. 
%
%
In the next step we construct a set $K$, which includes all the terminal symbols 
we have found in the production rule $P$, except for the terminal symbol $i$. If 
we reconsider the example in \autoref{fig:steps-overview}(b), the set of such 
terminal symbols would be $K = $\{``{\em John}", ``{\em Bob}"\}. 
We choose a random terminal symbol $i' \in K$. We will use this terminal symbol to
replace $i \in I$ to create a new input $I'$. In the example, if we choose ``{\em 
John}", and replace  ``{\em Mary}", the new sentence $I' \in \mathbb{I}_{G}$ will 
be ``{\em John saw my dog}". 

As a result of the design of the perturbation function, the choice of grammar
plays an important role in the success of \FT. The idea of the perturbation is 
to generate a substantial number of inputs with similar derivation trees.

\subsection{{Similar Sentences and Perturbation}}
\label{sec:similarity-perturbation}

\begin{figure}[h]
\begin{center}
\includegraphics[scale=0.8]{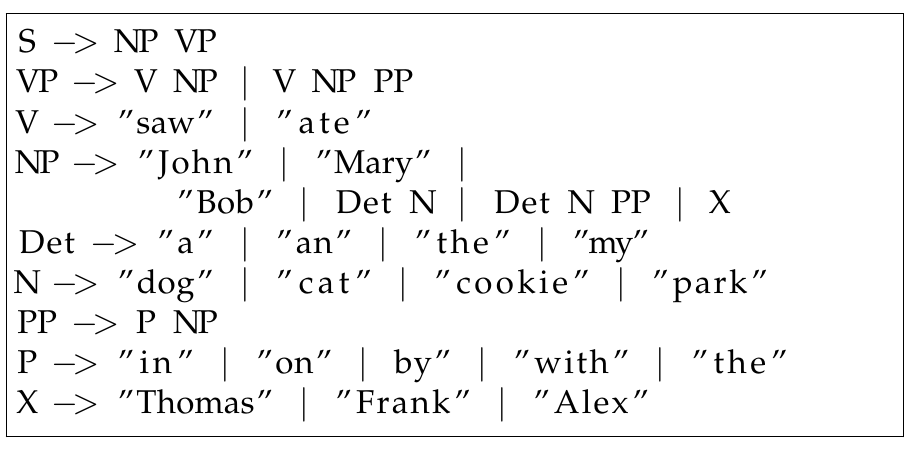}
\end{center}
\caption{Modified Example Grammar}
\label{fig:toy-gram-2}
\end{figure}

It is important to note that sentences that might appear similar, 
may not be considered similar (cf. Definition~\ref{def:similar-trees}) in \FT. 
This difference is best brought out with the help of an example.
Concretely, consider the grammar seen in \autoref{fig:toy-gram-2} and the 
derivation tree of the sentence ``{\em Frank saw my dog}" generated from this 
grammar. 

\begin{figure}[h]
\begin{center}
\includegraphics[scale=1.0]{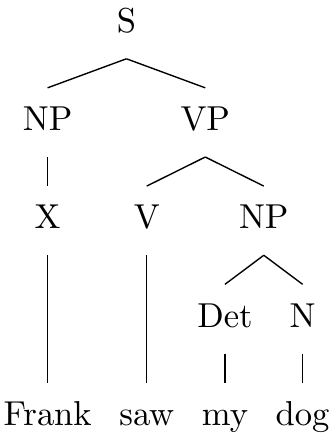}
\end{center}
\caption{Derivation tree for ``{\em Frank saw my dog}" }
\label{fig:diff-tree}
\end{figure}

Two sentences that conform to this grammar are $I_1 =$ ``{\em Mary saw my dog}" 
and $I_2 = $ ``{\em Frank saw my dog}". The derivation tree for $I_1$ can be seen 
in \autoref{fig:steps-overview}(a). As we can clearly see the structures of the 
derivation trees for $I_1$ and $I_2$ are different and as a result, these similar 
sentences are not considered similar inputs in \FT 
(cf. Definition \autoref{def:similar-trees}). In other words, \FT considers inputs 
to be {\em similar} only if they are derived {\em similarly} from the candidate 
grammar. This is a stricter condition over the similarity of the actual sentences. 
 The similar sentences (cf. Definition \autoref{def:similar-trees}) for $I_1$ would 
be  ``{\em Bob saw my dog}" and ``{\em John saw my dog}" according to \FT. Likewise, 
the similar sentences for $I_2$ would be ``{\em Thomas saw my dog}" and
``{\em Alex saw my dog}".

%



\section{Results}
\label{sec:results}

\subsubsection*{\textbf{Experimental Setup}}
\begin{figure}[h]
\includegraphics[scale=0.65]{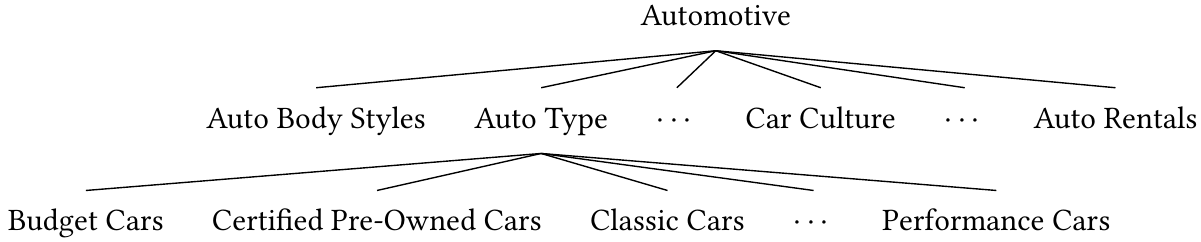}
\caption{IAB Content Taxonomy hierarchy}
\label{fig:iab}
\end{figure}

We evaluate \FT across three industrial text classification models provided by
uClassify~\cite{uclassify-url}, Aylien~\cite{aylien-url} and Rosette~\cite{rosette-url} 
text analytics. We have chosen these classifiers for two reasons. Firstly, these 
service providers are used in industry scale, such as in Amazon, Airbnb,  
Microsoft and Oracle among others. Secondly, our chosen service providers use 
text classifiers that categorise input text into a standard text classification 
taxonomy called the IAB content taxonomy~\cite{iab-taxonomy-url}. 
At a broader perspective, such a classification taxonomy acts as a
guideline on the types of classes that a text can be categorised to. 
In other words, this ensures a standardisation of classes across a 
variety of text classifiers.
%
%
A sample classification via the IAB (Interactive Advertising Bureau) Content Taxonomy can be 
found in \autoref{fig:iab}. 
``Automotive" is the broadest level of classification (Tier 1). Underneath this 
classification, there exists increasingly specific categories. For instance, Tier 2 
under the category "Automotive" includes ``Auto Body Styles", ``Auto Type" and 
``Auto Technology". Tier 3 is the most specific classification. Examples
under ``Auto Type" include ``Budget Cars", ``Classic Cars", ``Concept Cars" etc. 
For our evaluation, we have only considered the top tier classification. This is 
because we expect the classifier to at least have similar classification at the 
broadest category (i.e. Tier 1).

As we leverage differential testing, we aim to validate whether 
the output classes from two different text classifiers are similar. Since 
all our classifiers implement the IAB content taxonomy, we can compare their 
outputs coherently. Subsequently, we guide our test generation methodology 
to discover inputs that lead to vastly dissimilar classifier outputs 
(according to the IAB content taxonomy). 
%
We access the services of uClassify, Aylien and Rosette via client-side APIs. 
We engineer each API call to classify a sentence (as automatically generated 
via \FT) and to return a set of at most the five most likely results. 
For each test environment, we consider a pair of classifiers from different 
service providers to facilitate differential testing. To check the similarity 
between classifier outputs, we compute the Jaccard Index of outputs from two 
classifiers. If the computed Jaccard Index is below a certain threshold $J$ 
(c.f. Definition~\ref{def:error}), then we consider the input, leading to 
the respective Jaccard Index, as erroneous for at least one of the text classifier. 
The threshold $J$ is user defined and we evaluate \FT to check its sensitivity 
with respect to the threshold $J$. \revise{The threshold $J$ that we use for 
our evaluations can be found in \autoref{table:thresholds}.}

\revise{Although Google~\cite{google-cloud-nlp-url}, Baidu 
\cite{baidu-senta-url}, Facebook \cite{facebook-fasttext-url}
and Amazon \cite{amazon-nlp-url} do have certain NLP solutions, 
they do not offer a standard solution. \FT requires classifiers 
that are trained to provide a standard set of classification 
outputs or on a standard taxonomy. This is to have a reasonable 
expectation that the classifiers should have the same kind of 
outputs. As the training and testing data are proprietary for 
classifiers provided by Google~\cite{google-cloud-nlp-url}, 
Baidu~\cite{baidu-senta-url}, Facebook~\cite{facebook-fasttext-url}
and Amazon~\cite{amazon-nlp-url}, we cannot expect to have their 
outputs to be similar. In contrast, all our subject classifiers 
are expected to classify according to the 
IAB Taxonomy~\cite{iab-taxonomy-url}. Thus, we can compare the 
outputs of our subject classifiers to locate erroneous inputs.}

For the sake of brevity, we refer to Aylien as $\mathbb{A}$, Rosette as 
$\mathbb{R}$ and uClassify as $\mathbb{U}$ for the rest of this section. 
We also use the notations in \autoref{table:eval-notation} to describe 
the evaluation results. 

\begin{table}[t]
	\centering
	\caption{Notations used in Evaluation}
	\vspace*{-0.1in}
	\label{table:eval-notation}
	\begin{tabular}{| c | p{6cm} | }
	\hline
	\#inputs & Total number of \revise{unique} generated test inputs \\ \hline
	
	\#err & Total number of \revise{unique} erroneous inputs  \\ \hline
	
	$err_r$ & $\frac{\#err}{\#inputs}$ \\ \hline
	
	Imp\% &  Improvement of $err_r$ of \FT with respect to the $err_r$ of random test\\ \hline
\end{tabular}
	\vspace*{-0.15in}
\end{table}

\subsubsection*{\textbf{Choice of Input Grammars}}
We validate \FT using six different grammars (see Appendix for all the grammars 
used). As explained in \autoref{sec:method}, 
\FT essentially perturbs the derivation trees for an input generated from a grammar. 
Such a perturbation forms the crux of our systematic test generation while searching 
the neighbourhood of an erroneous input. We consider two inputs to be in the same 
neighbourhood if their derivation tree have the same structure 
(cf. Definition~\ref{def:error}). Thus, to continue test generation via \FT, the chosen 
grammar must encode a substantial number of inputs with the same derivation tree 
structure (see \autoref{fig:steps-overview}). To this end, we chose grammars that 
support production rules with multiple possible terminal symbols. For example, 
consider the grammar used in \autoref{sec:overview}. In this grammar, production 
rules from each non-terminal $V$, $NP$, $Det$, $N$ and $P$ lead to multiple possible 
terminal symbols. 


\subsubsection*{\textbf{Key Results}}
We construct three possible pairs of classifiers from the three text classifiers 
under test. Each pair of classifiers were validated with the six subject grammars 
chosen for evaluation. \autoref{table:average} outlines our key findings averaging 
over all such evaluation scenarios. The average is calculated over a varying 
threshold $[0.1,0.3]$ (cf. Definition~\ref{def:error}) to check the 
(dis)similarity of classifier outputs.  

In our evaluation, we intend to check whether our directed strategy indeed 
improves the state-of-the-art test generation methodologies for arbitrary 
machine-learning models. However, to the best of our knowledge, there does not exist 
any directed strategy for grammar-based test input generation with the objective to 
uncover errors in such models. Thus, to  evaluate the effectiveness of \FT, we compare 
it with a strategy that randomly generates sentences (Random) conforming to the input 
grammar and employs differential testing as embodied within \FT. We aim to show that 
if \FT generates more error inducing inputs than Random, then it is a step forward 
in designing directed, yet scalable methods for grammar-based test input generation 
targeting arbitrary machine-learning models.

As observed in \autoref{table:average}, \FT outperforms Random  
by a significant margin (up to 54\%). 
We attribute this improvement to the directed test 
strategy integrated within \FT. Specifically, \FT discovers more erroneous inputs 
than Random by exploiting the robustness property of common machine-learning models 
and realising this via a focused search in the neighbourhood of already discovered 
erroneous inputs. To evaluate the effectiveness of \FT in detail, we have answered 
the following research questions (RQs).   


\begin{table*}[h]
\caption{\revise{Key Results (the initial input is not erroneous)}}
\centering
\arrayrulecolor{black}
\begin{tabular}{!{\color{black}\vrule}c!{\color{black}\vrule}c!{\color{black}\vrule}c!{\color{black}\vrule}c!{\color{black}\vrule}c!{\color{black}\vrule}c!{\color{black}\vrule}c!{\color{black}\vrule}c!{\color{black}\vrule}c!{\color{black}\vrule}c!{\color{black}\vrule}c!{\color{black}\vrule}} 
\arrayrulecolor{black}\cline{1-1}\arrayrulecolor{black}\cline{2-11}
\multicolumn{1}{!{\color{black}\vrule}l!{\color{black}\vrule}}{} & \multicolumn{3}{c!{\color{black}\vrule}}{\FT} & \multicolumn{3}{c!{\color{black}\vrule}}{\FT - No Backtrack} & \multicolumn{3}{c!{\color{black}\vrule}}{Random} & \%impr                                      \\ 
\arrayrulecolor{black}\cline{1-1}\arrayrulecolor{black}\cline{2-11}
\multicolumn{1}{!{\color{black}\vrule}l!{\color{black}\vrule}}{} & \#inputs & \#errs  & $err_r$ & \#inputs & \#errs  & $err_r$                                   & \#inputs & \#errs  & $err_r$ & \multicolumn{1}{l!{\color{black}\vrule}}{}  \\ 
\hline
$\mathbb{R}$-$\mathbb{A}$                                                  & 1948.33  & 1757.50 & 0.90                      & 1953.17  & 1126.17 & 0.58                                     & 1798.33  & 1276.67 & 0.71                        & 27.06                                       \\ 
\hline
$\mathbb{U}$-$\mathbb{A}$                                                & 1920.33  & 1686.67 & 0.88                      & 1947.67  & 983.67  & 0.51                                     & 1778.67  & 1305.17 & 0.73                        & 19.70                                       \\ 
\hline
$\mathbb{R}$-$\mathbb{U}$                                               & 1917.17  & 1312.17 & 0.68                      & 1954.67  & 741.50  & 0.38                                     & 1798.00  & 797.50  & 0.44                        & 54.31                                       \\
\hline
\end{tabular}
\label{table:average}
\arrayrulecolor{black}
\end{table*}

\begin{table}[h]
\caption{\revise{Jaccard Thresholds}}
\centering
\begin{tabular}{| c | c | c | c | c | c | c | c |} 
\hline
   Grammars & \multicolumn{1}{l|}{A} & \multicolumn{1}{l|}{B} & \multicolumn{1}{l|}{C} & \multicolumn{1}{l|}{D} & \multicolumn{1}{l|}{E} & \multicolumn{1}{l|}{F}  \\ 
\hline
$\mathbb{R}$-$\mathbb{A}$  & 0.15                   & 0.15                   & 0.1                    & 0.15                   & 0.15                   & 0.15                    \\ 
\hline
$\mathbb{U}$-$\mathbb{A}$ & 0.15                   & 0.15                   & 0.1                    & 0.15                   & 0.1                    & 0.1                     \\ 
\hline
$\mathbb{R}$-$\mathbb{U}$ & 0.3                    & 0.15                   & 0.15                   & 0.3                    & 0.15                   & 0.15                    \\
\hline
\end{tabular}
\label{table:thresholds}
\end{table}

\begin{center}
\begin{tcolorbox}[width=\columnwidth, colback=gray!25,arc=0pt,auto outer arc]
\textbf{RQ1: Can the robustness property be leveraged for systematically testing real-world text classifiers?}
\end{tcolorbox}
\end{center}

\begin{figure*}[h]
\begin{center}
\begin{tabular}{cc}
\includegraphics[scale=0.5]{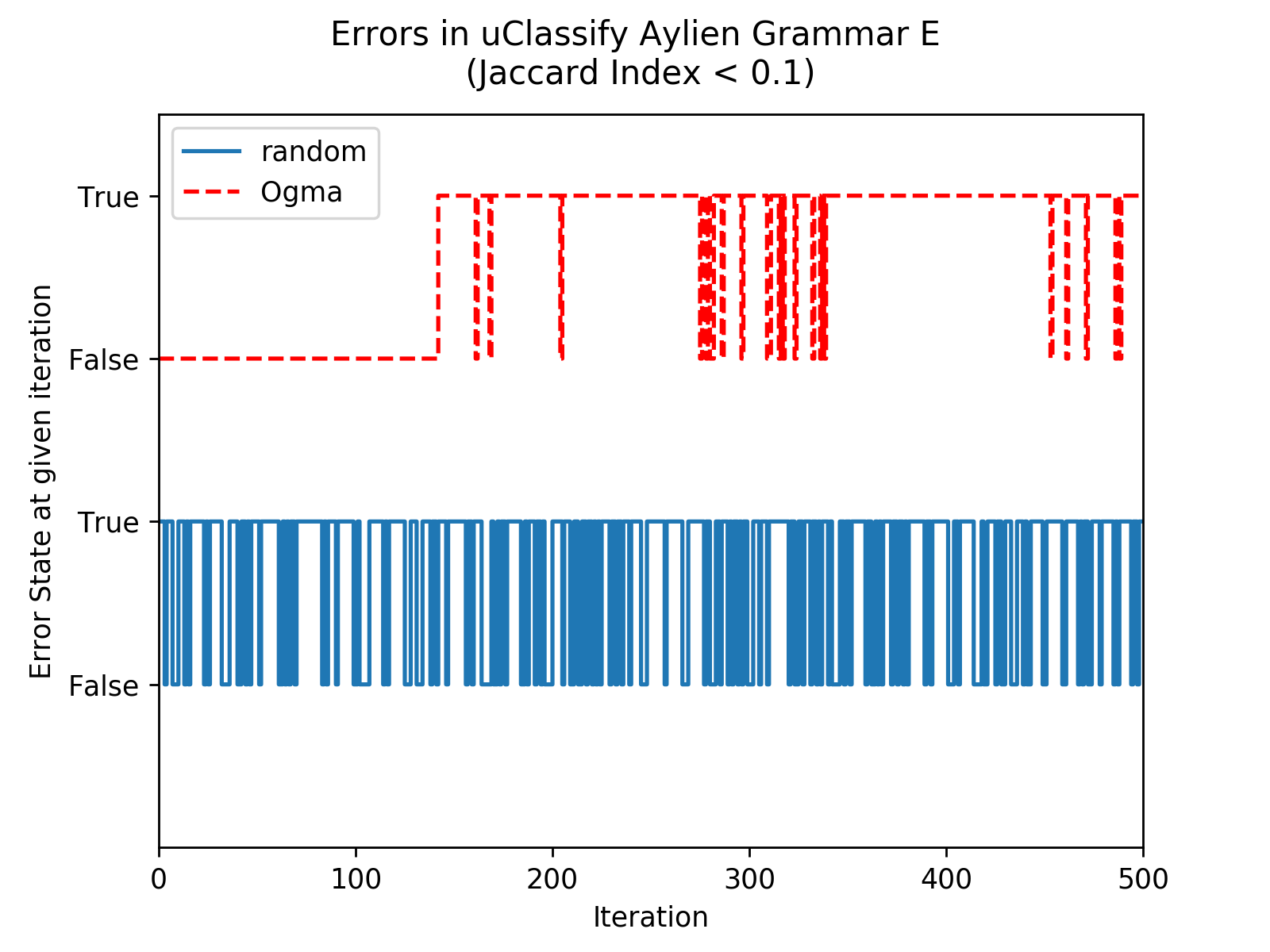} & 
\includegraphics[scale=0.5]{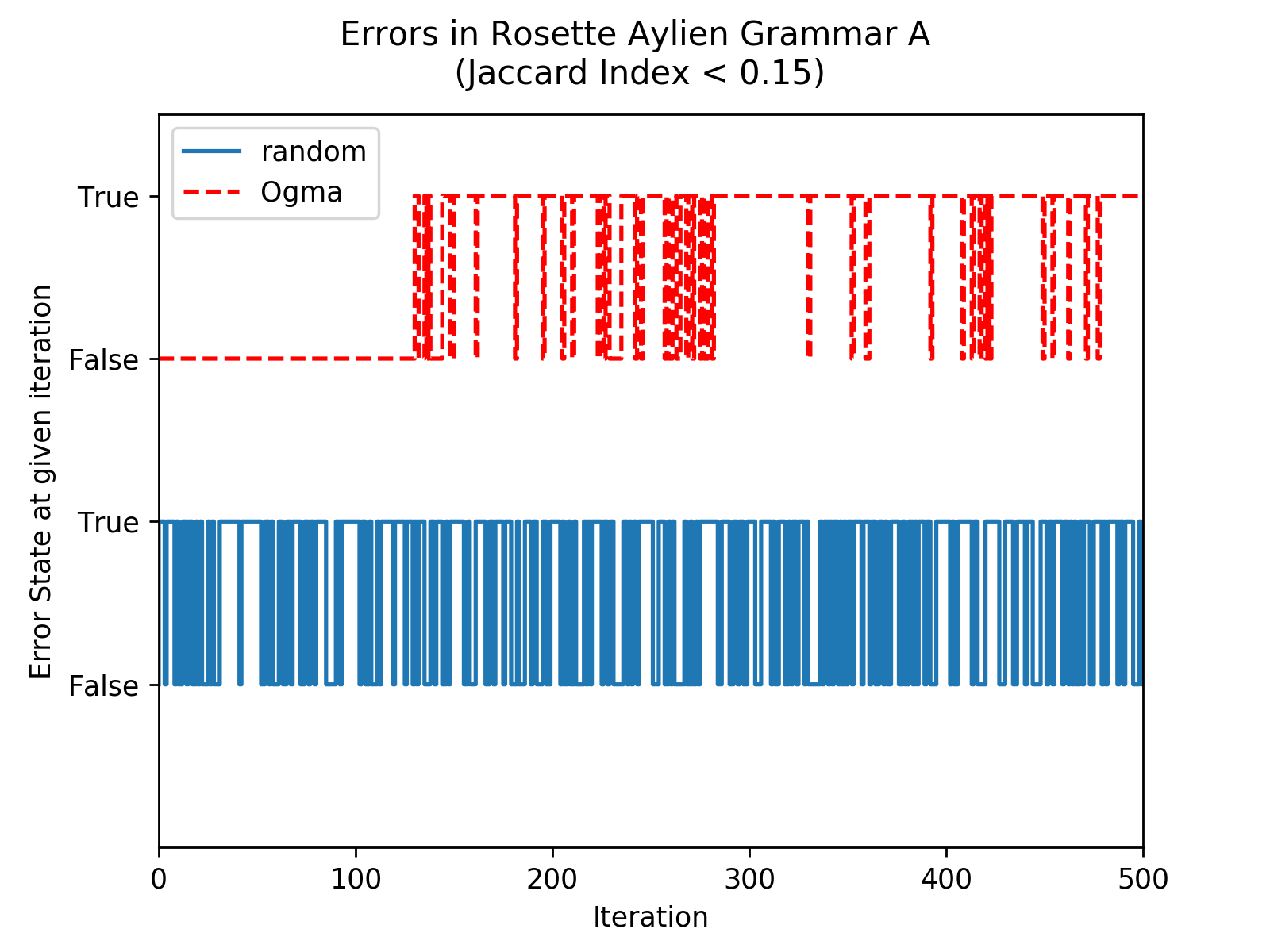} \\
{\bf (a)} & {\bf (b)} \\
\vspace*{-0.2in}
\end{tabular}
\end{center}
\caption{The rationale behind using robustness for error discovery}
\label{fig:RQ1}
\end{figure*}

\begin{figure}[h]
\includegraphics[scale=0.55]{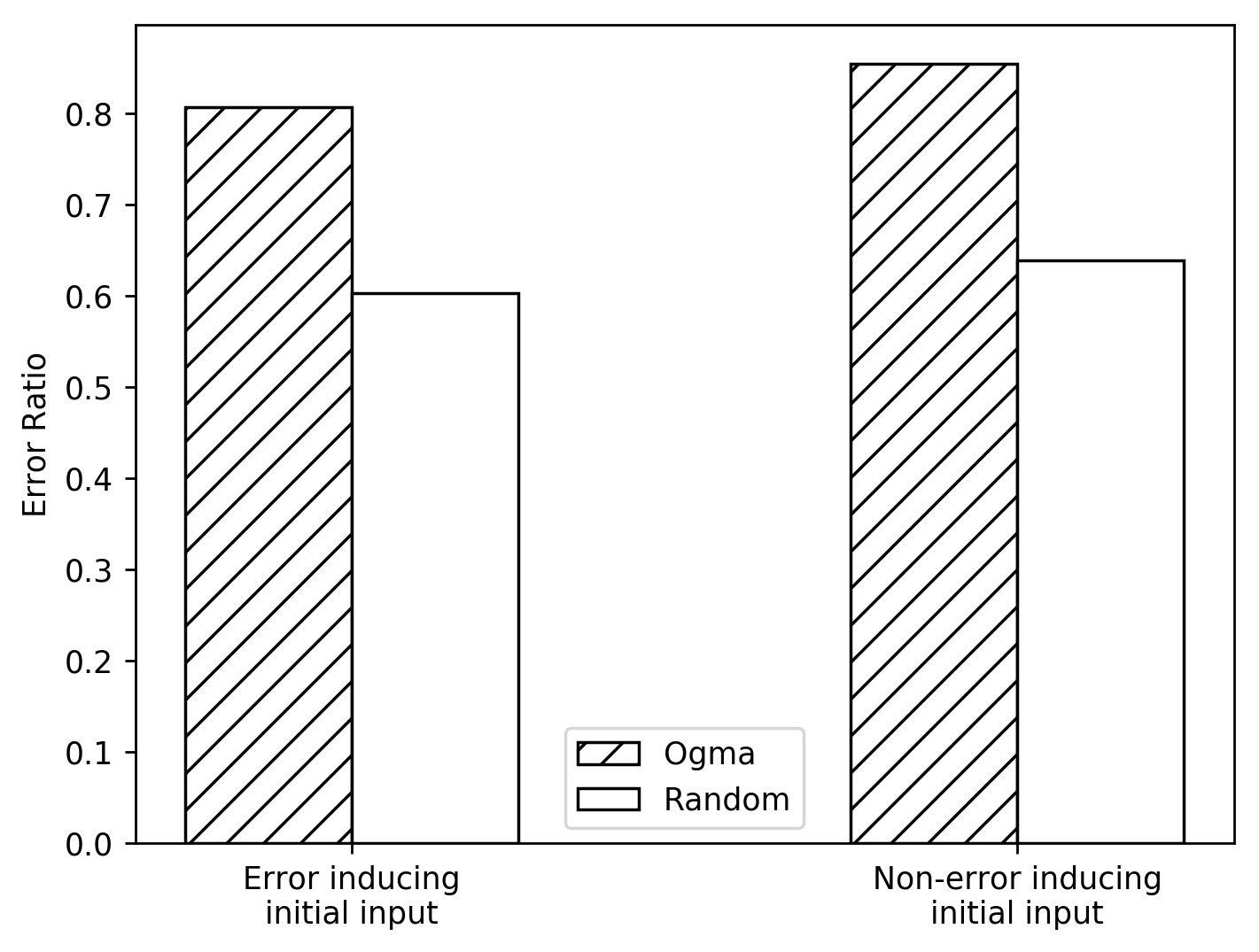}
\caption{Sensitivity of \FT w.r.t. the choice of initial input}
\label{fig:RQ3}
\end{figure}

Intuitively, robustness is a concept in machine learning which states 
that changing the input to any well trained machine learning model by 
some small value $\delta$ should not change the respective output 
dramatically. This means that similar inputs with similar outputs are 
likely to be clustered together. Thus, inputs similar to an error 
inducing input are also likely to lead to classification errors.



We present two cases (cf. \autoref{fig:RQ1}) where we have discovered 
clear indications of robustness playing a role in error discovery. 
In both \autoref{fig:RQ1}(a) and \autoref{fig:RQ1}(b), the graphs 
demonstrate whether a given test iteration leads to an error (output 
{\em True}) or not (output {\em False}). We have the following crucial 
observations from \autoref{fig:RQ1}. Firstly, we observe that once 
\FT finds an error inducing input (e.g. around iteration 200 in 
\autoref{fig:RQ1}(a)), it continues to discover more error inducing 
inputs (e.g. approximately until iteration 300). 
This is because \FT by design ensures to explore the neighbourhood of 
an input, whereas the robustness property of machine-learning 
models ensure 
that the neighbourhood of an error inducing inputs are 
likely to be error inducing too. Thus, \FT discovers a stretch of 
error inducing inputs, as observed between iterations 200 to 300 in 
\autoref{fig:RQ1}(a). Secondly, we observe from 
\autoref{fig:RQ1} that the directed search embodied in \FT is useful 
in terms of steering the execution to errors. For example, even though 
\FT discovers non-error inducing inputs, it can quickly revert to find 
error inducing inputs (e.g. approximately between iterations 300 and 350 in 
\autoref{fig:RQ1}(a)). Similar characteristics can also be observed in 
\autoref{fig:RQ1}(b). 
Due to the aforementioned characteristics, \FT significantly outperforms 
the random test generation. This validates our hypothesis that robustness 
of well trained machine learning model can indeed be leveraged to design 
systematic and scalable test generation methodologies. 



\begin{center}
\begin{tcolorbox}[width=\columnwidth, colback=gray!25,arc=0pt,auto outer arc]
\textbf{RQ2: How effective is \FT in terms of generating error inducing inputs for 
a variety of real-world text classifiers?}
\end{tcolorbox}
\end{center}

We measure the effectiveness of \FT as the ratio of the number of errors found 
with respect to the number of unique inputs generated (cf. \autoref{table:average}). 
On average the random approach discovers errors with a ratio 
of 0.7, 0.73 and 0.44 for the pairs of classifiers $\mathbb{R}$-$\mathbb{A}$, 
$\mathbb{U}$-$\mathbb{A}$, $\mathbb{R}$-$\mathbb{U}$, respectively. 
In comparison, the ratio of errors discovered via \FT are 0.9, 0.87 and 0.68 
respectively. Thus, on average, we observe that the directed search strategy 
embodied in \FT improves the effectiveness of test generation by 33.68\%.


\begin{center}
\begin{tcolorbox}[width=\columnwidth, colback=gray!25,arc=0pt,auto outer arc]
\textbf{RQ3: Does the effectiveness of \FT depend on initial test input?}
\end{tcolorbox}
\end{center}

We validated whether the initial input plays a major role in the effectiveness 
of \FT. To this end, we conducted two sets of experiments -- one where the initial 
input induced an error (i.e. two classifiers under test had dissimilar outputs) and 
another where the initial input was not an error inducing input (i.e. two classifiers 
under the test had similar outputs). We discovered that the initial input does not 
play a major role in the effectiveness of \FT. \autoref{fig:RQ3} outlines our 
finding. Specifically, \autoref{fig:RQ3} captures the average ratio of error inducing 
inputs discovered over all grammars and text classifiers. 


In \autoref{fig:RQ3}, the effectiveness of random test generation (in terms of 
discovering error inducing inputs) improves marginally by 1.57\% when initiated 
with an error inducing input. 
In general, the effectiveness of random test generation should be unaffected 
by the initial input, as each test input is generated independently. 
Concurrently, the effectiveness of \FT also improves 
by a negligible 4.94\% when the initial input is error inducing. Thus, we conclude 
that the initial input does not influence the effectiveness of our test generation 
methodologies significantly. 
%
However, as also seen in \autoref{fig:RQ3}, the relative improvement due to 
the directed strategy in \FT, over the random test generation strategy, remains 
over 33\% regardless of the category of initial input. 

\revise{The effectiveness of OGMA depends on the number of 
iterations it takes to reach the {\em first} non-error-inducing input. Subsequently, OGMA 
employs a backtracking strategy to prevent the exploration of non-error-inducing input space. 
When OGMA starts exploration with a non-error inducing input, it takes some iterations to 
reach the first input that induces error (e.g. see the flat portion of the OGMA curve until 
iteration 100 in Figure~8). However, when starting with a non-error-inducing input, we note 
that OGMA randomly samples the input space to reach an error-inducing input. From the Law of 
Large Number (LLN) in probability theory and as observed in the previous work~\cite{aequitas}, 
we can find the error inducing input with high probability within a few sampling instances. 
The number of such sampling instances are usually negligible when compared with the substantial 
number of test inputs (e.g. 2000 test iterations) generated by OGMA. In our evaluation, 
it takes an average of only 25.22 iterations to get to the first error 
inducing input. As a result, even though 
OGMA takes a few test iterations initially to reach the error-inducing input (when started 
with a non-error-inducing input), the effectiveness of OGMA is essentially unaffected by the 
type of initial test input.} 


\begin{table}[h]
\caption{Sensitivity of \FT w.r.t. the threshold $J$ for checking Jaccard Index}
\vspace*{-0.2in}
\label{classifiers}
\begin{center}
\begin{tabular}{| c | c | c | c | c | c | c | c |}
\cline{1-8}
& \multicolumn{3}{c|}{\FT} & \multicolumn{3}{c|}{Random} & \\ \cline{1-8}
& \#inputs & \#err & $err_r$ & \#inputs & \#err & $err_r$ & Imp\% \\ \cline{1-8}
0.05 & 172 & 41 & 0.23 &	 198	& 8 & 	0.04 & 489.97 \\ \cline{1-8}
0.15 & 189 & 100& 0.53 & 196 & 47 & 0.24 & 120.65 \\ \cline{1-8}
0.3 & 196 & 148 & 0.76 & 200& 124& 0.62 & 21.79 \\ \cline{1-8}
0.4 & 195 & 168 & 0.86 & 195 & 134& 0.69 & 25.37 \\ \cline{1-8}
0.45 & 196 & 184 & 0.93 & 193 & 187 & 0.96 & -3.11 \\ \cline{1-8}
0.5 & 193 & 189 & 0.98 & 197 & 184 & 0.93 & 4.85\\ \cline{1-8}
0.6 & 195& 184 & 0.94 & 197 & 184 & 0.93 & 1.0\\ \cline{1-8}
0.75 & 197 & 197 & 1 & 195 & 194 & 0.99 & 0.51 \\ \cline{1-8}

\end{tabular}
\end{center}
\label{table:thresholdSensitivity}
\vspace*{-0.1in}
\end{table}

\begin{figure}
\includegraphics[scale=0.55]{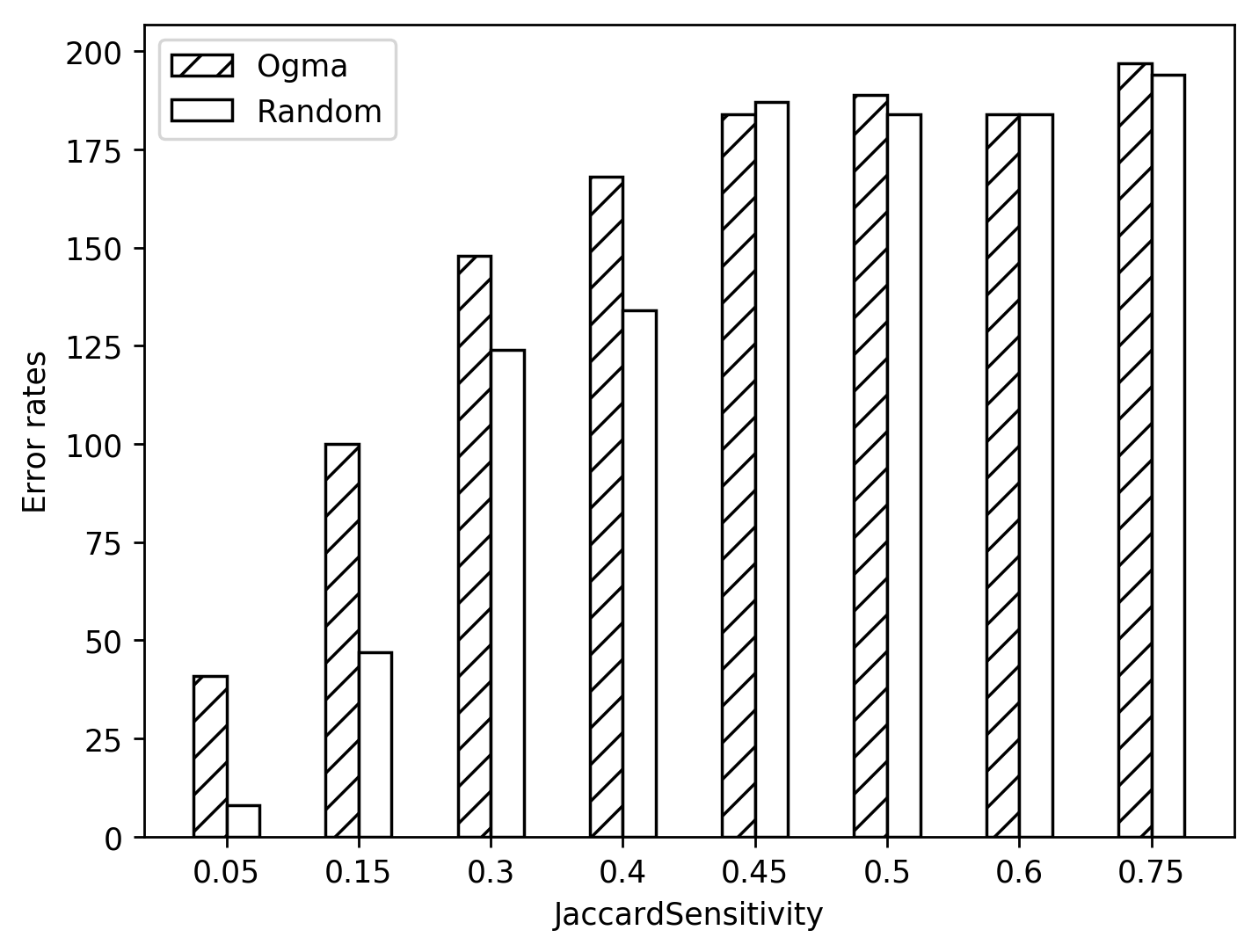}
\caption{Sensitivity of \FT w.r.t. the threshold $J$ for checking Jaccard Index}
\label{fig:RQ5}
\vspace*{-0.1in}
\end{figure}

\begin{center}
\begin{tcolorbox}[width=\columnwidth, colback=gray!25,arc=0pt,auto outer arc]
\textbf{RQ4: How sensitive is \FT w.r.t. the threshold $J$ (cf. Definition~\ref{def:error}) 
to check the similarity of classifier outputs? }
\end{tcolorbox}
\end{center}

\begin{figure*}[t]
\begin{center}
\begin{tabular}{ccc}
\includegraphics[scale=0.35]{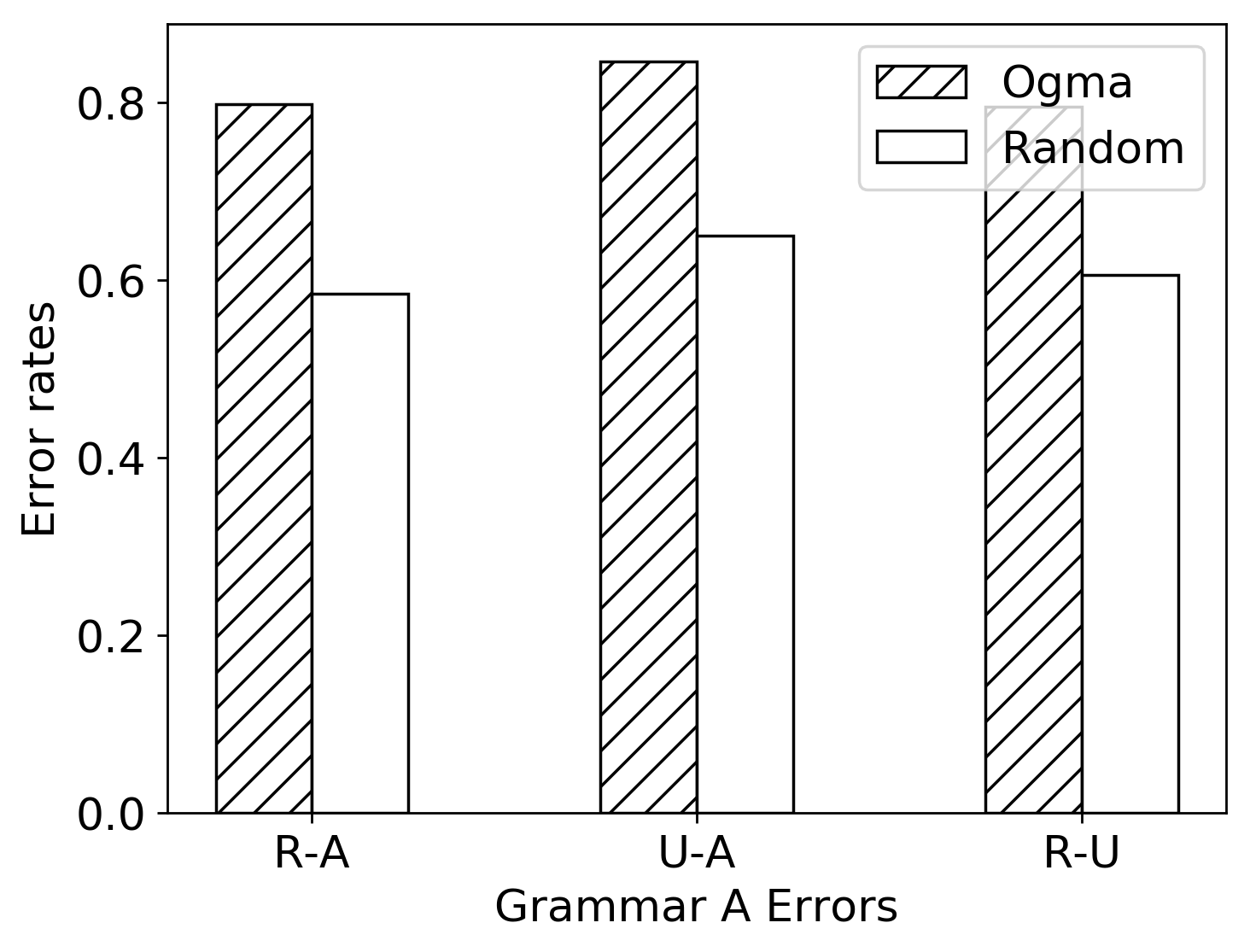} & 
\includegraphics[scale=0.35]{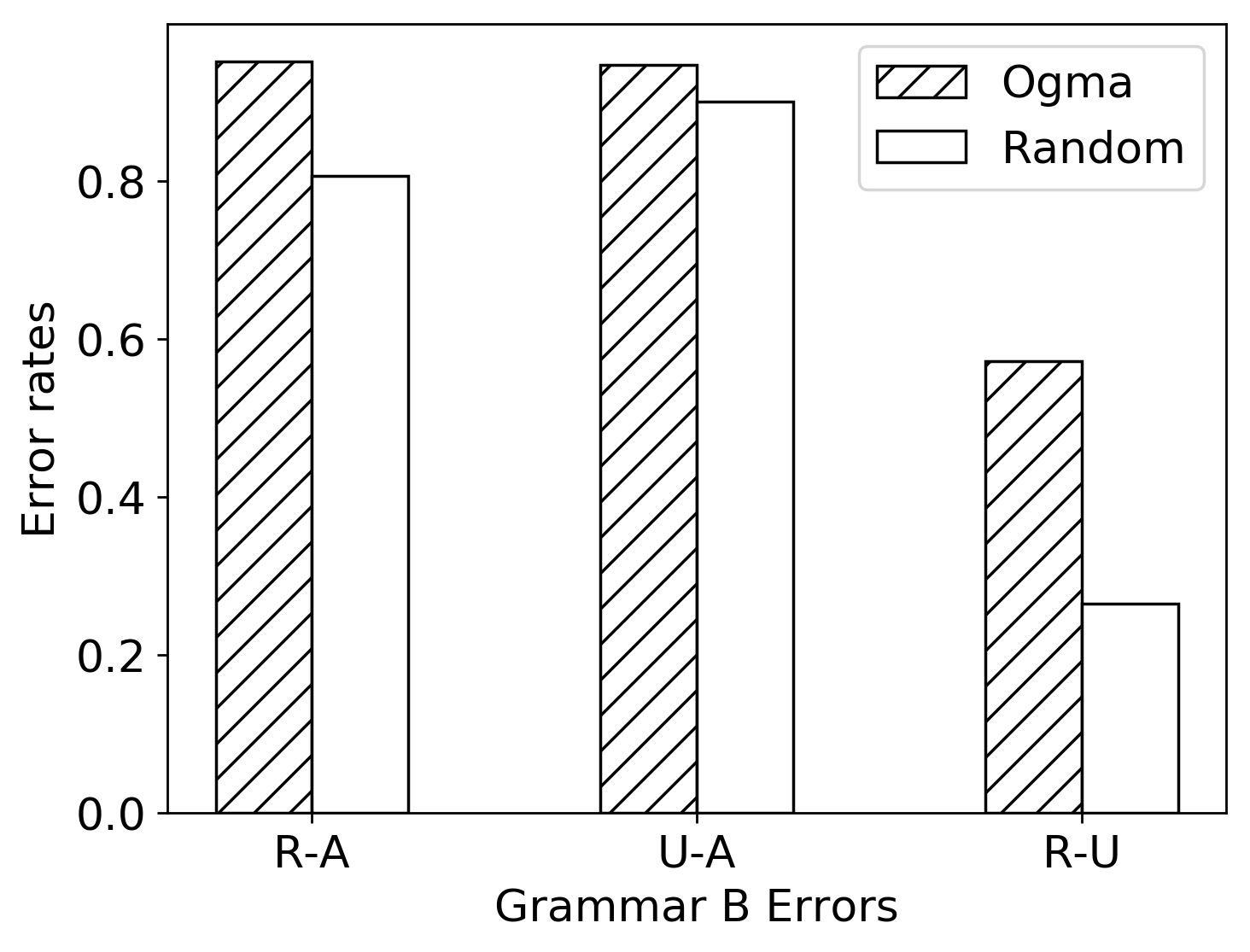} & 
\includegraphics[scale=0.35]{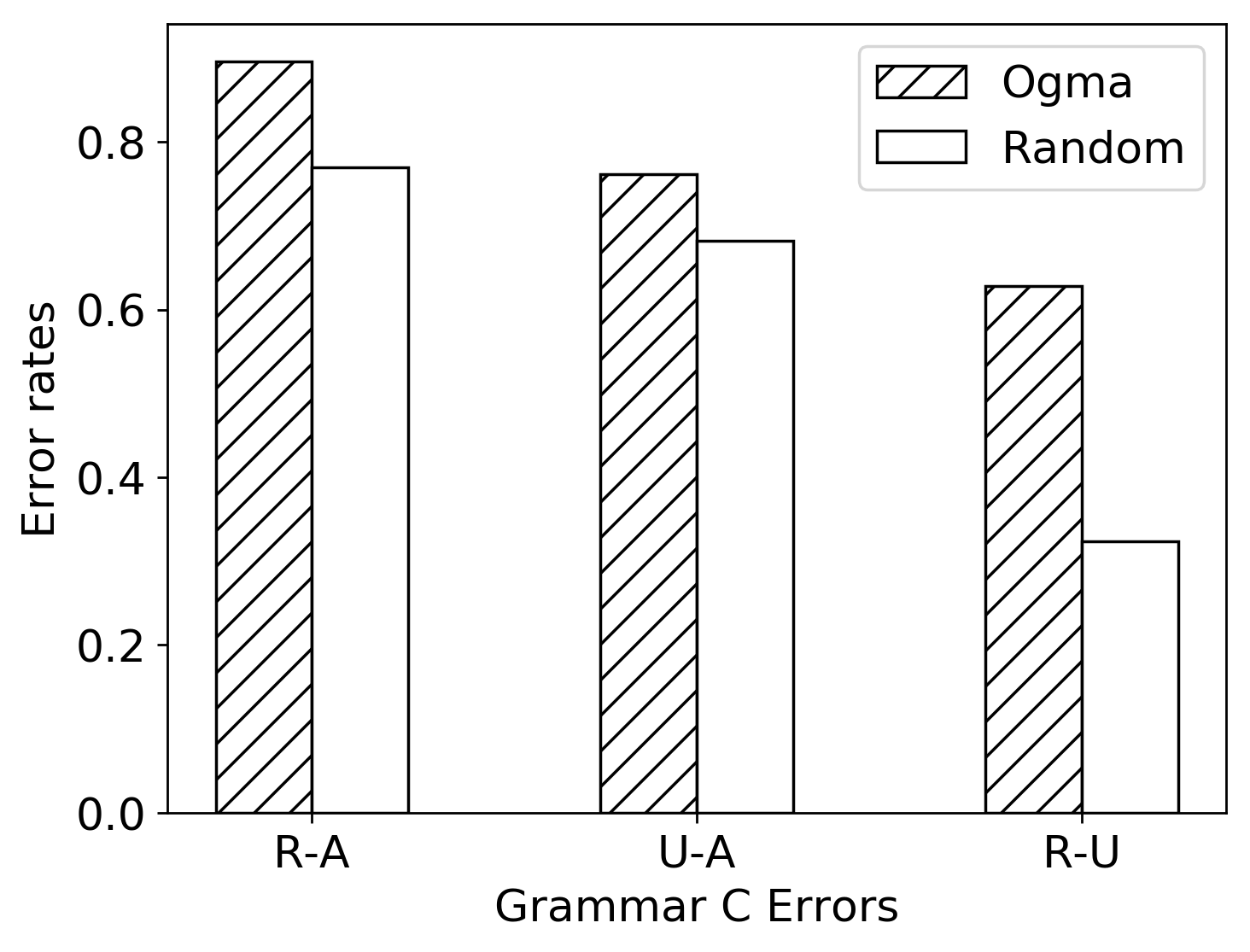}\\
{\bf (a)} & {\bf (b)} & {\bf (c)}\\
\\
\includegraphics[scale=0.35]{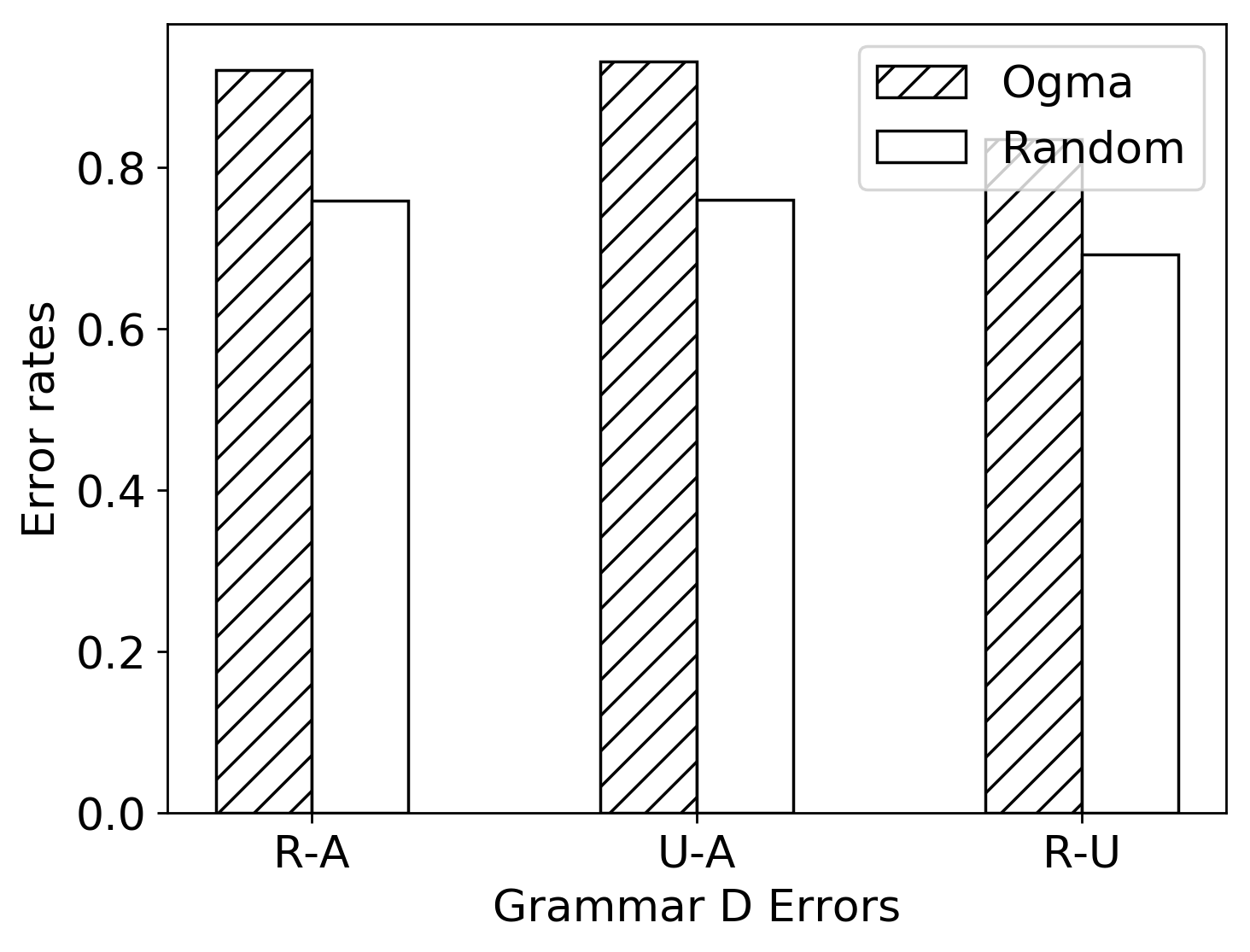} & 
\includegraphics[scale=0.35]{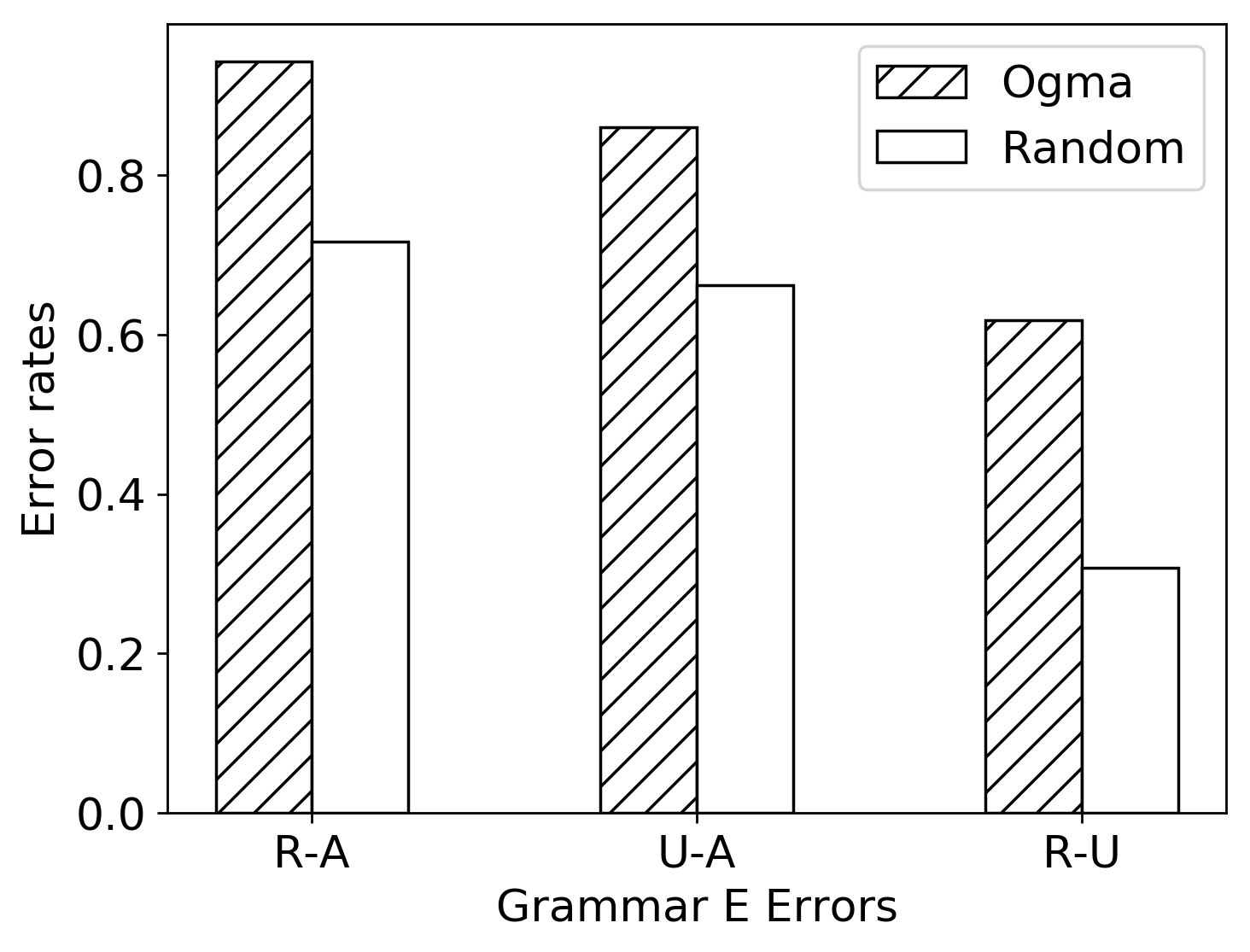} & 
\includegraphics[scale=0.35]{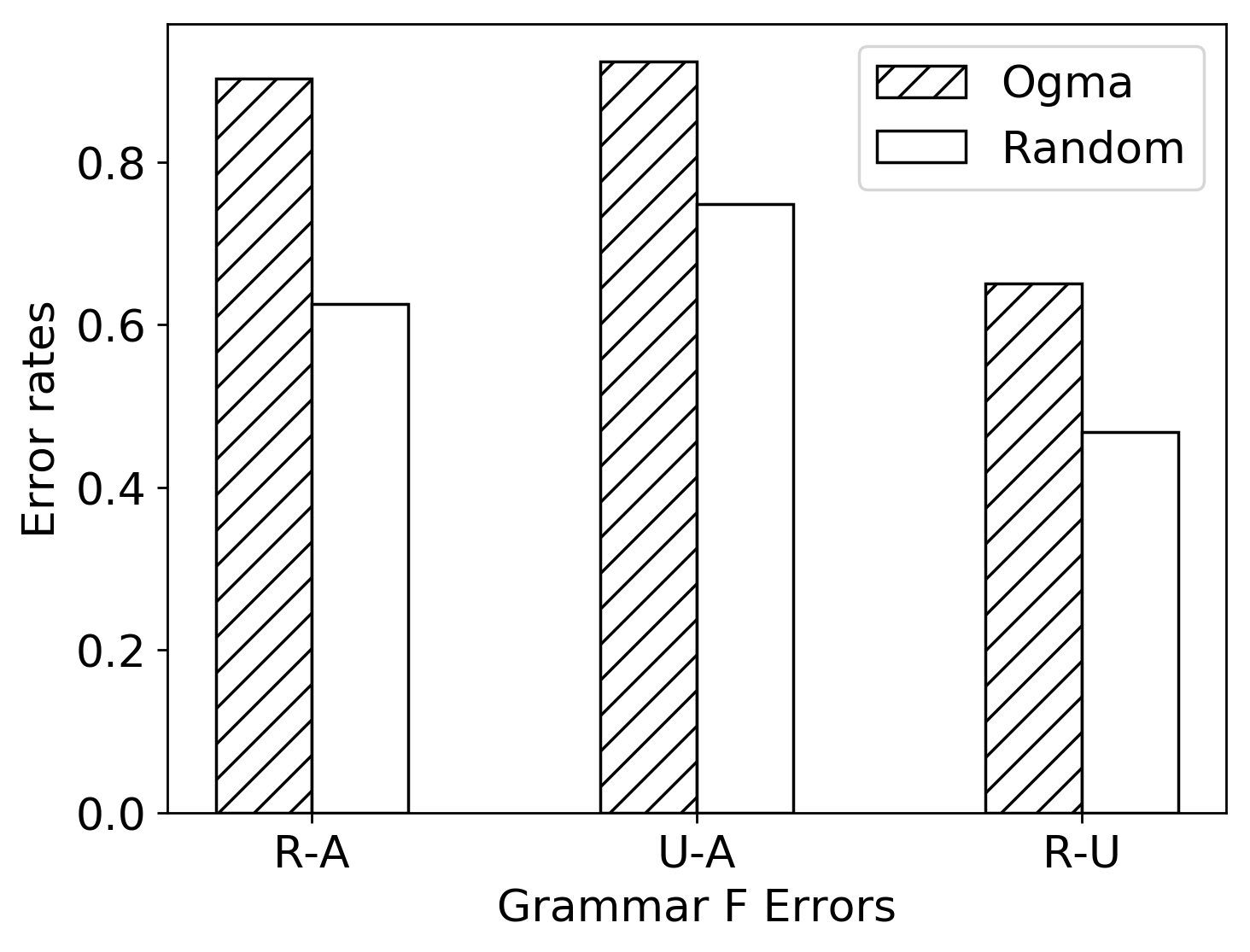}\\
{\bf (d)} & {\bf (e)} & {\bf (f)} \\
\end{tabular}
\end{center}
\caption{Sensitivity of \FT with respect to different grammars (see Appendix for the grammars)}
\label{fig:RQ4}
\end{figure*}

To answer this research question, we varied the threshold $J$ 
(cf. Definition~\ref{def:error}) for grammar A (see Appendix). 
The initial input for the test generation led to a Jaccard 
Index $> 0.15$, but $< 0.3$. Thus, for threshold values $[0.05,0.15]$, the 
initial input was not error inducing, whereas for threshold 
$\geq 0.3$, the initial input was error inducing. Finally, the reported 
values in this experiments were averaged over all possible pairs 
of classifiers (i.e. $\mathbb{R}$-$\mathbb{A}$, 
$\mathbb{U}$-$\mathbb{A}$ and $\mathbb{R}$-$\mathbb{U}$).

\revise{A small Jaccard index threshold captures very low overlap between two classifier 
outputs. Thus, for Jaccard Index threshold 0.1, an error inducing input exhibits vastly 
dissimilar outputs between two classifiers. We recommend to set such low Jaccard Index 
threshold when the two classifier outputs are expected to have some dissimilarity. Thus, 
an error inducing input will capture scenarios where the dissimilarity is substantial. 
We recommend to set high Jaccard Index threshold when the level of tolerance in a classifier 
output is low (for example, in safety-critical domains). In such cases, 
even a small deviation in classifier outputs can be classified as errors. We leave 
the choice of Jaccard Index threshold to the user, as it might depend on the type of 
applications being targeted.}


We observe a direct correlation between the chosen threshold $J$ 
and the effectiveness of \FT (cf. \autoref{table:thresholdSensitivity}). 
In particular, a low threshold value for $J$ (cf. Definition~\ref{def:error}) 
indicates that the tested classifier outputs have vastly dissimilar 
content. Thus, the lower the threshold $J$, the lower is also 
the probability to discover erroneous inputs. In other words, if 
we keep the threshold $J$ low, it is difficult for a random test 
generation strategy to discover error inducing inputs. As a result, 
for such scenarios, the directed test strategy in \FT outperforms 
random test generation by a significant margin (up to 489\%). 
In contrast, for a higher threshold $J$, even a slightly dissimilar 
classifier outputs might be categorized as errors. As such, for 
higher threshold (e.g. between 0.45 and 0.75), the effectiveness 
of \FT and the random test generation strategy is similar. 

\autoref{fig:RQ5} provides the trend of discovered error inputs 
with respect to the threshold $J$. The number of errors found by 
\FT is consistently higher than the random approach except for 
threshold value 0.45. For threshold value 0.45, random strategy 
is marginally better due to the ease of finding error inputs 
with high probability. 
%
The observations in \autoref{table:thresholdSensitivity} and in 
\autoref{fig:RQ5} reveal that \FT should be used for finding error 
inducing inputs where the error condition is strict (i.e. low 
threshold $J$ for the computed Jaccard Index). This is because such 
error inducing inputs are unlikely to be discovered via a random 
search, while \FT can discover these inputs effectively by leveraging 
the robustness property of machine-learning models. 

%
%

\begin{center}
\begin{tcolorbox}[width=\columnwidth, colback=gray!25,arc=0pt,auto outer arc]
\textbf{RQ5: How sensitive is \FT w.r.t. the chosen grammar? }
\end{tcolorbox}
\end{center}

\begin{table}[h]
\caption{Sensitivity of \FT w.r.t. Grammars}
\vspace*{-0.2in}
\label{classifiers}
\begin{center}
\begin{tabular}{| c | c | c | c | c |}
\hline
& \multicolumn{2}{c|}{Grammar A - F} & \multicolumn{2}{c|}{$G_{bad}$} \\ \hline
& \%unique inps & \%error inps & \%unique inps & \%error inps \\ \hline
$\mathbb{R}$-$\mathbb{A}$  & 97\% &	88\% &	52\% &	22\% \\ \hline
$\mathbb{U}$-$\mathbb{A}$ & 96\% & 	84\% &	49\% &	34\%  \\ \hline
$\mathbb{R}$-$\mathbb{U}$ & 96\% &	66\% &	51\% &	28\% \\ \hline

\end{tabular}
\end{center}
\label{table:badGrammar}
\vspace*{-0.1in}
\end{table}

\begin{figure}[H]
\begin{center}
\includegraphics[scale=0.8]{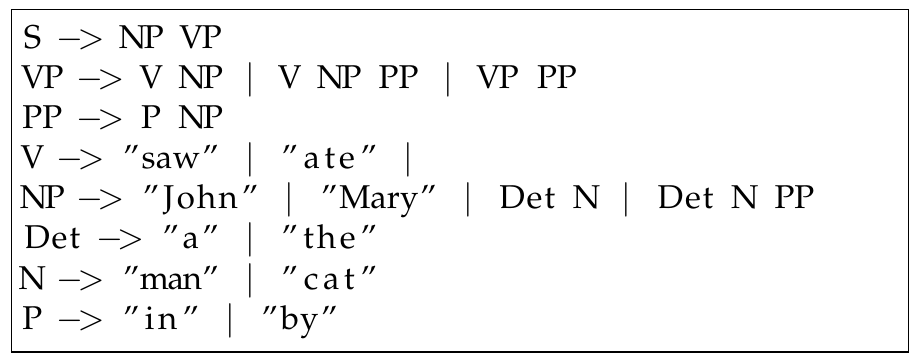}
\end{center}
\caption{Grammar with a few terminal symbols ($G_{bad}$)}
\label{fig:bad-grammar}
\end{figure}

As discussed earlier in this section, we employ a directed search in the 
neighbourhood of an error inducing input. This is accomplished by only 
perturbing a leaf node of the derivation tree, yet keeping the structure 
of the derivation tree similar. As such, we have chosen grammars (see 
Appendix) to generate a substantial number of test inputs by 
perturbing only leaf nodes of the derivation tree for a given input. 


We evaluate the effectiveness of \FT for six different grammars chosen for 
our evaluation and our findings are demonstrated via \autoref{fig:RQ4}. 
For each set of experiments, we measure the ratio of error inducing inputs 
(with respect to the total number of generated inputs) discovered for both 
random testing and \FT. As observed from \autoref{fig:RQ4}, our \FT approach 
is consistently more effective than random test generation and its effectiveness 
is not compromised across a variety of grammars.  
%
%
Specifically, we obtain a maximum improvement of up to 94\% (for classifiers 
$\mathbb{R}$-$\mathbb{U}$ and with Grammar C) and an average improvement of up 
to 33\% across all grammars and classifiers. 


\smallskip \noindent 
\revise{
\textbf{Sensitivity to Grammars with a few terminal symbols:}
}
\revise{
Additionally, we have also evaluated the three pairs of classifiers on a input
grammar with a few terminal symbols. 
This grammar, as seen in \autoref{fig:bad-grammar} 
has very few terminal symbols. This grammar is used with our \FT approach 
for 100 iterations. We aim to fine the number of unique inputs and the number of 
unique errors we can generate using this grammar. Intuitively, we do not expect the 
grammar with a few terminal symbols to be able to generate a lot of unique 
sentences because it has very few options for perturbations and it is likely that 
\FT won't be able to generate a lot of unique inputs.
}

\revise{
We measure the unique inputs and unique errors generated as a percentage of
the total number of inputs generated by \FT. The
data, as seen in \autoref{table:badGrammar} shows that the grammar with a few 
terminal symbols produces on an average only 52\% unique inputs, in contrast to an 
average of 96\% for the other grammars (Grammars A - F) and 28\% unique error 
inputs in comparison to the average of 79\%  unique error inputs of the ``good" 
grammars (Grammars A - F). 
}

\begin{center}
\begin{tcolorbox}[width=\columnwidth, colback=gray!25,arc=0pt,auto outer arc]
\textbf{RQ6: Does \FT find errors for  different use cases and classifiers?}
\end{tcolorbox}
\end{center}


\newrevise{
To evaluate RQ6, we have evaluated the sentiment analysis tool 
provided by Google's Natural Language API \cite{google-cloud-nlp-url}. We have 
employed differential testing with Rosette's sentiment analysis service 
\cite{rosette-url}. The outputs of the Rosette API are [``POSITIVE", ``NEGATIVE", 
``NEUTRAL"]. The Google API returns a score, $sc \in (-1, 1)$ and magnitude, 
$mg \in (0, \infty)$. Using the Interpreting sentiment analysis values as a 
guide~\cite{google-nlp-interpreting-sentiments}, we classify 
the output as [``NEGATIVE"], [``NEUTRAL"] and [``POSITIVE"] when 
$sc \in (-1, -0.25]$, $~sc \in (-0.25, 0.25)$ and $sc \in [0.25, 1)$, respectively.
An input is considered an error if the outputs of the Google sentiment analysis 
API and Rosette sentiment analysis API are different. 
}

\begin{table*}
\caption{Number of errors discovered in sentiment analysis using Rosette and 
Google sentiment analysis API}
\label{table:rq6-errors}
\centering
\arrayrulecolor{black}
\setlength{\fboxrule}{3pt}
\fcolorbox{blue}{white}{
\begin{tabular}{!{\color{black}\vrule}c!{\color{black}\vrule}c!{\color{black}\vrule}c!{\color{black}\vrule}c!{\color{black}\vrule}c!{\color{black}\vrule}c!{\color{black}\vrule}c!{\color{black}\vrule}c!{\color{black}\vrule}c!{\color{black}\vrule}c!{\color{black}\vrule}} 
\arrayrulecolor{black}\cline{1-1}\arrayrulecolor{black}\cline{2-10}
\multicolumn{1}{!{\color{black}\vrule}l!{\color{black}\vrule}}{} & \multicolumn{3}{c!{\color{black}\vrule}}{\FT} & \multicolumn{3}{c!{\color{black}\vrule}}{Random} & \multicolumn{3}{c!{\color{black}\vrule}}{\FT - No Backtrack}  \\ 
\arrayrulecolor{black}\cline{1-1}\arrayrulecolor{black}\cline{2-10}
Grammar                                                          & \#inputs & \#errs & $err_r$                          & \#inputs & \#errs & $err_r$                        & \#inputs & \#errs & $err_r$                                          \\ 
\hline
A                                                                & 195      & 160    & 0.82                           & 195      & 59     & 0.30                         & 196      & 83     & 0.42                                           \\ 
\hline
B                                                                & 194      & 173    & 0.89                           & 197      & 71     & 0.36                         & 197      & 100    & 0.51                                           \\ 
\hline
C                                                                & 199      & 161    & 0.81                           & 198      & 51     & 0.26                         & 196      & 60     & 0.31                                           \\ 
\hline
D                                                                & 198      & 138    & 0.70                           & 199      & 53     & 0.27                         & 196      & 49     & 0.25                                           \\ 
\hline
E                                                                & 195      & 146    & 0.75                           & 197      & 47     & 0.24                         & 193      & 52     & 0.27                                           \\ 
\hline
F                                                                & 198      & 173    & 0.87                           & 199      & 64     & 0.32                         & 193      & 93     & 0.48                                           \\
\hline

\end{tabular}
}
\arrayrulecolor{black}
\end{table*}


\newrevise{
We use this experiment to check OGMA’s differential testing capability. We present 
the result in Table~\ref{table:rq6-errors}. We run OGMA for 200 iterations and we evaluate 
three error discovery strategies as seen before, i.e., OGMA, random and OGMA with 
backtracking disabled. From Table~\ref{table:rq6-errors}, we observe that the results are 
in line with our findings in IAB classification use cases. Specifically, OGMA discovers 
the most number of errors, averaging an error ratio of 0.81 across the six grammars. 
The random and OGMA- No Backtrack strategies have an error ratio average of 0.29 and 
0.37, respectively.
}

\begin{center}
\begin{tcolorbox}[width=\columnwidth, colback=gray!25,arc=0pt,auto outer arc]
\textbf{$\boldsymbol{\mu}$RQ: Can we use the error inducing inputs generated by 
\FT to improve the accuracy of classifiers? }
\end{tcolorbox}
\end{center}

As part of this research question, we intend to check the usage of error 
inducing inputs generated by \FT. A natural usage of these inputs is to 
retrain the classifier under test. Such a retraining can be accomplished 
by augmenting the training sets with the generated error inducing inputs. 
However, as we only had usage-level access to the text classifiers from 
Rosette, Aylien and uClassify, we were unable to retrain these classifiers. 
Thus, for this research question, we evaluated two classifiers from 
scikit-learn implementations of a regularized linear model with stochastic 
gradient descent and the multinomial Naive Bayes classifier. 
%
%
The objective of these classifiers is to classify a given sentence based 
on which grammar they were generated from. We used the following grammars 
seen in \autoref{fig:toy-gram-1-urq} and \autoref{fig:toy-gram-2-urq} in the evaluation:  
\begin{figure}[h]
\begin{center}
\includegraphics[scale=0.8]{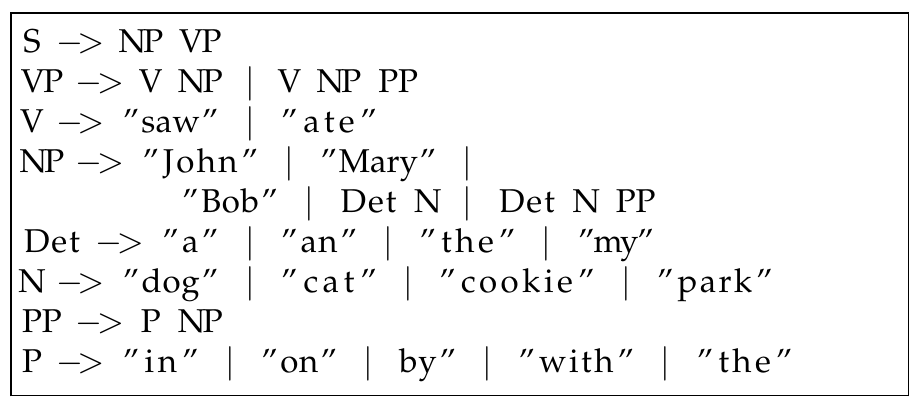}
\end{center}
\caption{Toy Grammar 1}
\label{fig:toy-gram-1-urq}
\end{figure}

\begin{figure}[h]
\begin{center}
\includegraphics[scale=0.8]{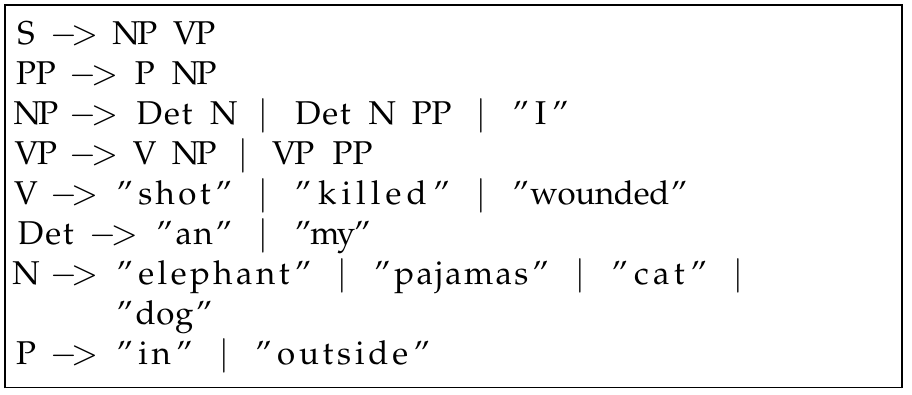}
\end{center}
\caption{Toy Grammar 2}
\label{fig:toy-gram-2-urq}
\end{figure}

Initially, the classification accuracy was 99.75\% for the classifiers. 
Subsequently, we used \FT to generate inputs such that the outputs of 
the chosen two classifiers are different. We add a sample of the 
generated erroneous test inputs into the training set and retrain the  
classifier. To generate the correct labels for the erroneous test inputs, 
we considered one classifier to be the oracle and assign the output 
generated by the oracle as the label. Subsequently, we train the other 
classifier with the augmented training set. \newrevise{It is important to note 
that there may be other ways (e.g. Transduction\cite{learning-transduction}) 
to find the ground truth labels, but investigating such methodologies is beyond 
the scope of this work.}

\newrevise{
Although in certain cases it might be possible to repair the 
machine-learning (ML) model with few representative inputs, we believe 
it is necessary to generate a significant number of erroneous inputs 
(when possible). This is because of two reasons. 
Firstly, the erroneous 
inputs generated by \FT can be used to retrain the underlying ML algorithm 
and reduce its erroneous behaviour. The exact percentage of inputs that 
need to be added may depend on the application and the ML algorithm. 
In our evaluation, for example, augmenting the training set by 15\% with 
the error inducing inputs led to the largest reduction in errors. Secondly, 
even though the repair might be achieved by a few representative error 
inputs, it is crucial to test the repaired model with a substantial number 
of error inducing inputs generated by \FT. This is to check whether the 
repaired model indeed reduced the error rate. In the absence of a 
substantial number of error inducing inputs, the designer will not be able 
to validate a repaired model.
}

We generated 1000 test inputs via \FT to discover the number 
of error inducing inputs before and after retraining. Since \FT has 
randomness involved in its core, we repeated the test generation 50 
times and take the average over all 50 iterations. Our findings are 
summarized in \autoref{table:retrain-errors}. On average, \FT generated 
553 error inducing inputs (out of 1000) before retraining, whereas the 
number of error inducing inputs reduced to 
\newrevise{ as low as 294 (i.e. 47.01\% decrease)} 
after the retraining. This experiment clearly shows that the test 
inputs generated by \FT can be utilized to reduce the erroneous 
behaviours in classifiers. 

%

\begin{table}
\centering
\arrayrulecolor{black}
\caption{Number of error inducing inputs after retraining. The number of 
error inputs added is shown as a percentage of the size of original 
training set.}
\label{table:retrain-errors}
\begin{tabular}{| c | c | c | c |} 
\hline
\% added & \#Errors & \makecell{Accuracy\% \\ SGDClassifier}  & 
\makecell{Accuracy\% \\ Multinomial NB } \\ \hline
0\% 			 & 553 	& 99.76	    &  99.76    \\ \hline
2\%              & 477  &  99.76    &  99.76    \\ \hline
5\%              & 439  &   99.76   &  99.76    \\ \hline
7\%              & 370  &  99.76    &  99.76    \\ \hline
10\%             & 409  &  99.76    &  99.76    \\ \hline
12\%             & 351  &   99.76   &   99.76   \\ \hline
15\%             & 293  &   99.76   &   99.76   \\ \hline
17\%             & 387  &   99.74   &   99.74   \\ \hline
20\%             & 454  &   99.74   &   99.74   \\ \hline
22\%             & 449  &   99.74   &   99.74   \\ \hline
25\%             & 417  &    99.73  &   99.73   \\ \hline
\end{tabular}
\arrayrulecolor{black}
\end{table}


\subsubsection*{\textbf{Examples of Error Inducing Inputs}}
In this section, we introduce some of the interesting error inducing inputs 
automatically discovered by \FT. For instance, consider the following 
sentence generated from one of our subject grammars: 
\begin{align*}
the\ monkey\ shot\ Bob
\end{align*}
The text classifier $\mathbb{U}$ returns the following result (top three categories) 
where the first element in the pair captures the classification class (according IAB 
content Taxonomy Tier 1) and the second element captures the weight (i.e. a score 
reflecting how likely is the respective category): 
\begin{enumerate}
\item 'HOBBIES AND INTERESTS', 0.371043
\item 'SOCIETY', 0.167253
\item 'SPORTS', 0.118665
\end{enumerate}
Another classification for the example sentence
\begin{align*}
I\ shot\ John\ with\ Mary
\end{align*}
leads to the following classification classes: 
\begin{enumerate}
\item 'SOCIETY', 0.840454
\item 'ARTS AND ENTERTAINMENT',  0.159546
\item 'SPORTS', 6.65587$e^{-11}$
\end{enumerate}
As observed from the preceding examples, the computed categories were clearly erroneous. 

We contacted the developers of the service providers of text classifiers and 
pinpointed them to the erroneous inputs. 
Developers confirm that these are indeed erroneous behaviours of 
the classifiers. 
They also confirmed that the primary reason for such erroneous behaviours 
is that the respective classifiers were inadequately trained for the type 
of text inputs generated by \FT. Thus, for these texts, the classifiers 
failed to provide a reasonable classification class. This experience 
clearly indicates the utility of \FT, as the directed test strategy embodied 
within \FT can rapidly discover such erroneous behaviours due to inappropriate 
training. Moreover, as observed in our $\boldsymbol{\mu}$RQ, \FT can also 
augment the training set with the generated erroneous inputs. This, in turn, 
helps to improve the accuracy of classifiers, as observed in our experiments.

 \section{Related Work}
 \label{sec:relatedWork}
 
In this section, we review the related literature and position our 
work on testing machine-learning systems.  

\smallskip\noindent
\textbf{Testing of machine-learning models:}
DeepXplore~\cite{DBLP:conf/sosp/PeiCYJ17} is a whitebox differential testing
algorithm for systematically finding inputs that can trigger inconsistencies
between multiple deep neural networks (DNNs). The neuron coverage was used as a 
systematic metric for measuring how much of the internal logic of a DNNs 
had been tested. More recently, DeepTest~\cite{tian2017deeptest} leverages 
metamorphic relations to identify erroneous behaviors in a DNN. The usage 
of metamorphic relations somewhat solves the limitation of differential 
testing, especially to lift the requirement of having multiple DNNs implementing 
the same functionality. A feature-guided black-box approach is proposed 
recently to validate the safety of deep neural networks~\cite{wicker2017feature}. 
This work uses their proposed method to evaluate the robustness of neural networks 
in safety-critical applications such as traffic sign recognition.
DeepGauge~\cite{deepgauge} formalizes a set of testing criteria based on multi
level and -granularity coverage for testing DNNs and measures the testing quality.
\textsc{Aequitas}\xspace~\cite{aequitas} aims to uncover fairness violations in
machine learning models. DeepConcolic~\cite{deepconcolic} designs a coherent 
framework to perform concolic testing for discovering violations of robustness.
\revise{DeepHunter~\cite{deephunter} and TensorFuzz~\cite{tensorfuzz} propose 
coverage guided fuzzing for Neural Networks.}

\revise{Unlike adversarial text generation~\cite{textbugger}, the goal of \FT is 
completely different. \FT abstracts the input space via a grammar and explores 
the input space with the objective of generating erroneous inputs. 
As a result, the erroneous inputs generated by \FT is not limited 
to only adversarial texts and they do not need to focus on semantic 
similarities. Nevertheless, it is possible for \FT to explore 
semantically equivalent sentences, as long as they conform to the input 
grammar. Indeed, the set of sentences generated by \FT captures 
a variety of texts and they are not restricted to unobservable input 
perturbations. Moreover, \FT guarantees that the input perturbations 
still lead to valid input sentences (according to the grammar). 
Adversarial perturbations, e.g. TextBugger~\cite{textbugger}, might 
not guarantee the conformance with a grammar. Finally, \FT does 
not need any seed input to commence test generation. Thus, in contrast 
to most adversarial testing, \FT can work without seed inputs and 
also for models where the training data is sensitive.}


The aforementioned works are either not applicable for structured 
inputs~\cite{aequitas} or they require a set of concrete seed inputs to initiate 
the test generation process~\cite{DBLP:conf/sosp/PeiCYJ17,tian2017deeptest,wicker2017feature}. 
On the contrary, \FT encodes input domain via grammars and systematically generates 
inputs conforming to the grammar by exploiting the robustness property. Due to the 
grammar-based input generation, \FT can explore an input subspace that could be beyond 
the capability of techniques relying on concrete seed inputs. Presence of an input 
grammar is also common for several machine learning models, especially for models 
in the domain of text classification. Moreover, the objective of the works, as explained 
in the preceding paragraph, is largely to evaluate salient properties, e.g., fairness 
and robustness, of a given machine-learning model. In contrast, our \FT approach is 
targeted to discover classification errors in machine-learning models in a generic 
fashion, while leveraging the robustness property of these well trained models. 


\smallskip\noindent
\textbf{Verification of Machine Learning models:}
AI\textsuperscript{2}~\cite{ai2} uses abstract interpretation to verify the 
robustness of a given input against adversarial perturbations.
AI\textsuperscript{2} leverages zonotopes to approximate ReLU outputs. 
The authors guarantee soundness, but not precision. ReluVal~\cite{ReluVal} uses
interval arithmetic~\cite{interval-analysis} to estimate a neural network's
decision boundary by computing tight bounds on the output of a network for a given
input range. The authors leverage this to verify security properties of a Deep
Neural Network. Similarly, Reluplex~\cite{reluplex} uses SMT solvers to verify 
these security properties. They present an SMT solver and encode properties of
interest into this SMT solver. Dvijotham et al.~\cite{dual-verification} 
transform the verification problem into an unconstrained dual formulation 
using Lagrange relaxation and use gradient-descent to solve the respective 
optimization problem. 

In contrast to these works, our \FT approach has the flavor of testing. 
Specifically, our \FT approach does not generate false positives, i.e., all witnesses 
generated by \FT indeed capture erroneous behaviours in test classifier(s). 
Moreover, these witnesses generated by \FT can be used to retrain the test 
classifiers and thus reducing the number of erroneous classifications.

\smallskip\noindent
\textbf{Search based testing:}
Search-based testing has a long standing history in the domain of software engineering. 
Common techniques for search-based software testing are hill climbing, simulated annealing and genetic
algorithms~\cite{search_based_testing_survey}. 
These have been applied extensively to test applications that largely fall in the 
class of deterministic software systems. With this work we aim to uncover ways
to adapt these techniques to statistical software in general. 

\smallskip\noindent
\textbf{Choice of grammar-based equivalence:}
\revise{Grammar-based testing is applicable to a wide-range of real-world software, 
as observed in several existing works in the software engineering research 
community~\cite{grammarfuzz,langfuzz}. These works, however, target traditional 
software (i.e. not ML-based applications). The objective of our work is a 
novel grammar-based testing that exploits the intrinsic properties in 
machine-learning systems. For several real-world software (e.g. malware 
detectors for Javascript), the grammars are already available. Moreover, 
for several real-world systems, existing works show that such grammars 
can be constructed with little manual effort~\cite{ase16-fuzz} or they 
can even be mined automatically~\cite{ase16-grammarfuzz}. Thus, we believe 
that it is justifiable to rely on the presence of a grammar (encoding the 
input space) or to construct them with reasonable manual effort. In our 
evaluation, we can easily construct several grammars according to a template 
and they facilitate in discovering numerous errors in the NLP classifiers. 
In the future, such a grammar can be mined automatically, yet we believe 
that is orthogonal to the objective of our paper.}


 \section{Threats to Validity}
 \label{sec:threatsToValidity}
 
\smallskip\noindent
\textbf{Choice of Grammar:}
\FT implements a perturbation algorithm which perturbs the structure of the 
derivation tree of an input. The key requirement of \FT is that there should 
be many inputs which have the same structure for their derivation trees. 
This is not possible with grammars that have only one terminal symbols in 
their production rules. A derivation tree constructed from such a grammar 
will not be perturbed by \FT, and would lead to very restricted testing. 
However, the rationale behind perturbation in \FT is to exploit the robustness 
property in machine-learning models for scalable testing.
Specifically, we postulated that inputs having similar derivation tree 
structure are likely to be classified similarly and our empirical results 
validated this.

\smallskip\noindent
\textbf{Robustness:}
\FT is based on the hypothesis that 
the machine-learning models under test exhibit robustness.  
This is a reasonable assumption, as we expect the models under test 
to be deployed in real-world settings. As evidenced by our evaluation, 
\FT approach, which is based on the aforementioned hypothesis, was 
effective to localize the search in the neighbourhood of regions 
exhibiting erroneous behaviours.


\smallskip\noindent
\textbf{Complex Inputs:}
%
\newrevise{
Currently, \FT only works on input domain encoded by context free grammars. 
This includes 
natural language processing tools (as evaluated in our work) and 
malware detectors targeting certain programming and scripting 
languages~\cite{malware-http,malware-powershell}, among others.
The grammar helps 
us to encode a large number of inputs and explore the input space beyond the 
training set in a systematic fashion. In our evaluation, we can easily 
construct several grammars according to a template and they facilitate in 
discovering numerous errors in the NLP classifiers.}

\newrevise{
\FT is not evaluated on more complex input structures such as  images and 
 videos. To adapt our \FT approach for such complex inputs, a model that 
encodes these inputs is needed. This can be accomplished in a future 
extension of \FT. }

\smallskip\noindent
\textbf{Size of Input Text:}
We have tested classifiers that are claimed to not perform well for short text. 
It was brought to our notice that the classifier models need more 
context for the task of classification. We cannot conclude the effectiveness 
of \FT for longer texts. However, the open architecture of \FT allows for 
extensive evaluation of grammars generating longer texts. 

\smallskip\noindent
\textbf{Incompleteness:}
\revise{
In our evaluation, we have tested \FT for only up to 2000 iterations. It is possible
that we have not captured all the test cases which induce errors. By design \FT 
is not complete in terms of generating erroneous inputs.}

\section{Conclusion}
\label{sec:conclusion}

In this paper, we present \FT, a fully automated technique to generate grammar-based 
inputs which exhibit erroneous behaviours in \revise{machine learning based 
natural language processing models}. At the core of 
\FT lies a novel directed search technique. The key insight behind \FT is to 
exploit the robustness property inherent in any well trained machine-learning 
model. \FT provides comprehensive empirical proof for errors in text classifiers. 
To the best of our knowledge, \FT is the only grammar-based machine learning 
testing solution to date. We provide a generic and modular framework to any user 
of our tool to extend the application of \FT beyond classifiers. 
We also try and retrain a toy classifier models to show the potential use cases of 
these discovered erroneous behaviours. 

\newrevise{
\FT directs the search process in the input space to maximise the 
number of errors found. These errors may not necessarily be due to the same 
defect of the model. Thus, we believe that \FT is a powerful tool to discover 
erroneous inputs that may be caused due to a variety of defects embodied 
within the model. In other words, \FT should be used as a testing tool for 
NLP models to discover errors. In its current state, 
\FT is not capable to pin down the root cause in the model for a given erroneous 
input. This requires further development in the fault localisation research.} 
In future, we plan to extend the capability 
of \FT to automatically localise the cause of errors discovered in these
machine learning based natural language processing models. \newrevise{It would
also be desirable to integrate \FT with a system that can determine the ground 
truth label of the discovered inputs (e.g. Transduction
\cite{learning-transduction}) to effectively 
retrain the classifiers to alleviate the errors we have discovered.}

\FT lifts the state of the art by introducing a novel approach to testing for 
machine-learning models. We envision to extend \FT beyond just text classifier 
testing and we hope it can be used to test any machine-learning model whose 
input domain can be formalised not only via grammars, but also via other techniques 
such as via leveraging logic based on satisfiability modulo theory (SMT). We 
would also like to extend \FT to video and image inputs.  We hope that the central 
idea behind our \FT approach would influence the rigorous software engineering 
principles and help validate machine-learning applications. 
For reproducibility and advancing the state of research, we have made our tool and 
all experimental data publicly available:
\begin{center}
\url{https://github.com/sakshiudeshi/Ogma-Data}
\url{https://github.com/sakshiudeshi/Ogma}
\end{center}

\IEEEpeerreviewmaketitle

\balance
\bibliographystyle{plainurl}
\bibliography{GMLFuzz}

\newpage

\appendices
\section{Grammars and Additional Graphs}
\label{sec:appendix-grammars}

In the appendix below we provide all the grammars that we used for testing 
and the additional experimental results that we obtained.


\begin{figure}[H]
\begin{center}
\includegraphics[scale=0.8]{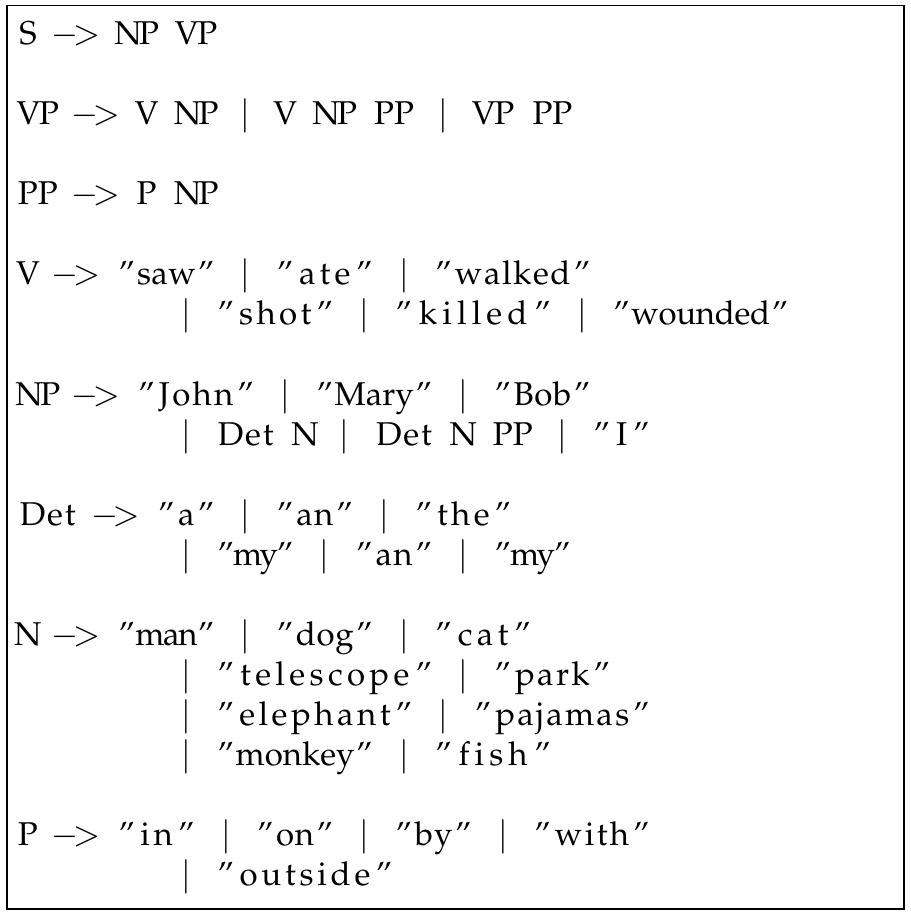}
\end{center}
\caption{Grammar A}
\label{fig:grammarA}
\end{figure}

\begin{figure}[H]
\begin{center}
\includegraphics[scale=0.8]{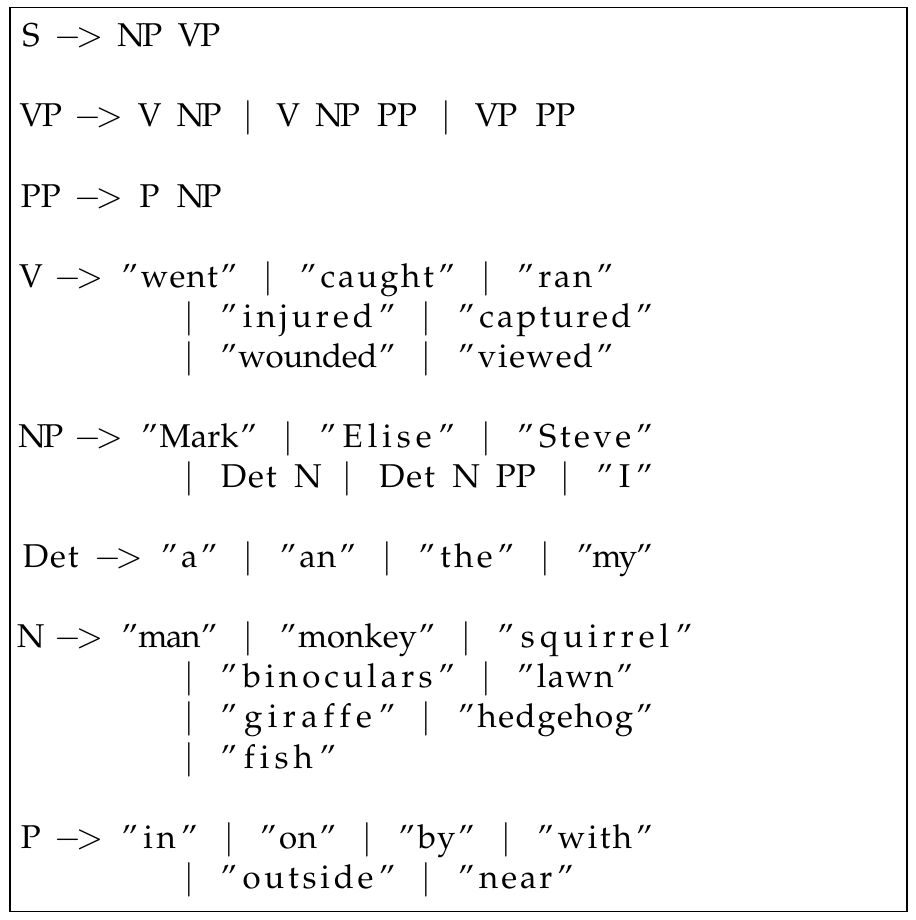}
\end{center}
\caption{Grammar B}
\label{fig:grammarB}
\end{figure}

\begin{figure}[H]
\begin{center}
\includegraphics[scale=0.8]{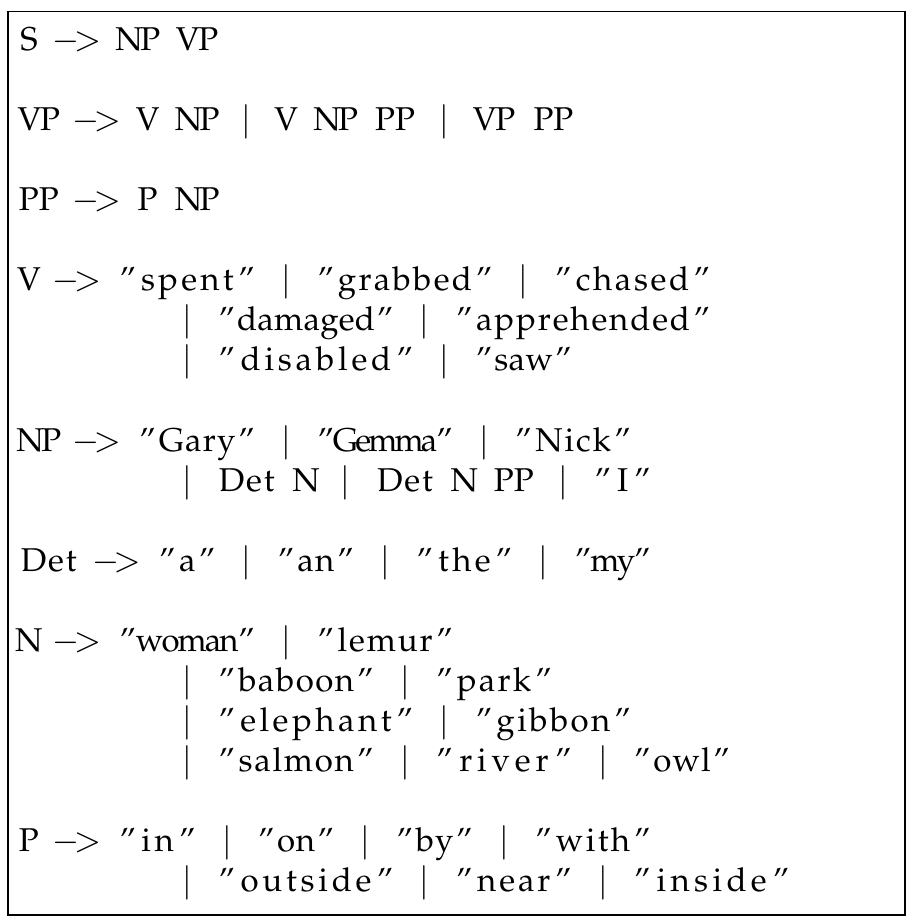}
\end{center}
\caption{Grammar C}
\label{fig:grammarC}
\end{figure}

\begin{figure}[H]
\begin{center}
\includegraphics[scale=0.8]{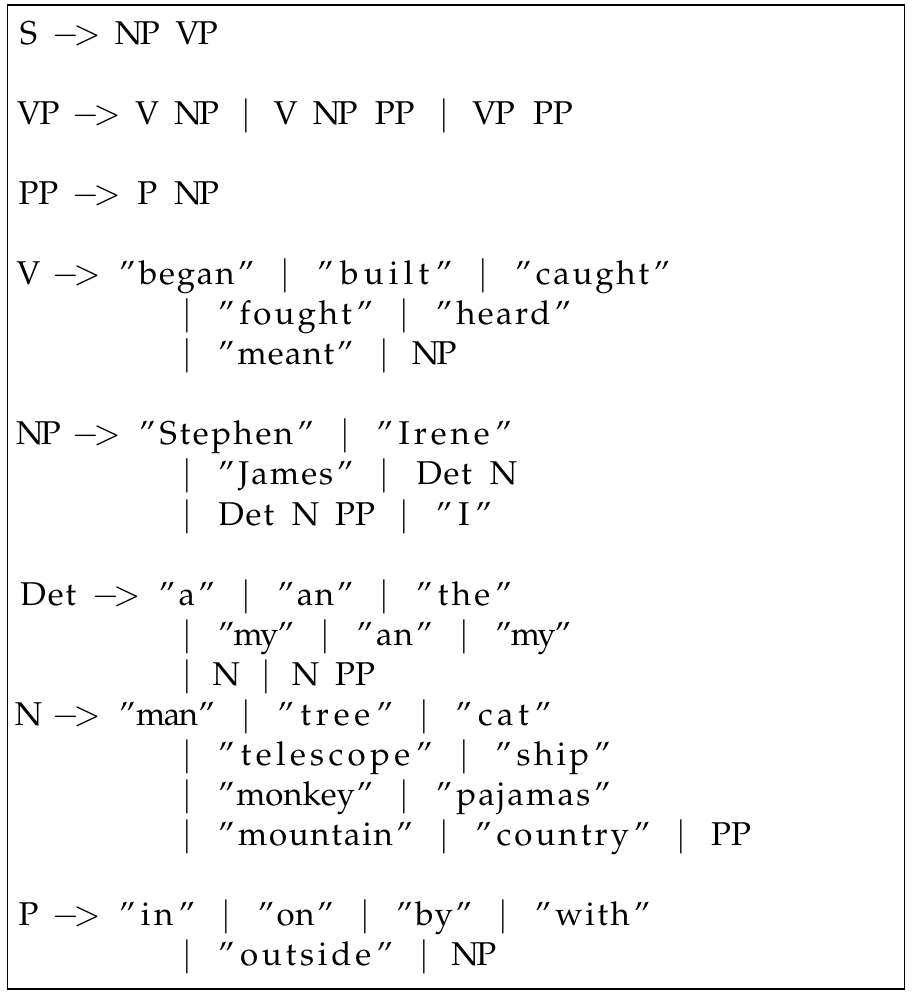}
\end{center}
\caption{Grammar D}
\label{fig:grammarD}
\end{figure}

\newpage

\begin{figure}[H]
\begin{center}
\includegraphics[scale=0.8]{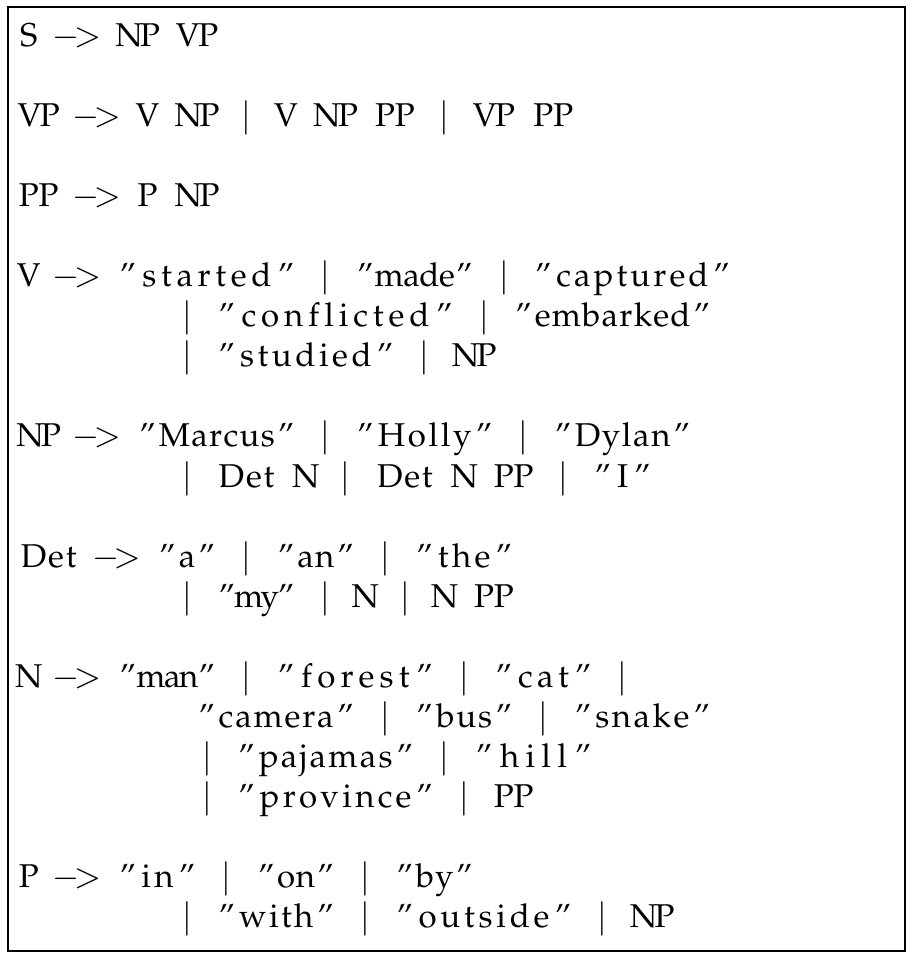}
\end{center}
\caption{Grammar E}
\label{fig:grammarE}
\end{figure}

\begin{figure}[H]
\begin{center}
\includegraphics[scale=0.8]{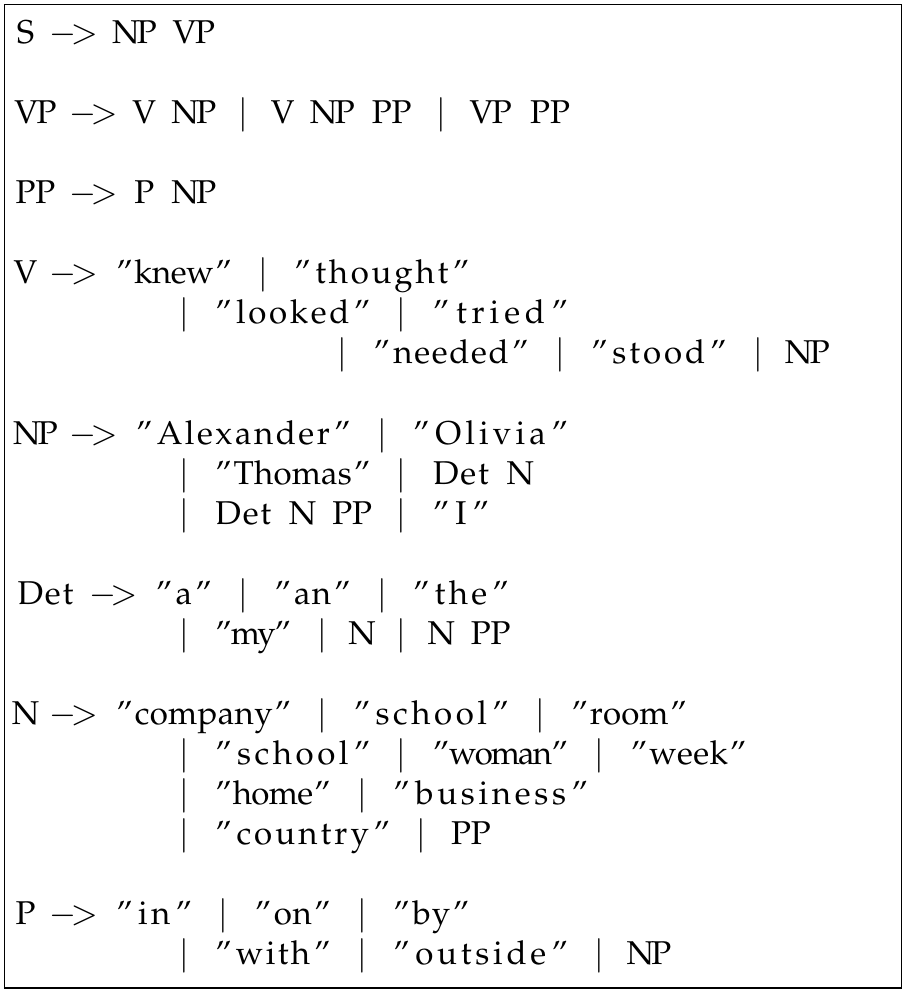}
\end{center}
\caption{Grammar F}
\label{fig:grammarF}
\end{figure}



\begin{figure*}[t]
\begin{center}
\begin{tabular}{ccc}
\includegraphics[scale=0.3]{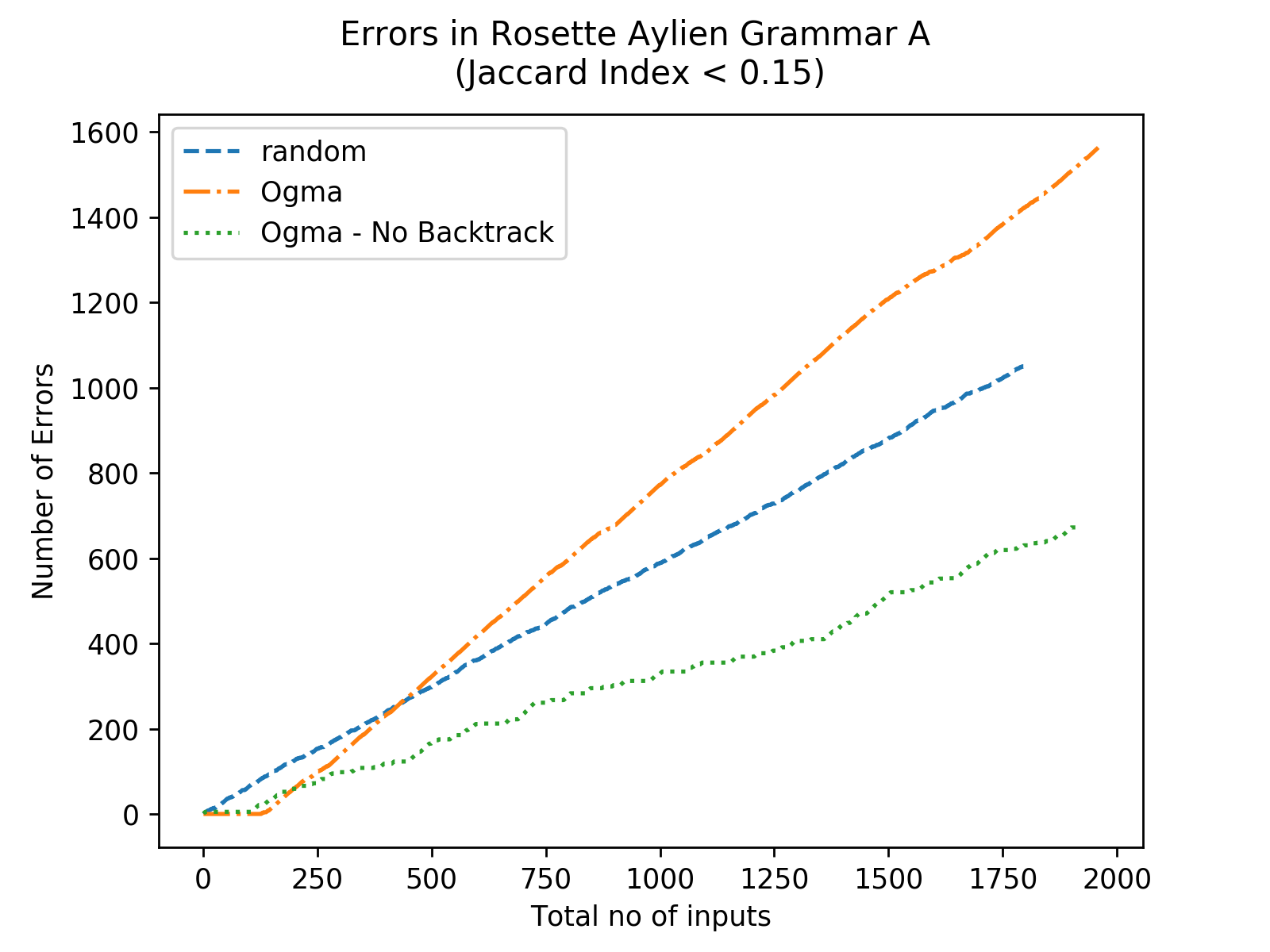} & 
\includegraphics[scale=0.3]{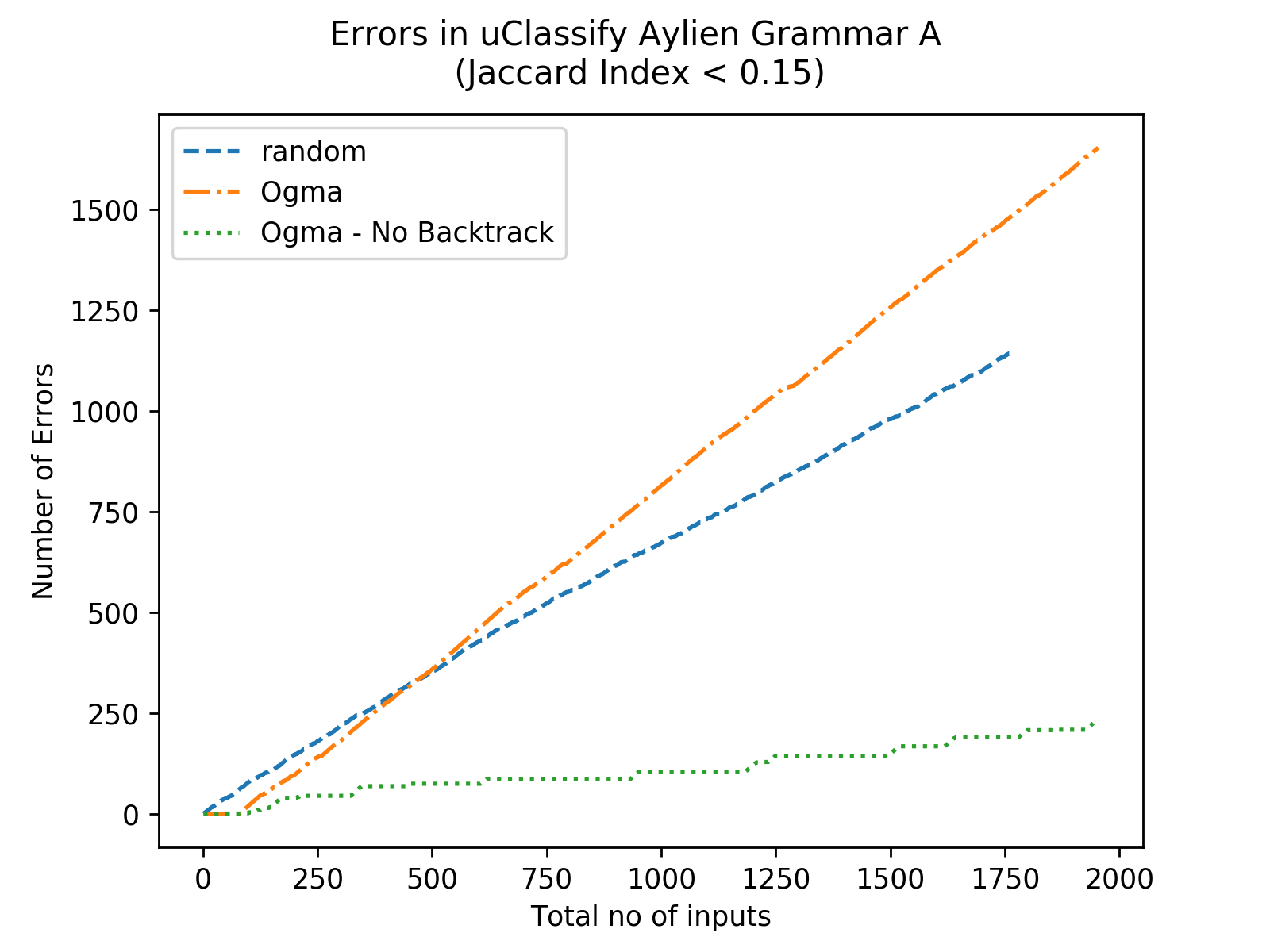} & 
\includegraphics[scale=0.3]{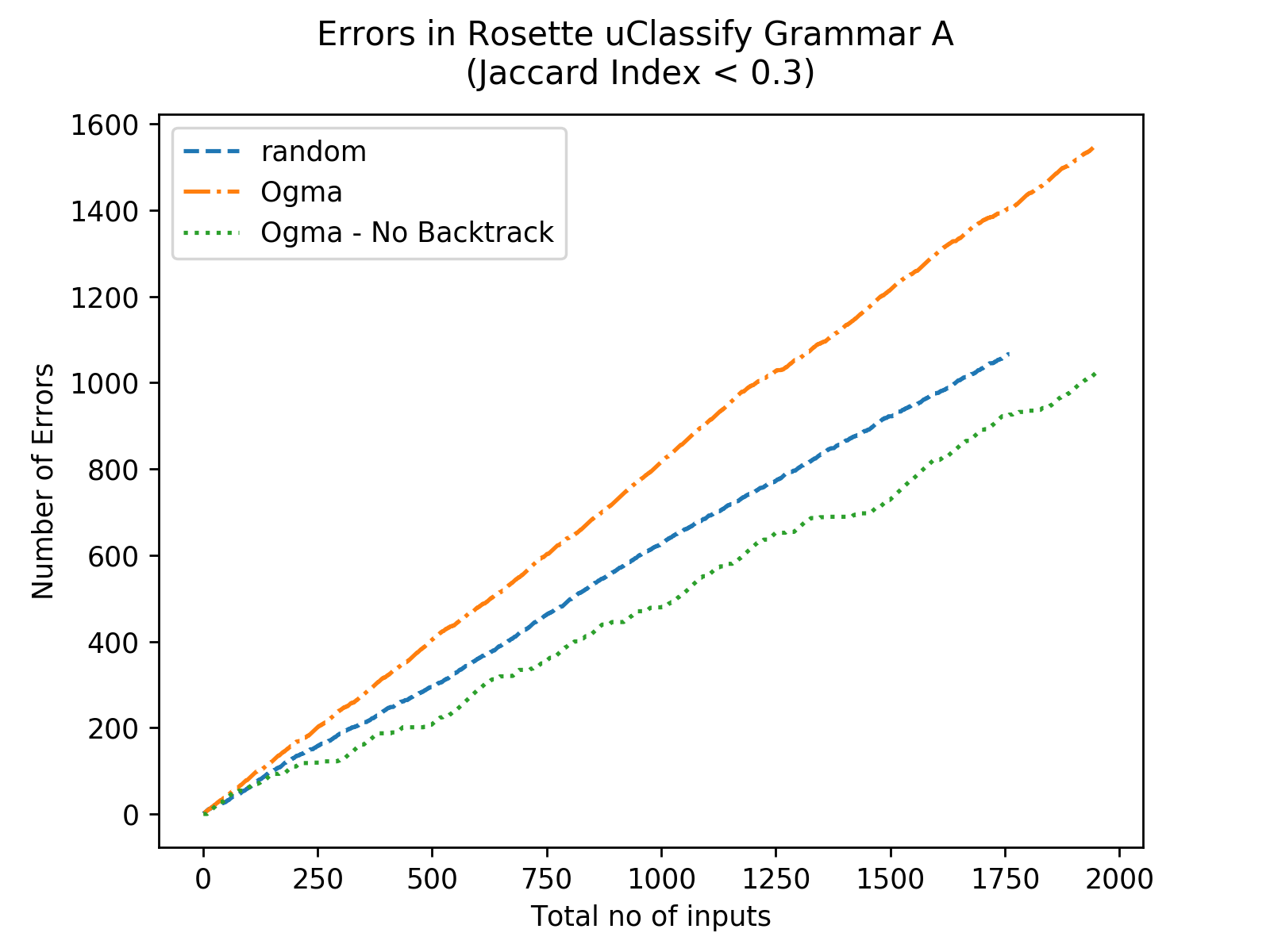}\\

\\
\includegraphics[scale=0.3]{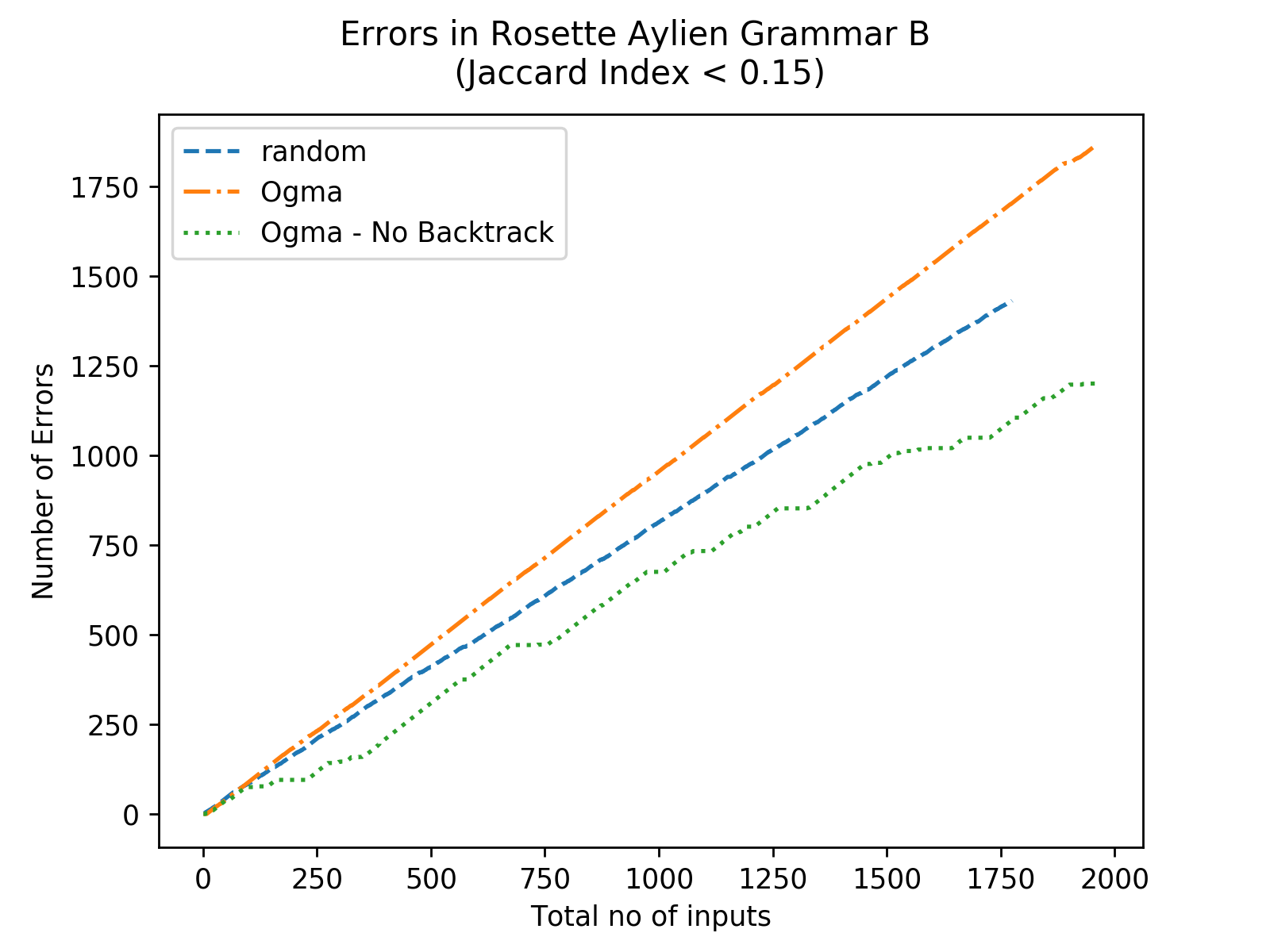} & 
\includegraphics[scale=0.3]{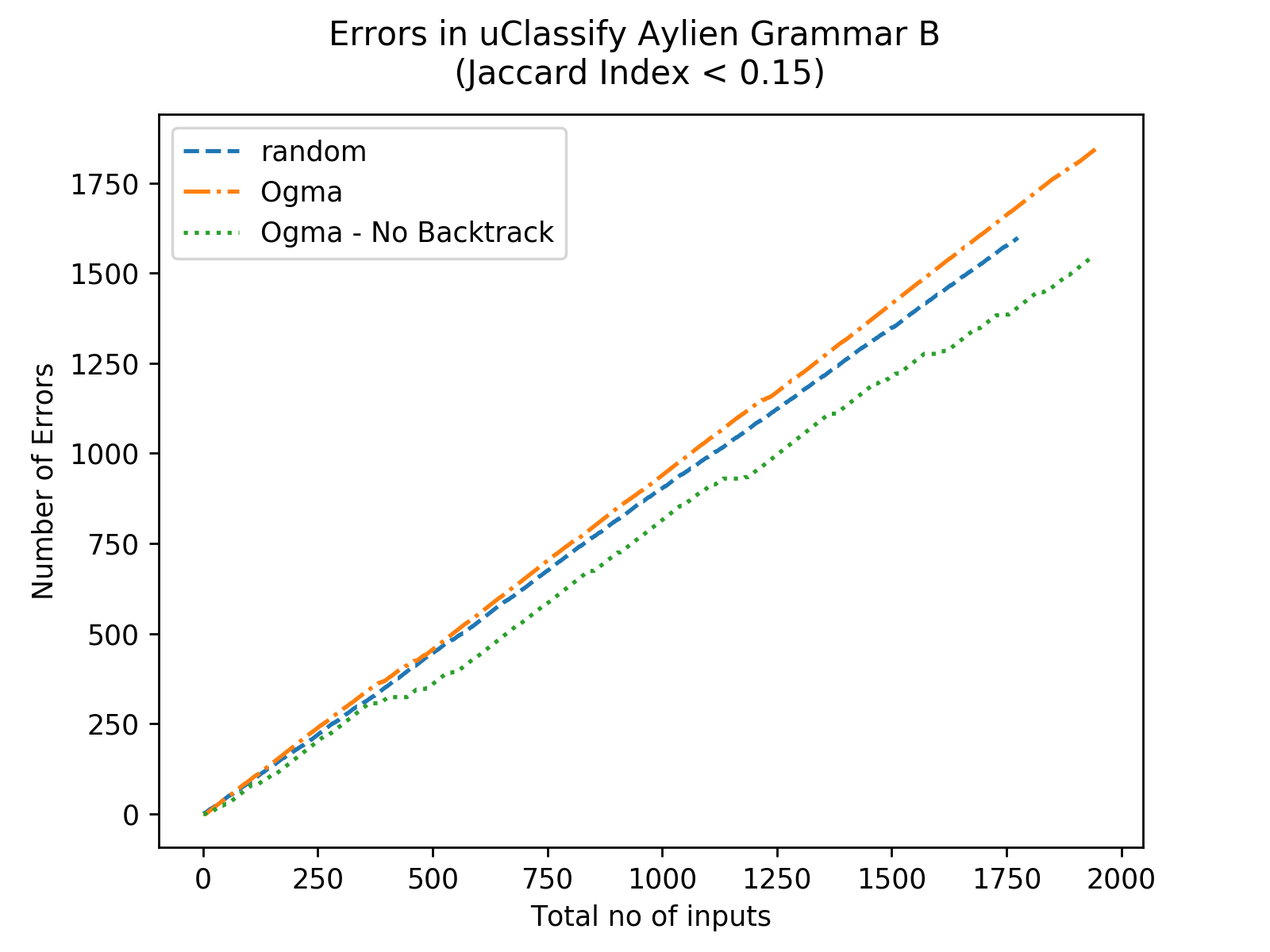} & 
\includegraphics[scale=0.3]{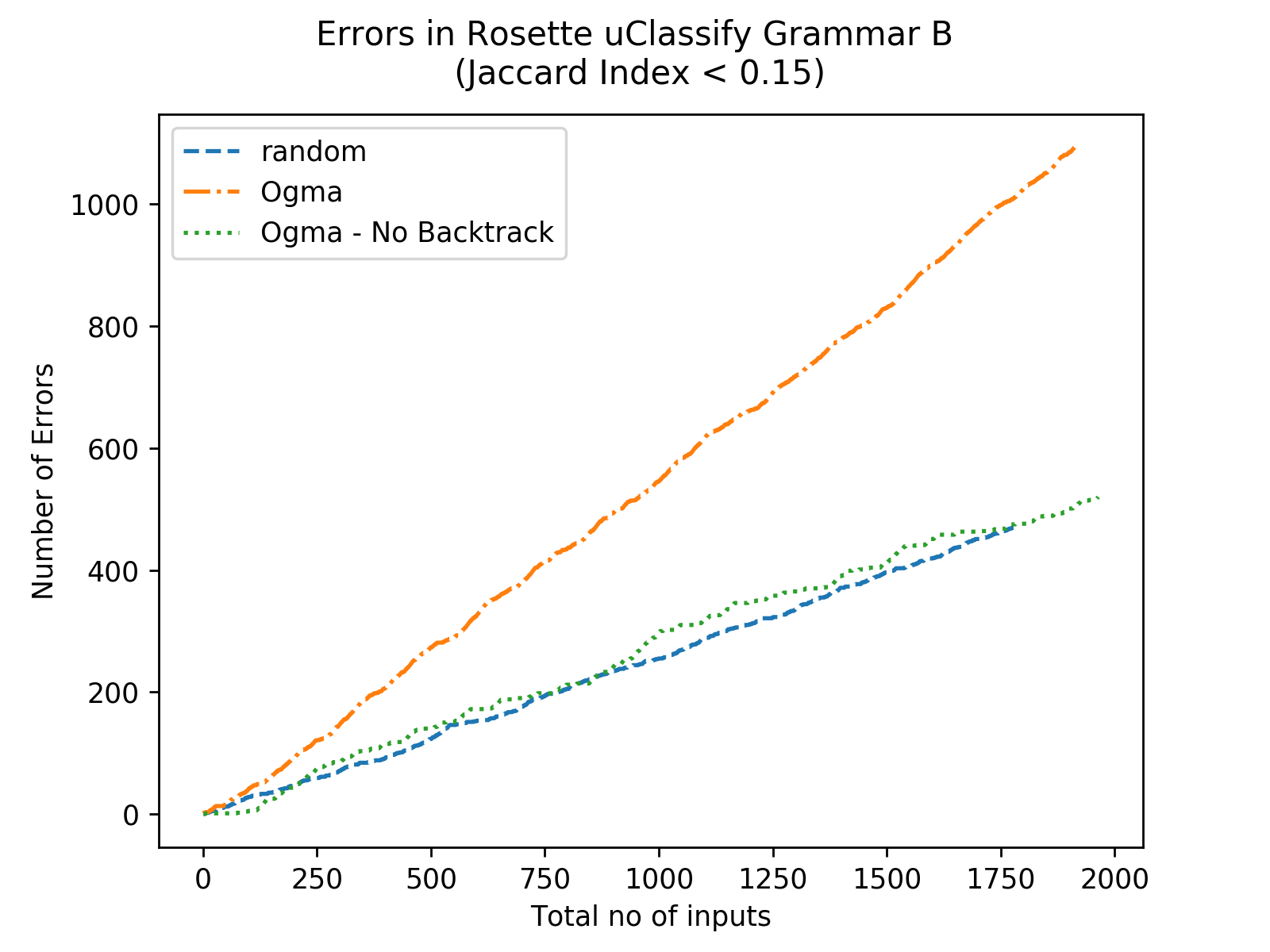}\\

\\
\includegraphics[scale=0.3]{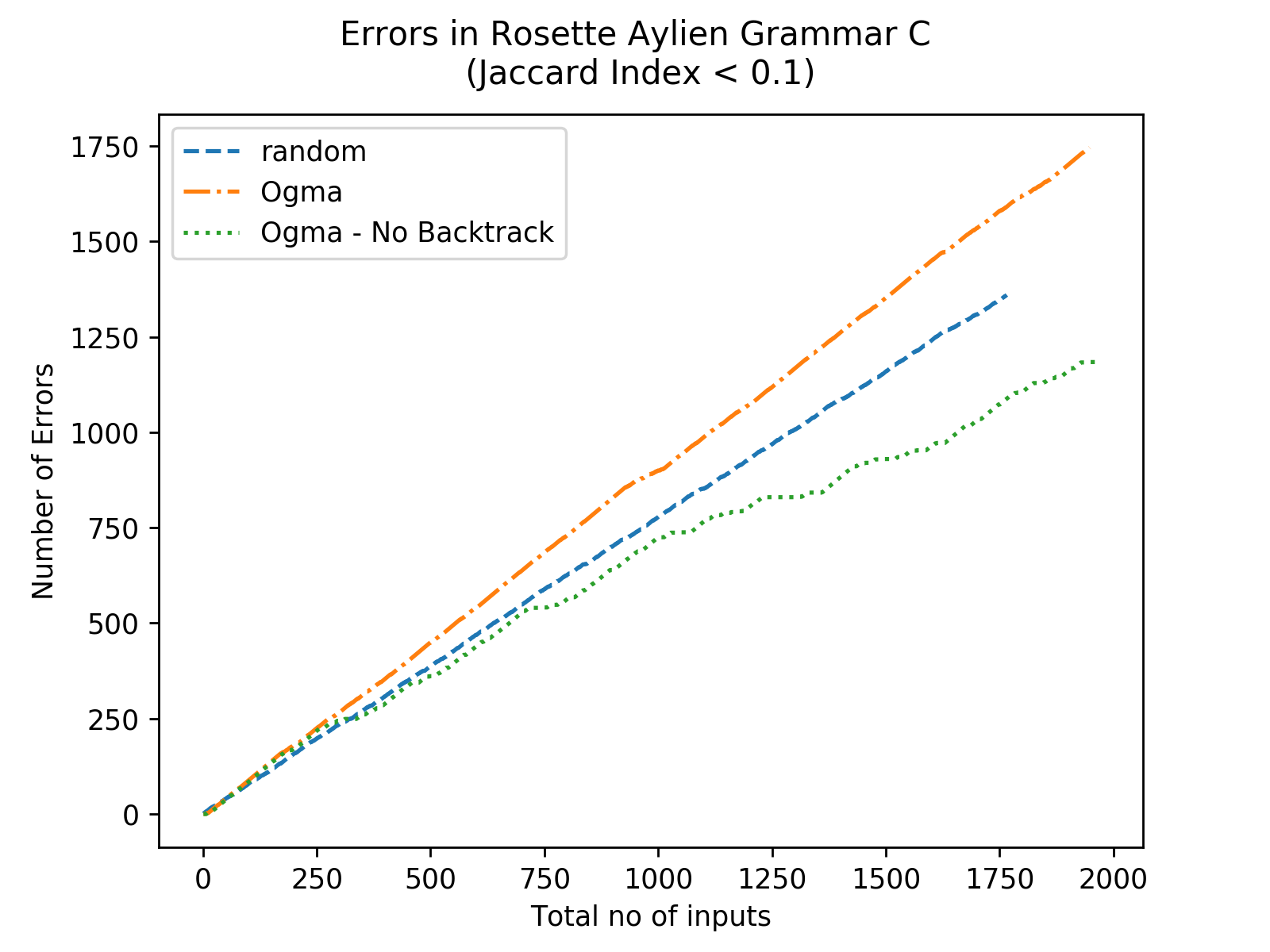} & 
\includegraphics[scale=0.3]{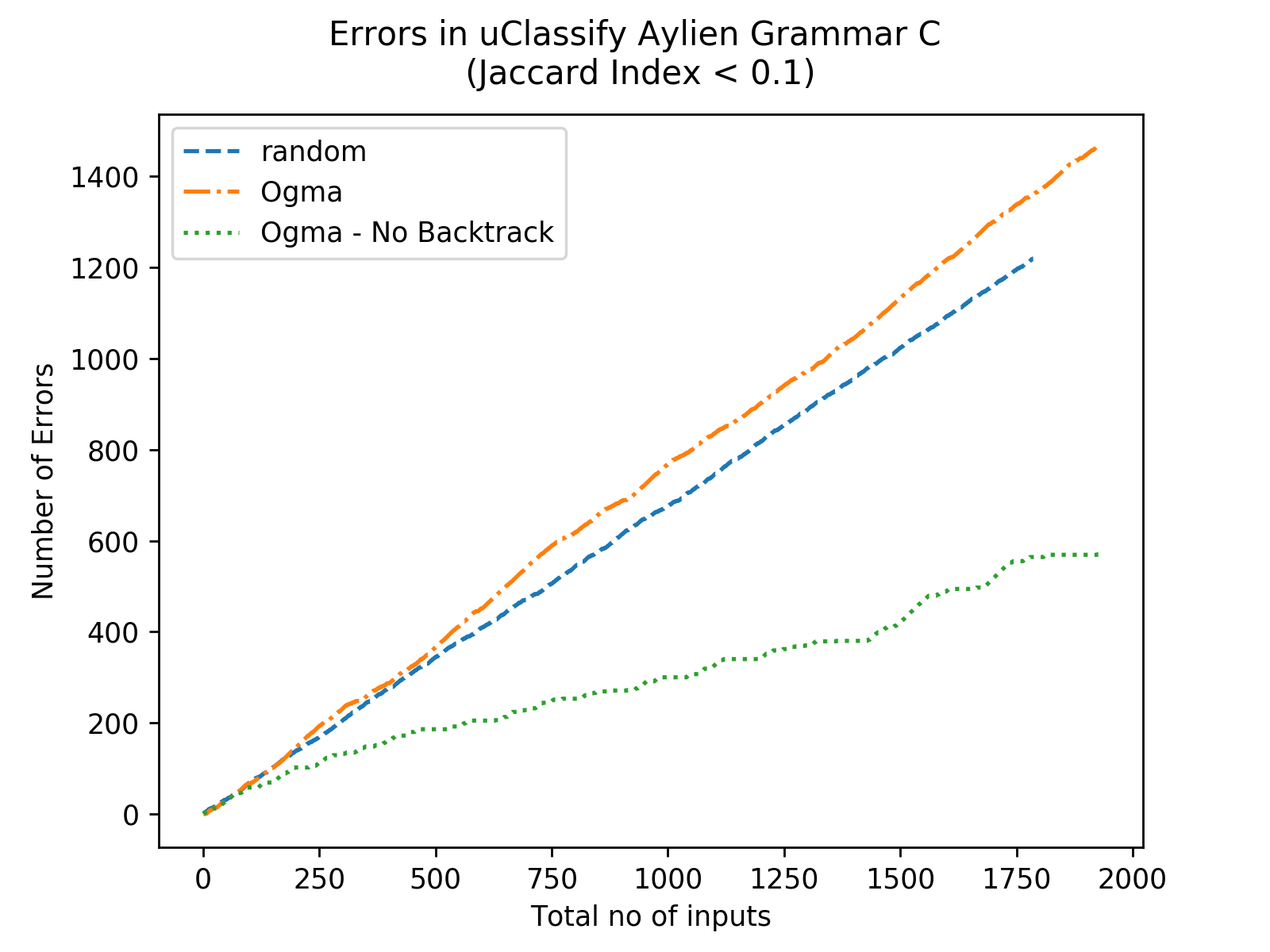} & 
\includegraphics[scale=0.3]{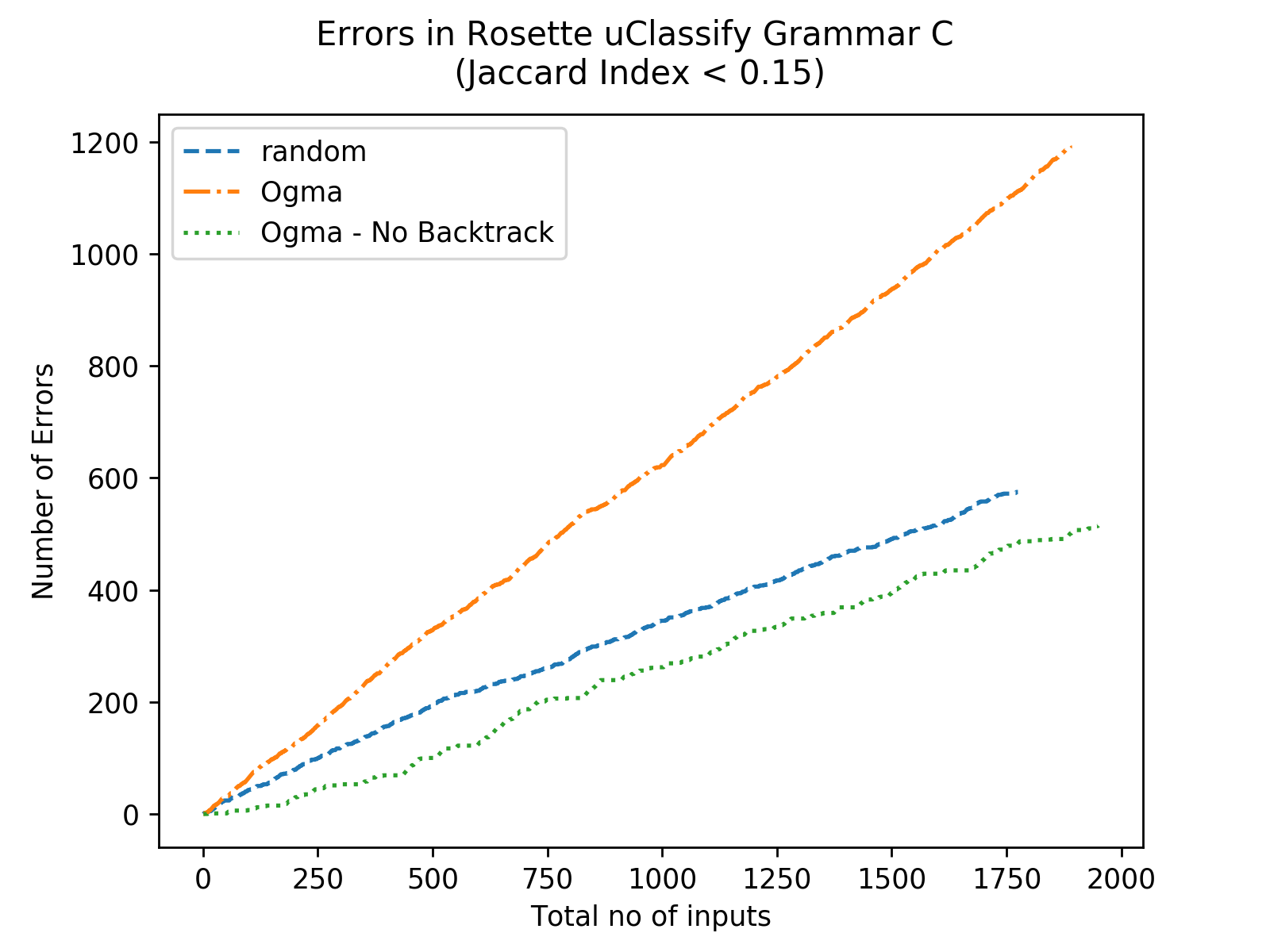}\\

\\
\includegraphics[scale=0.3]{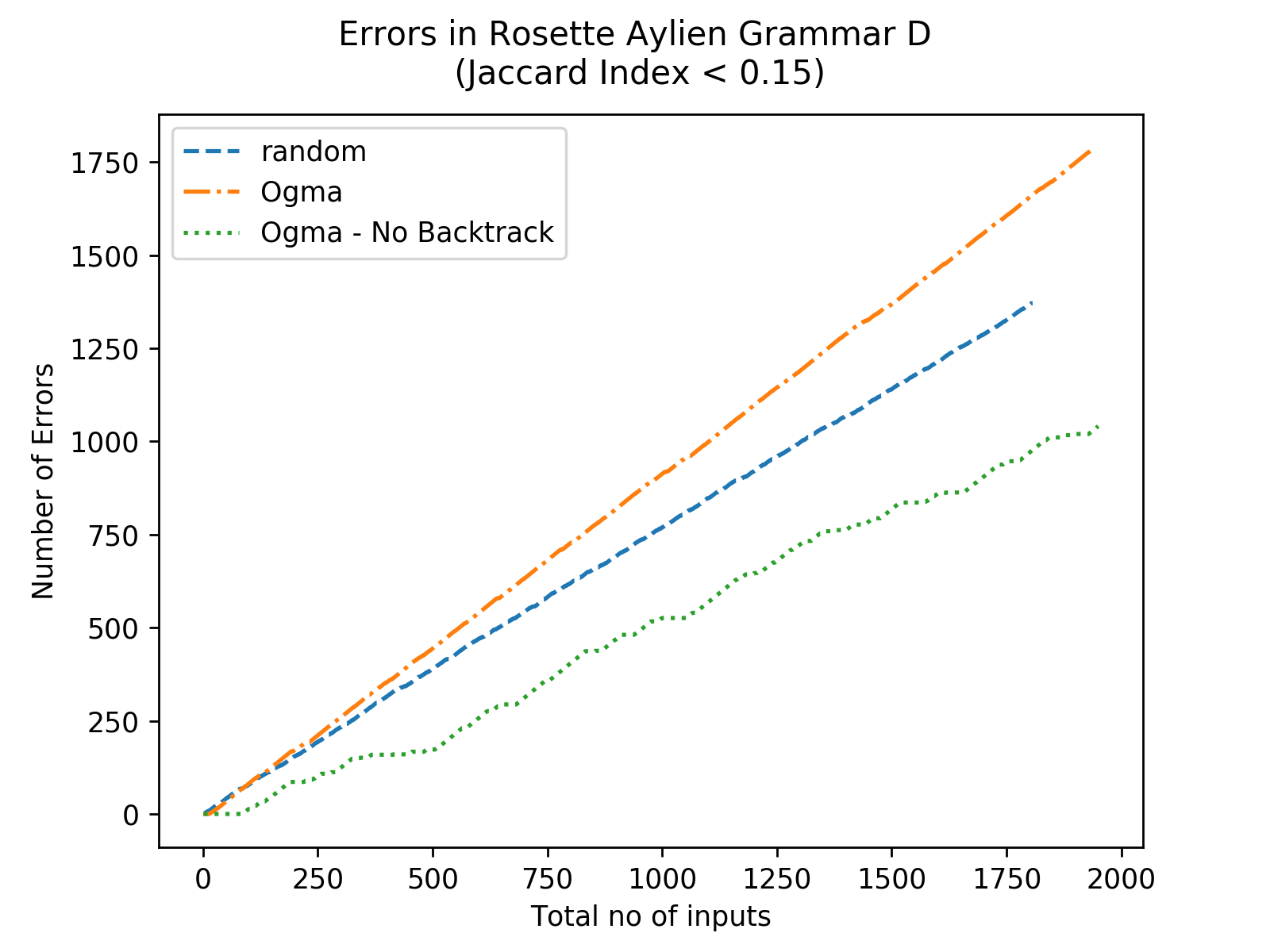} & 
\includegraphics[scale=0.3]{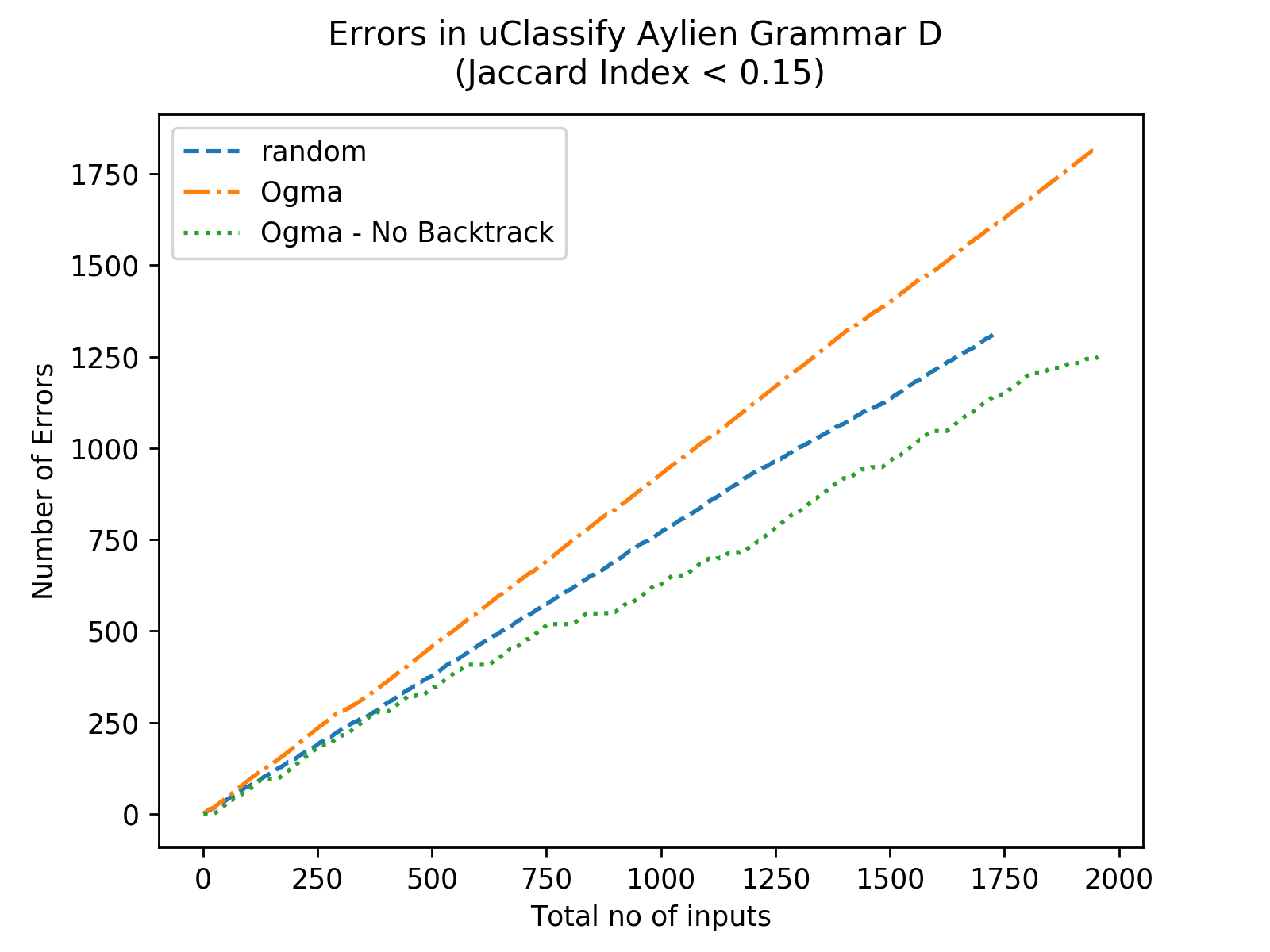} & 
\includegraphics[scale=0.3]{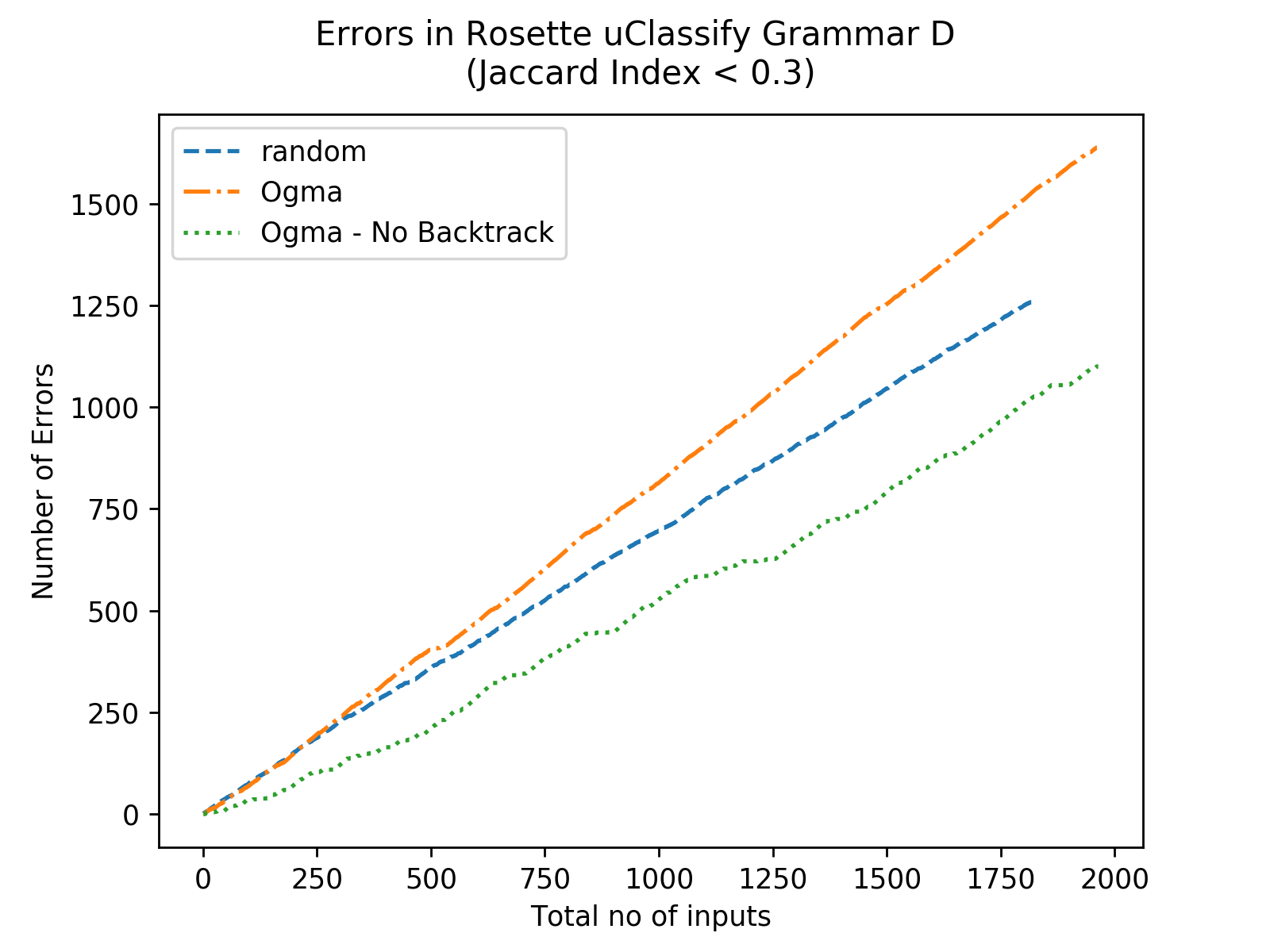}\\

\\
\includegraphics[scale=0.3]{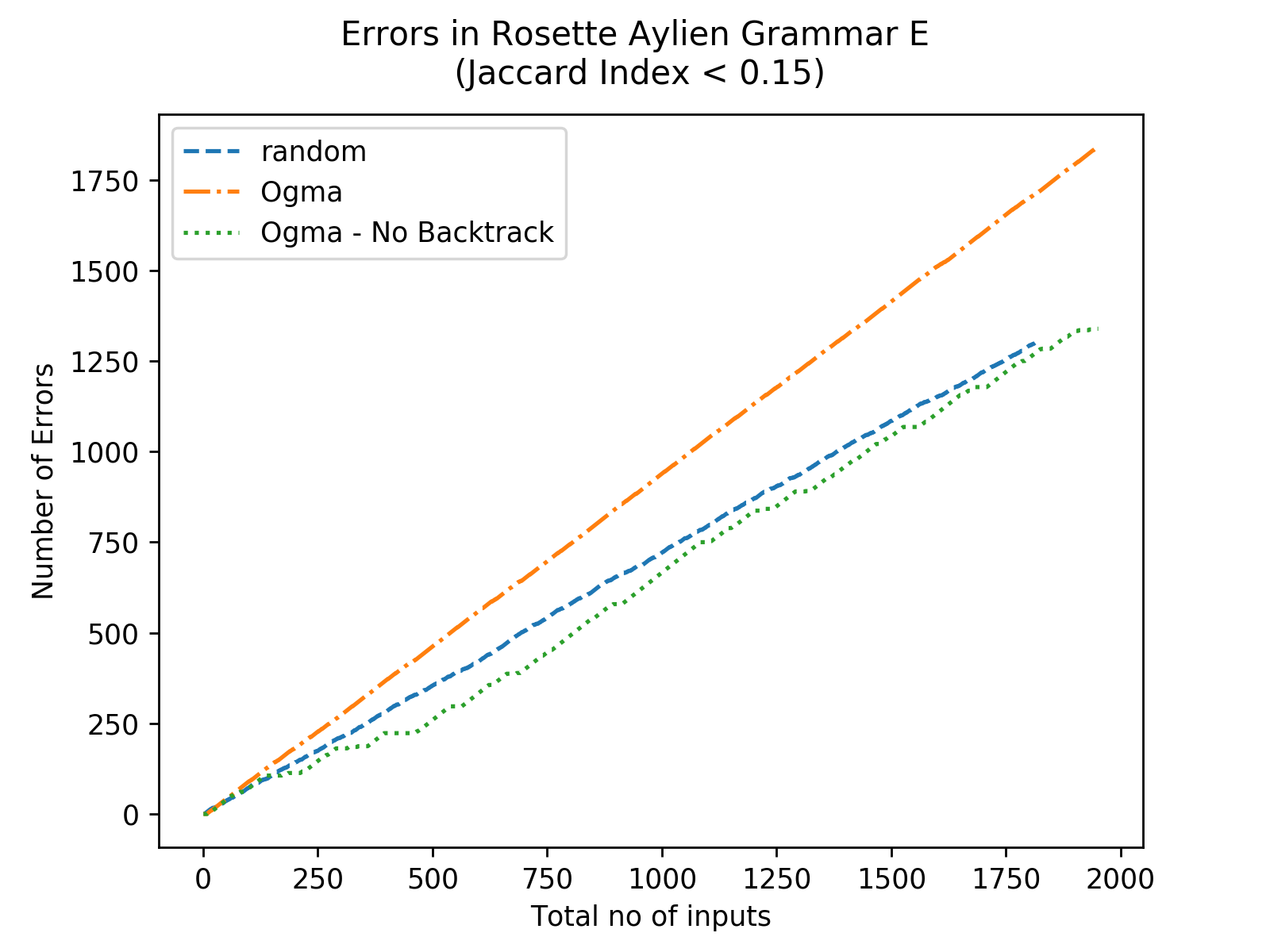} & 
\includegraphics[scale=0.3]{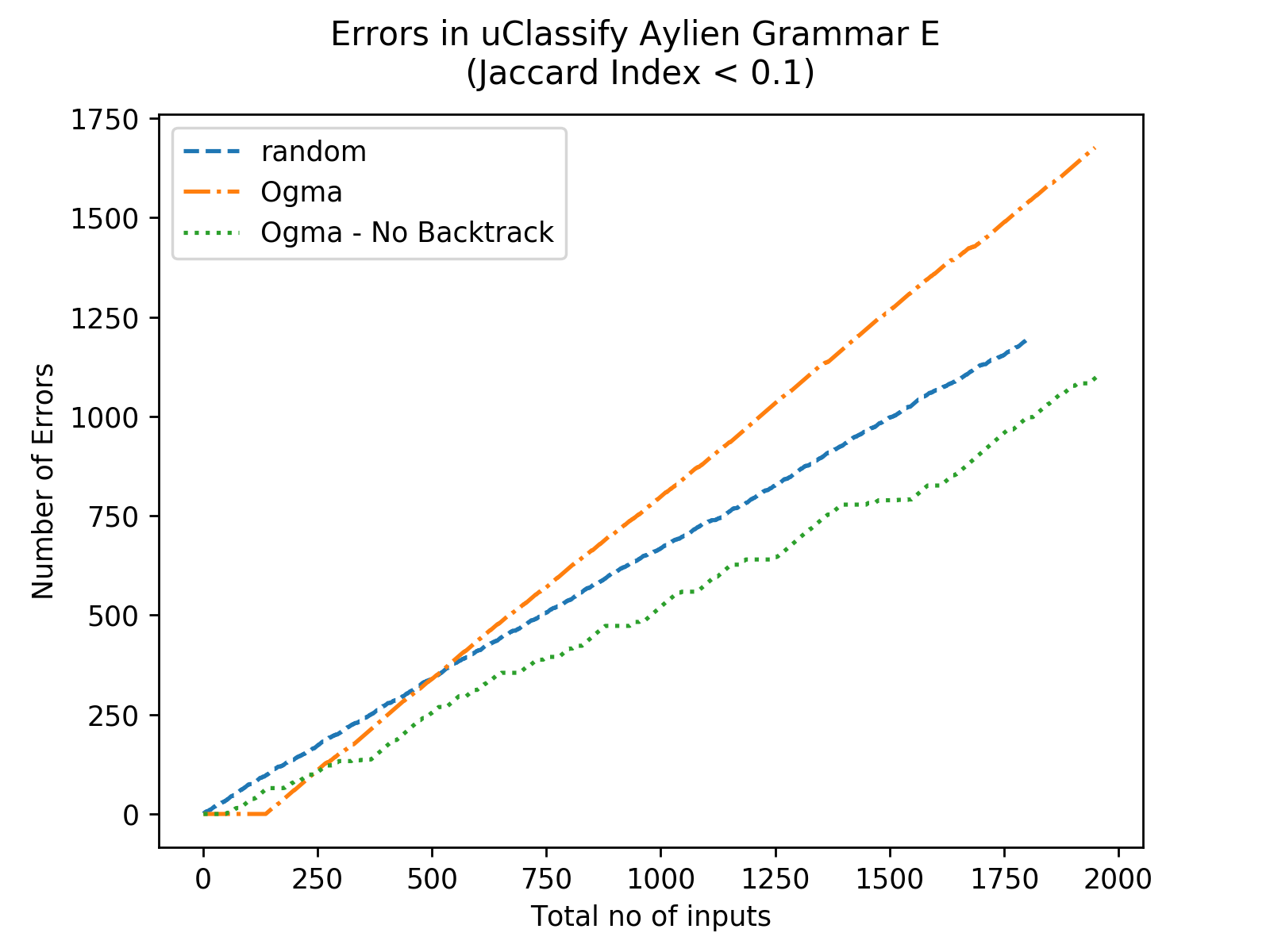} & 
\includegraphics[scale=0.3]{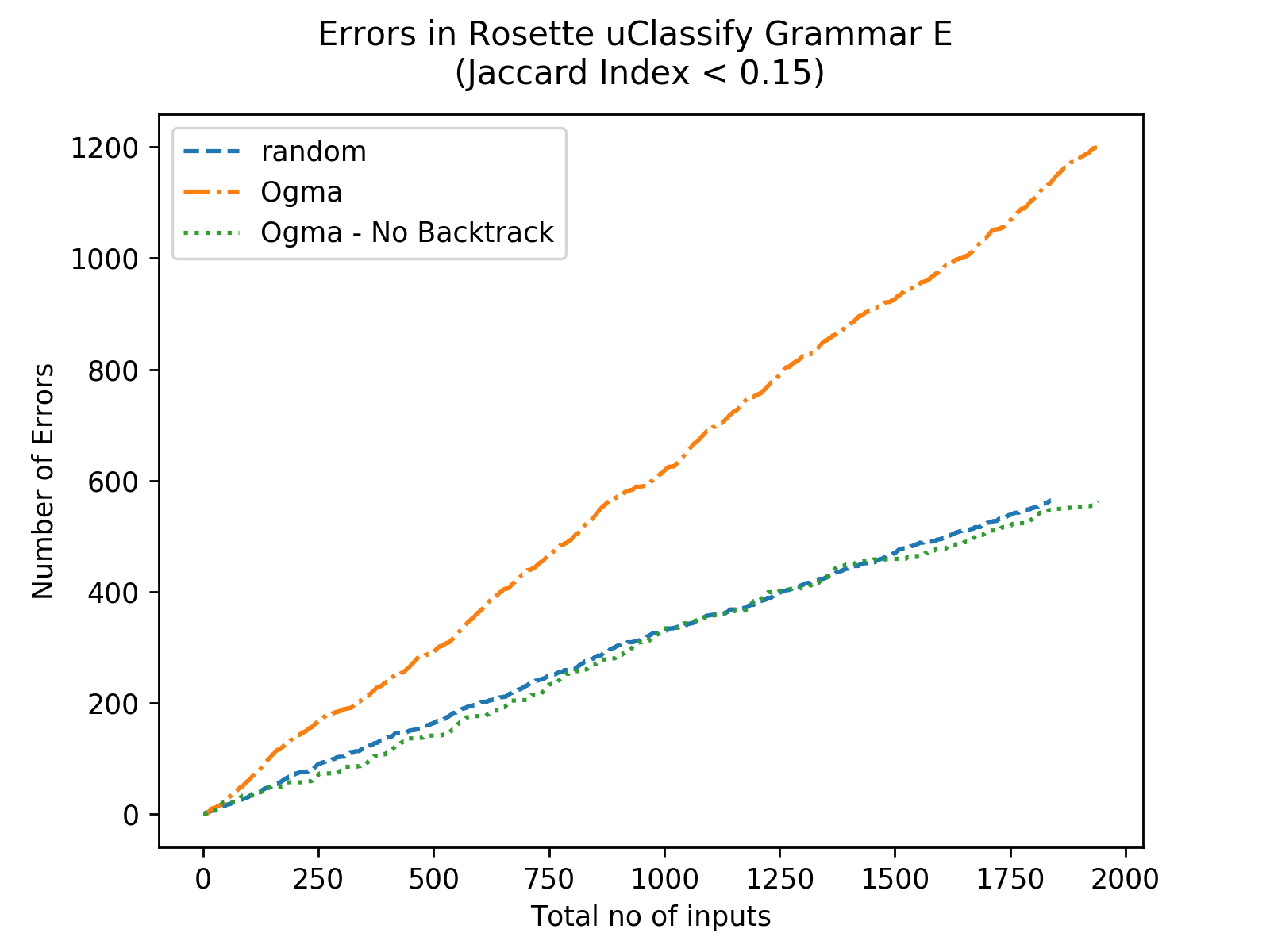}\\

\\
\includegraphics[scale=0.3]{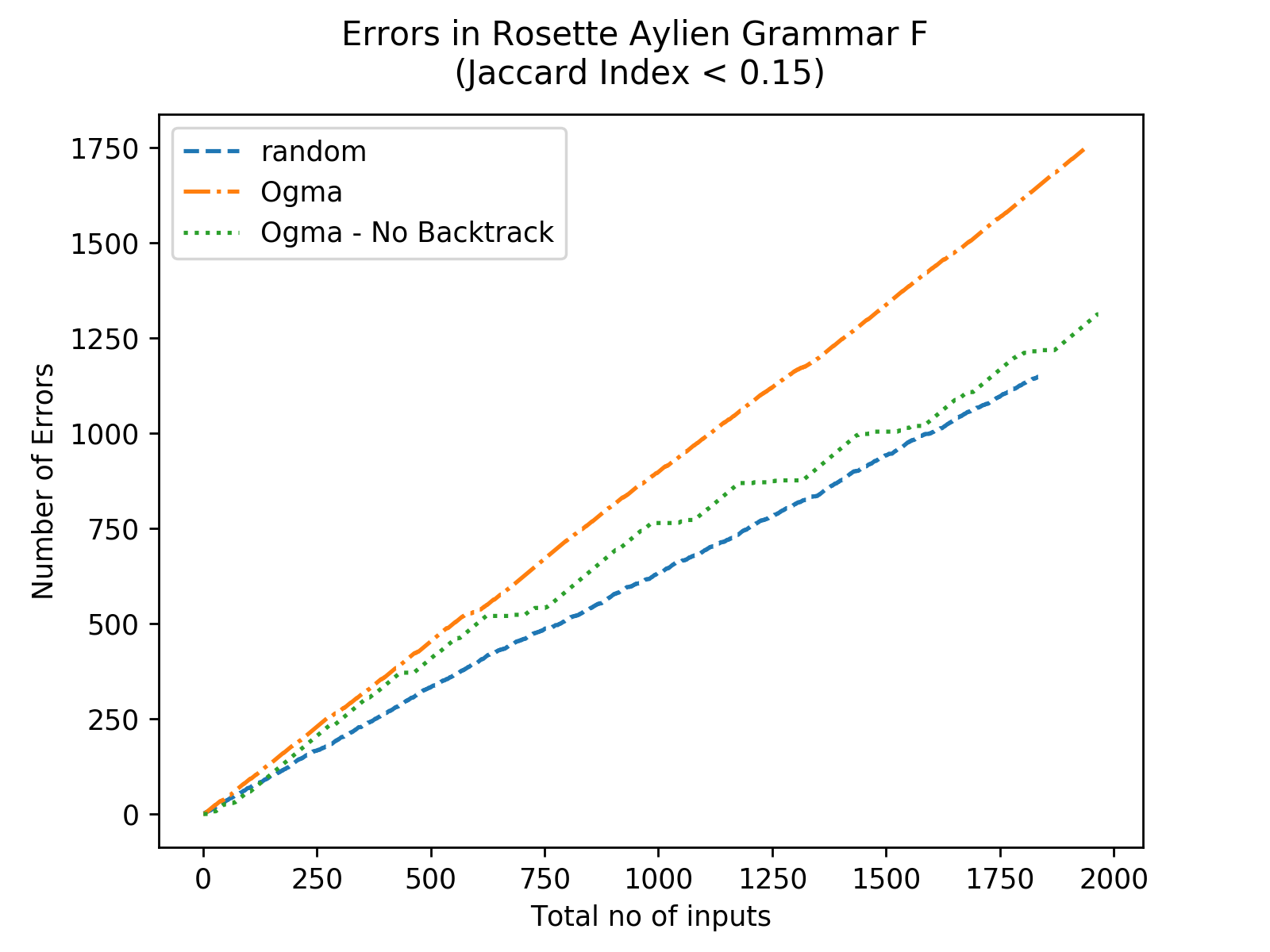} & 
\includegraphics[scale=0.3]{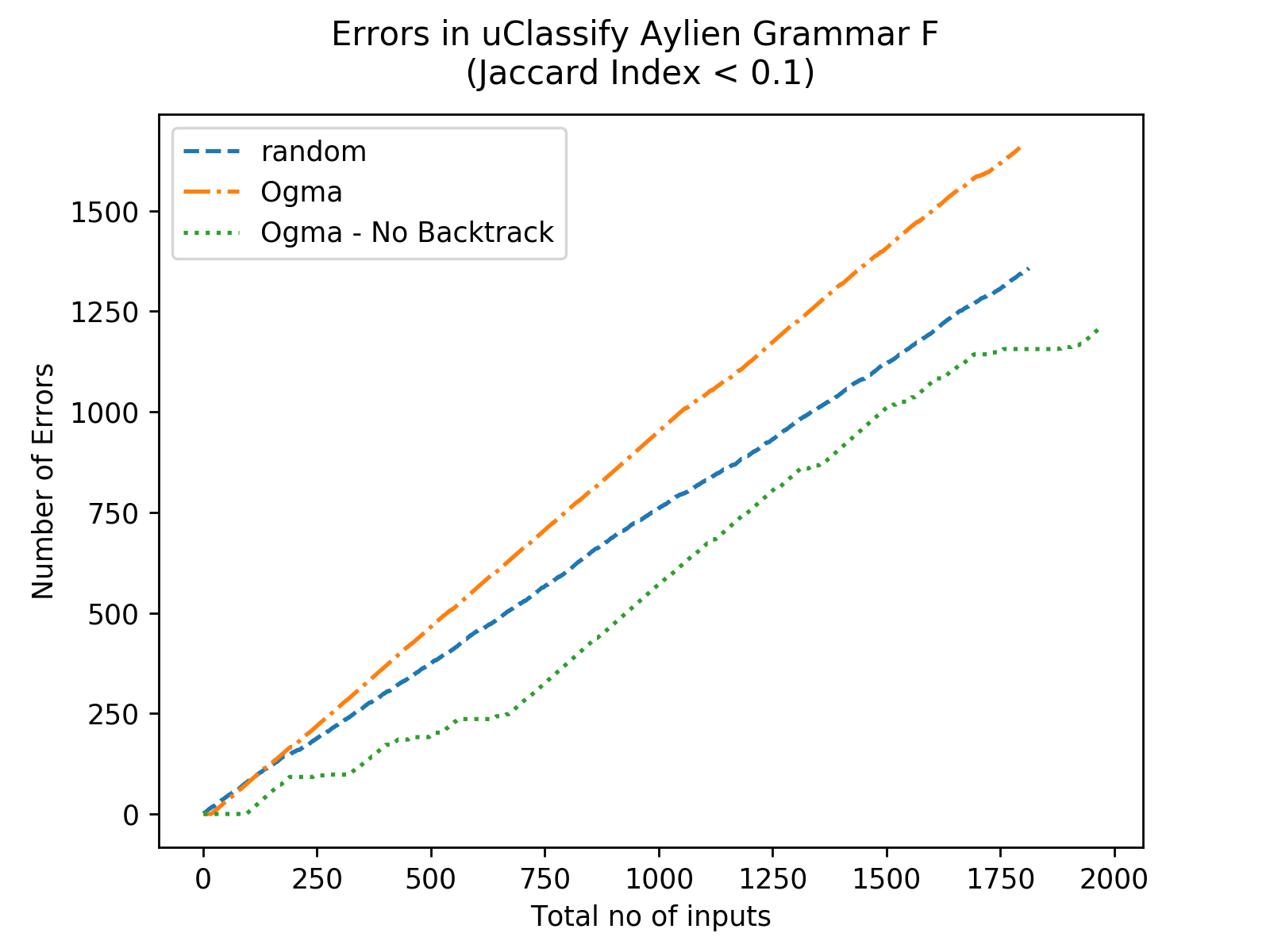} & 
\includegraphics[scale=0.3]{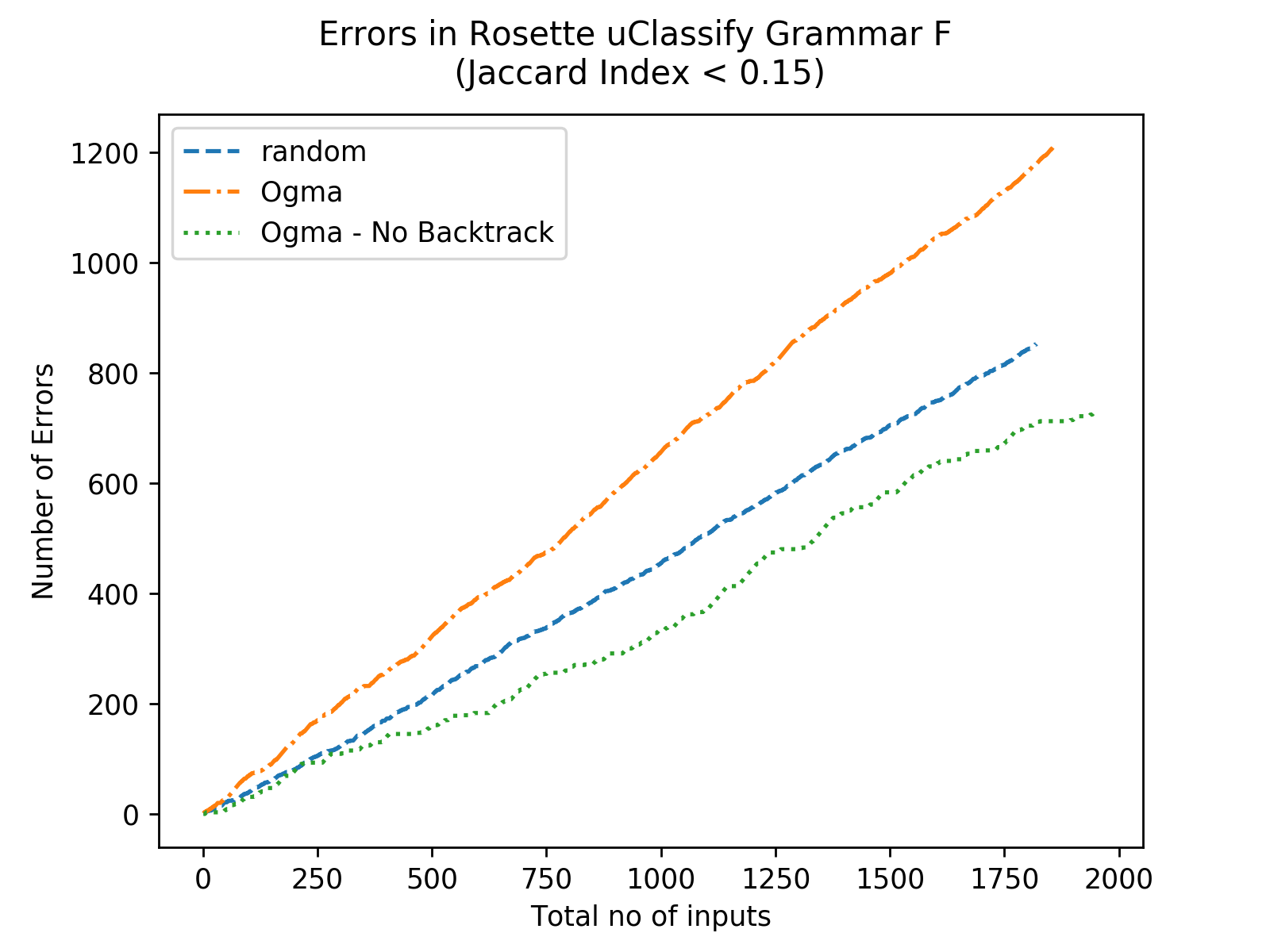}\\

\\
\end{tabular}
\end{center}
\caption{Results with initial input being non-error inducing}
\label{fig:non-error-start}
\end{figure*}

\newpage

\begin{figure*}[t]
\begin{center}
\begin{tabular}{ccc}
\includegraphics[scale=0.3]{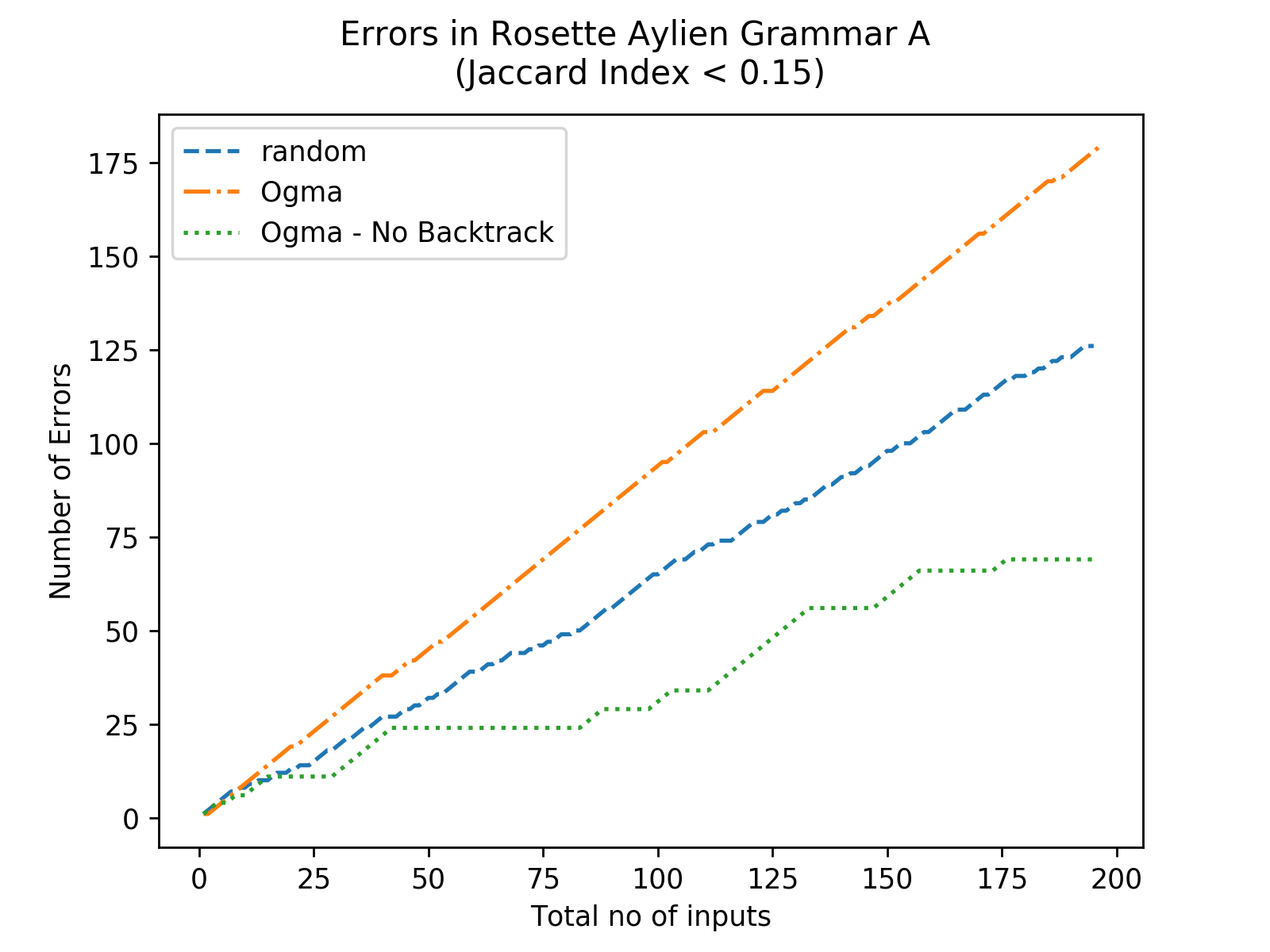} & 
\includegraphics[scale=0.3]{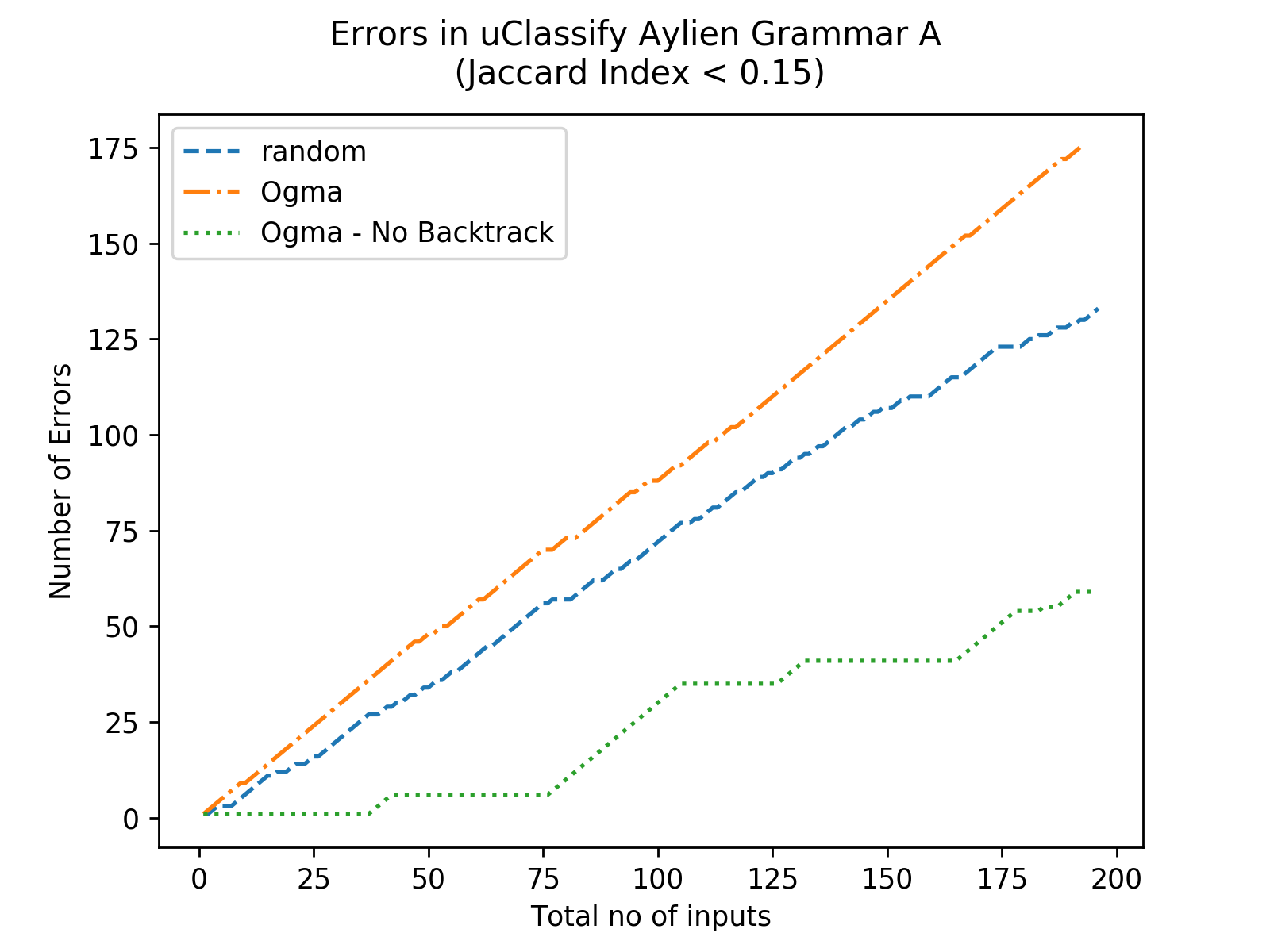} & 
\includegraphics[scale=0.3]{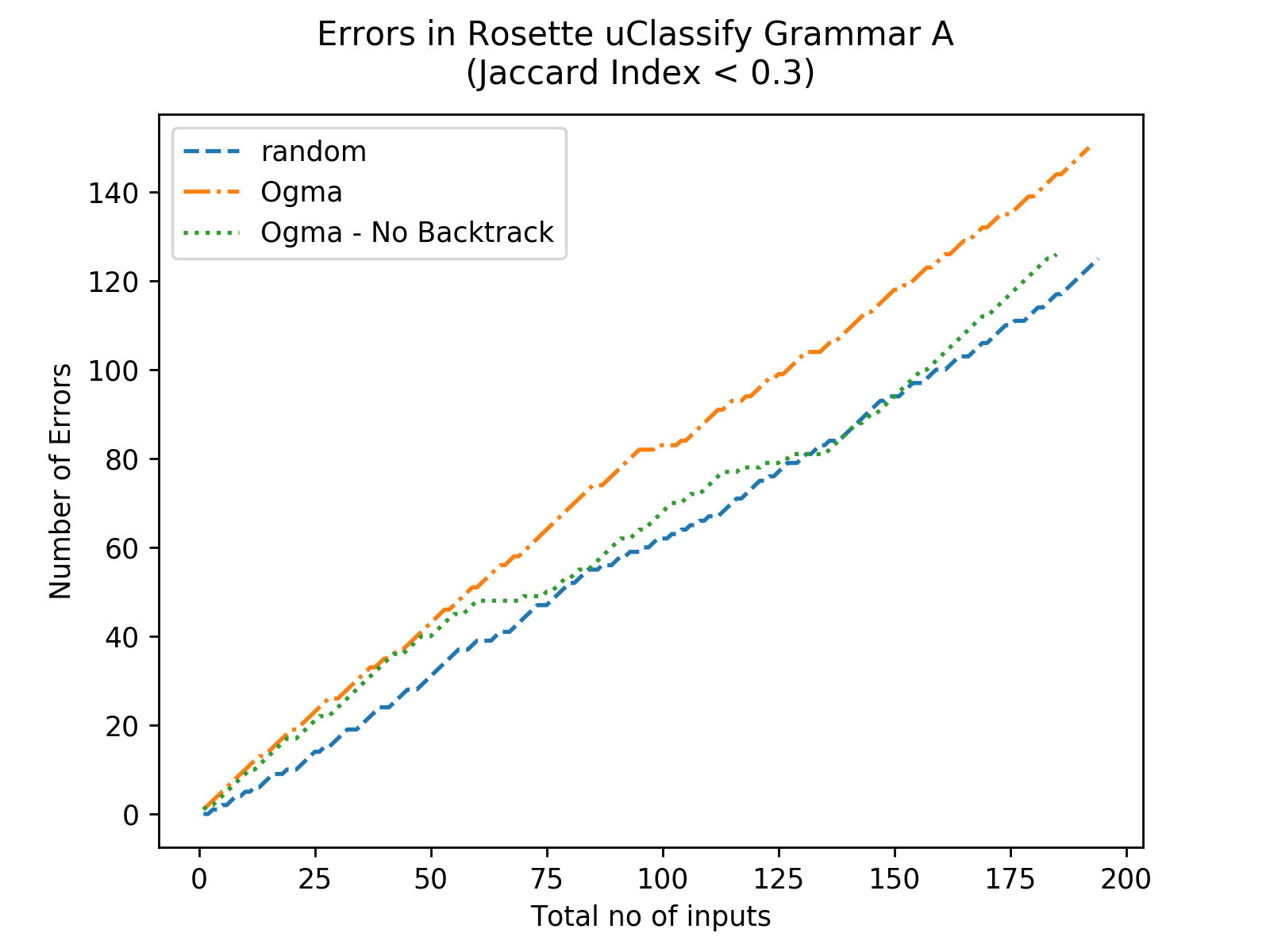}\\

\\
\includegraphics[scale=0.3]{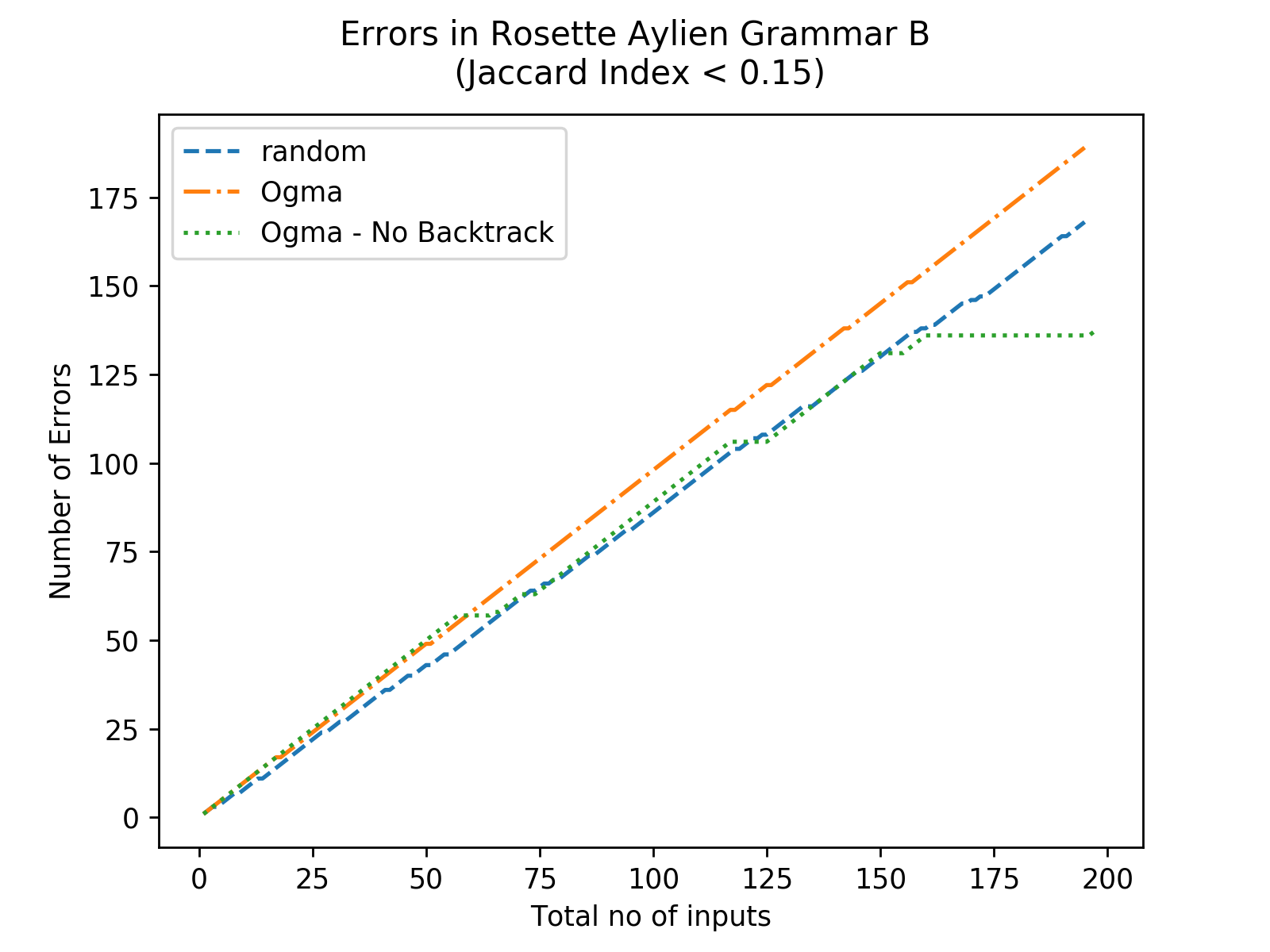} & 
\includegraphics[scale=0.3]{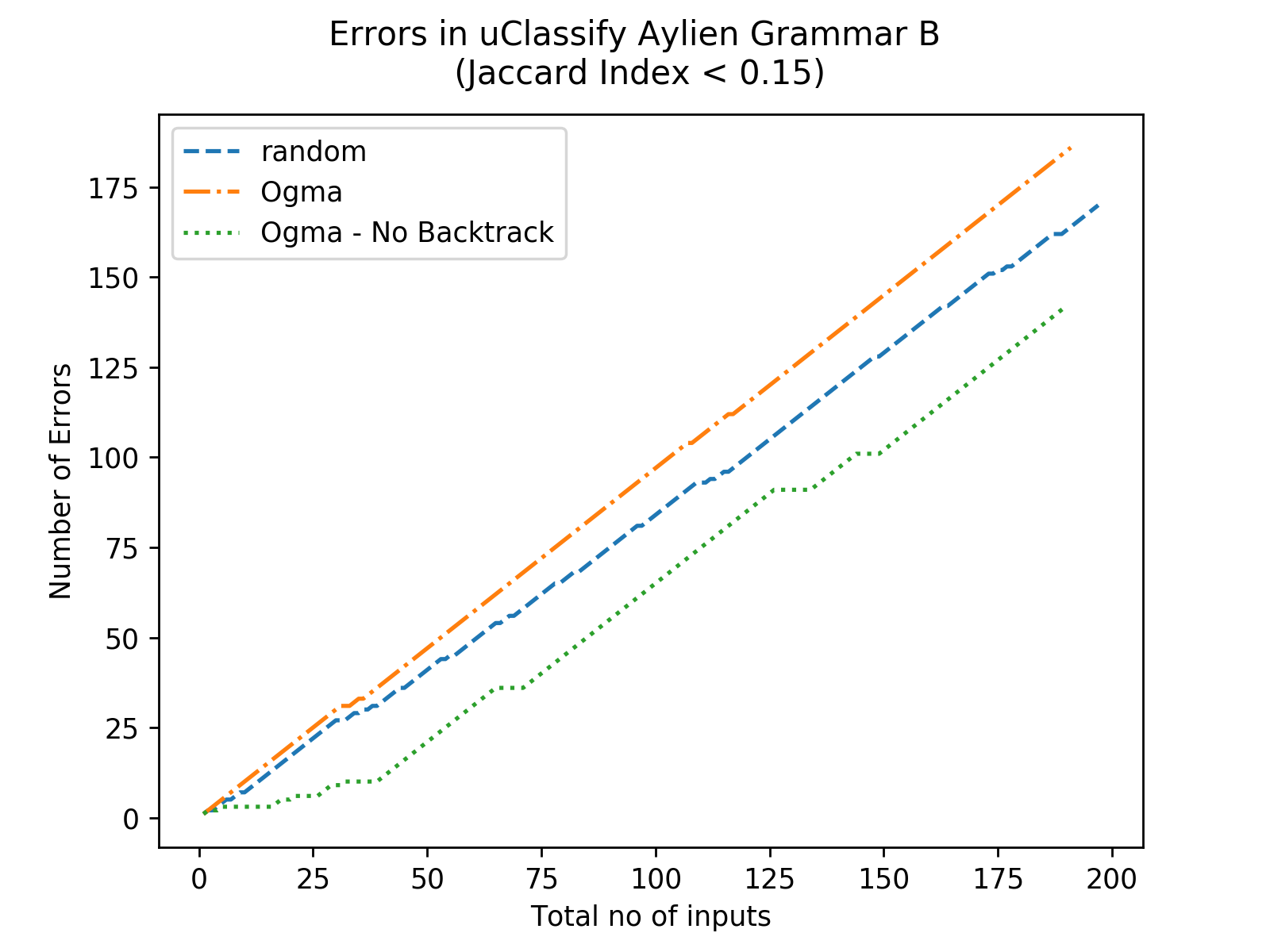} & 
\includegraphics[scale=0.3]{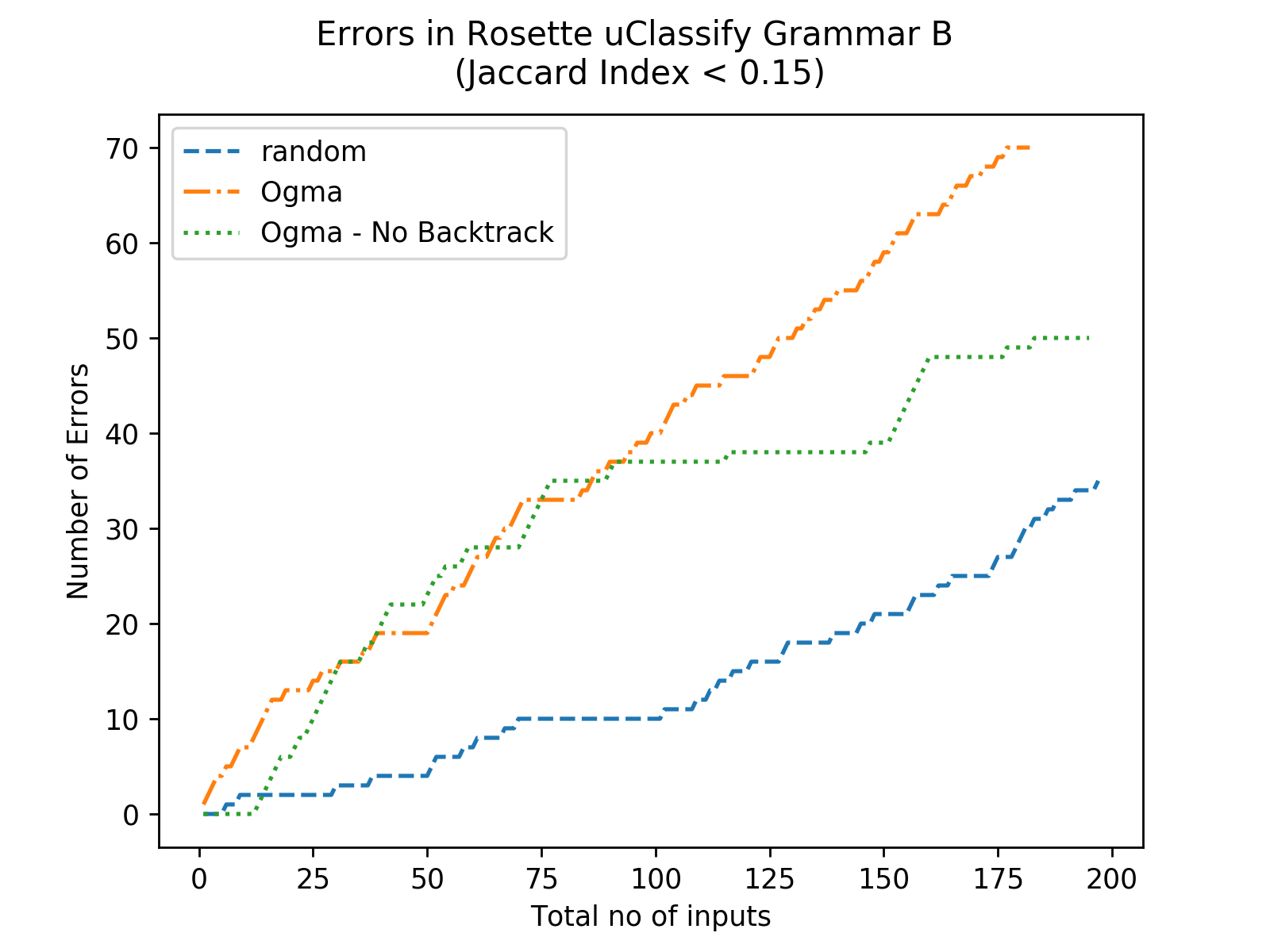}\\

\\
\includegraphics[scale=0.3]{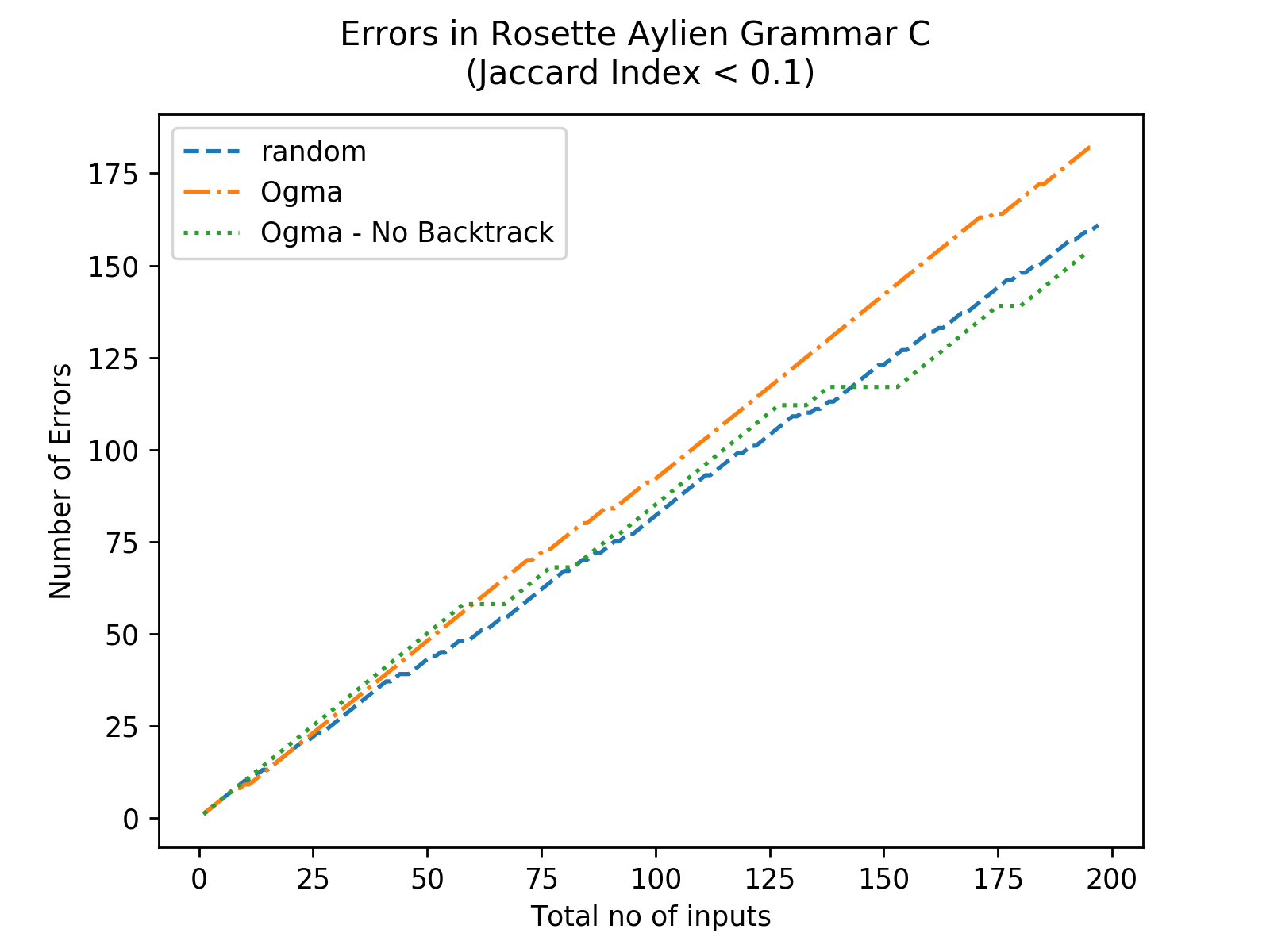} & 
\includegraphics[scale=0.3]{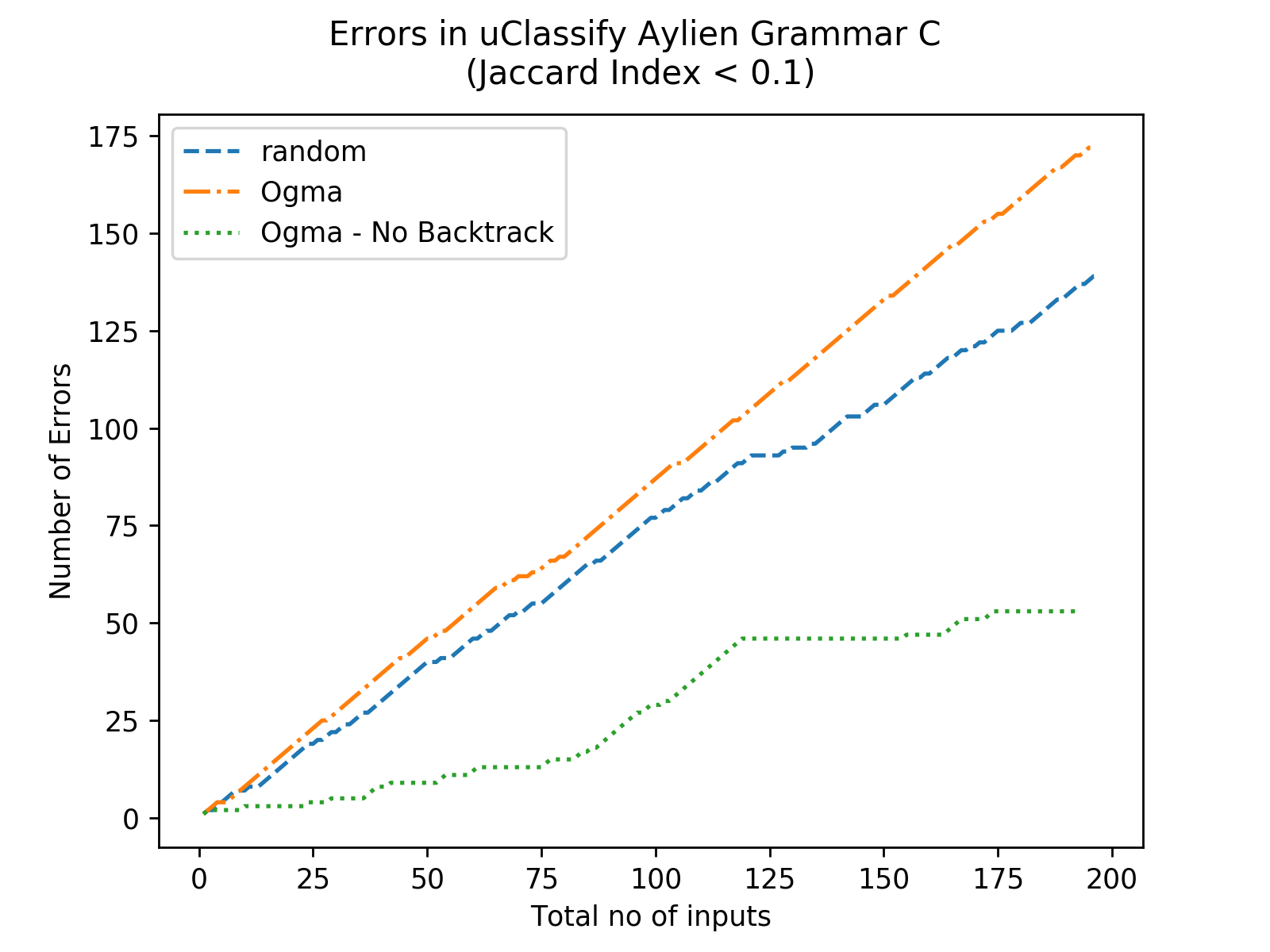} & 
\includegraphics[scale=0.3]{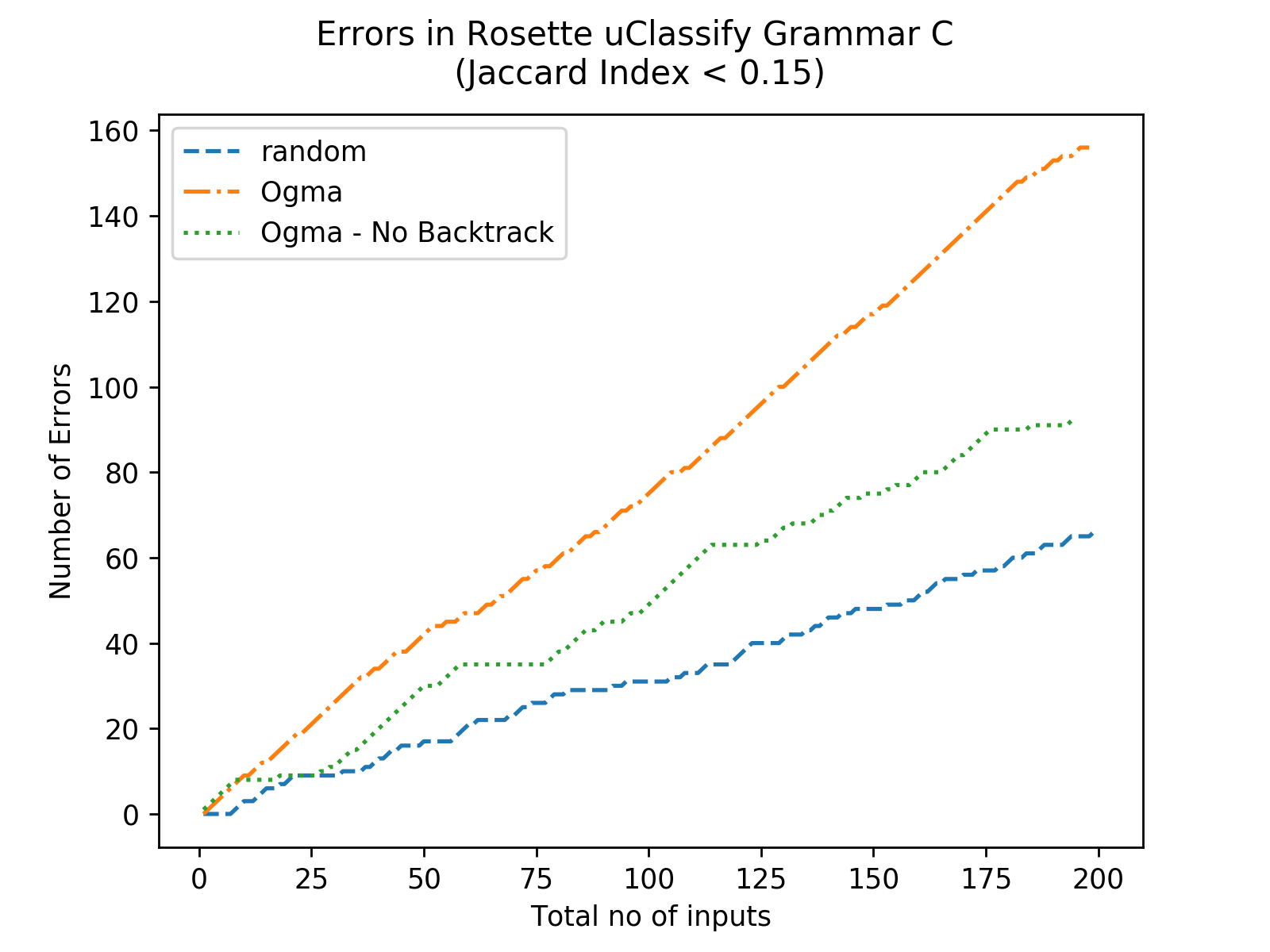}\\

\\
\includegraphics[scale=0.3]{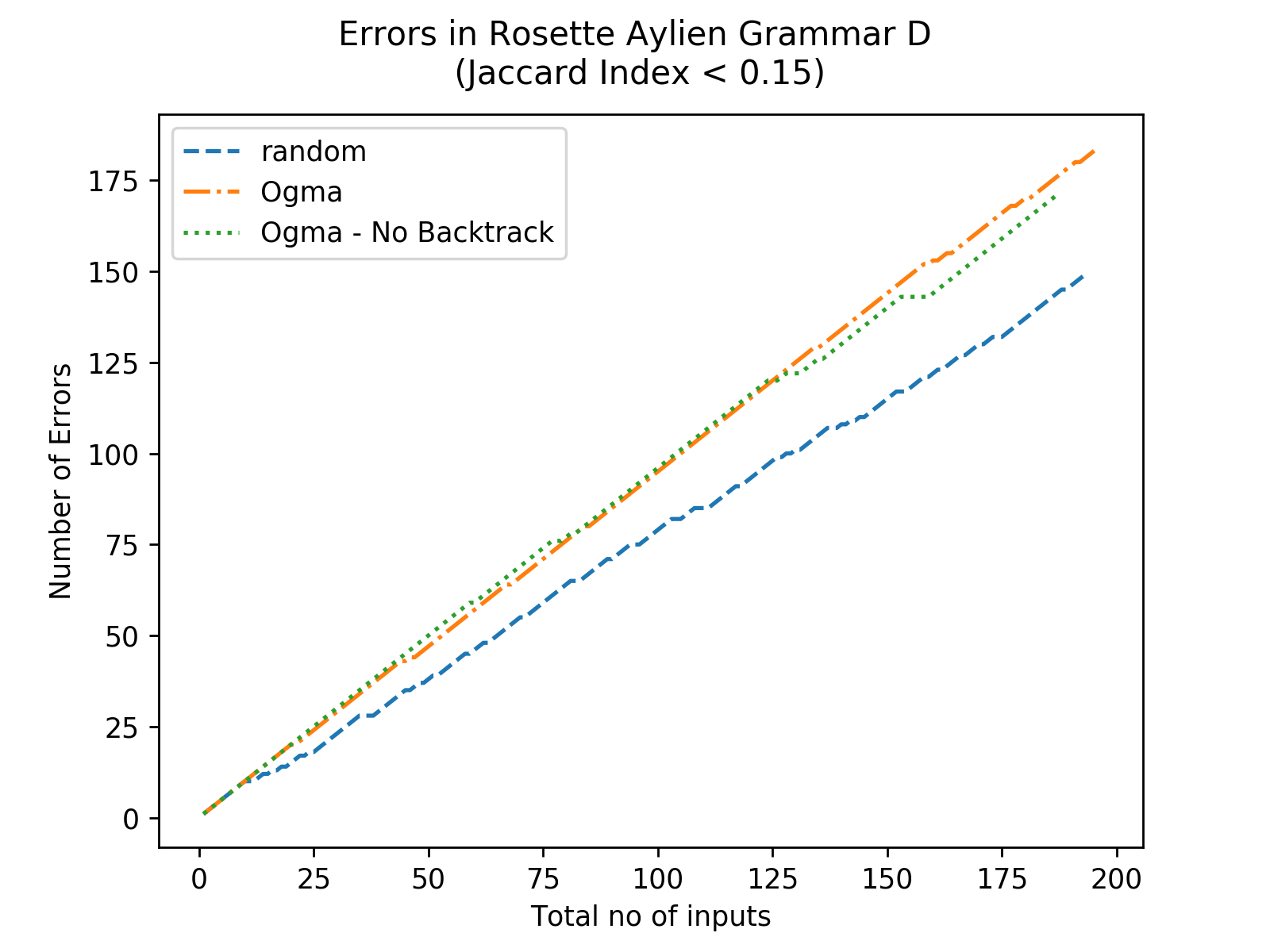} & 
\includegraphics[scale=0.3]{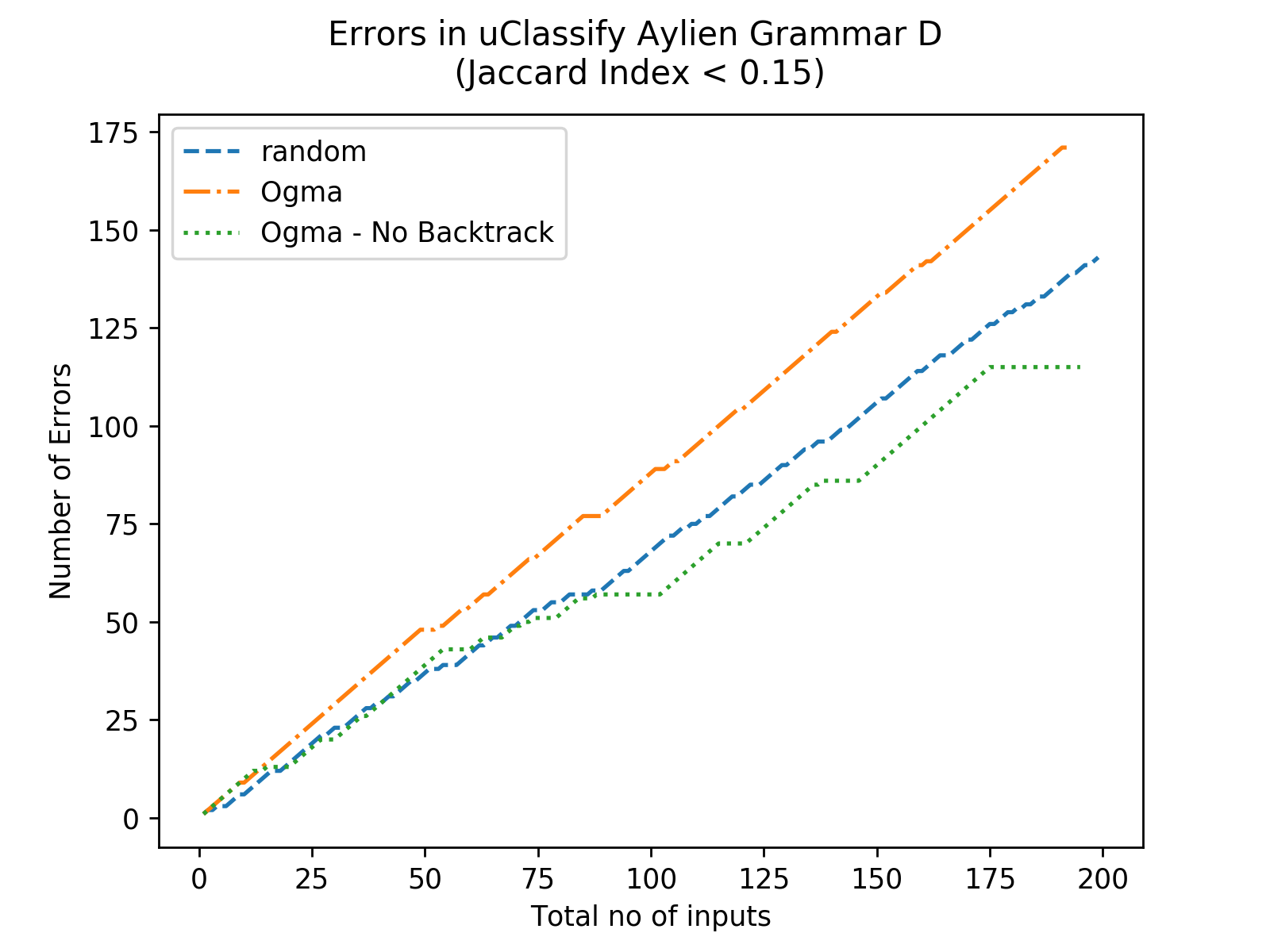} & 
\includegraphics[scale=0.3]{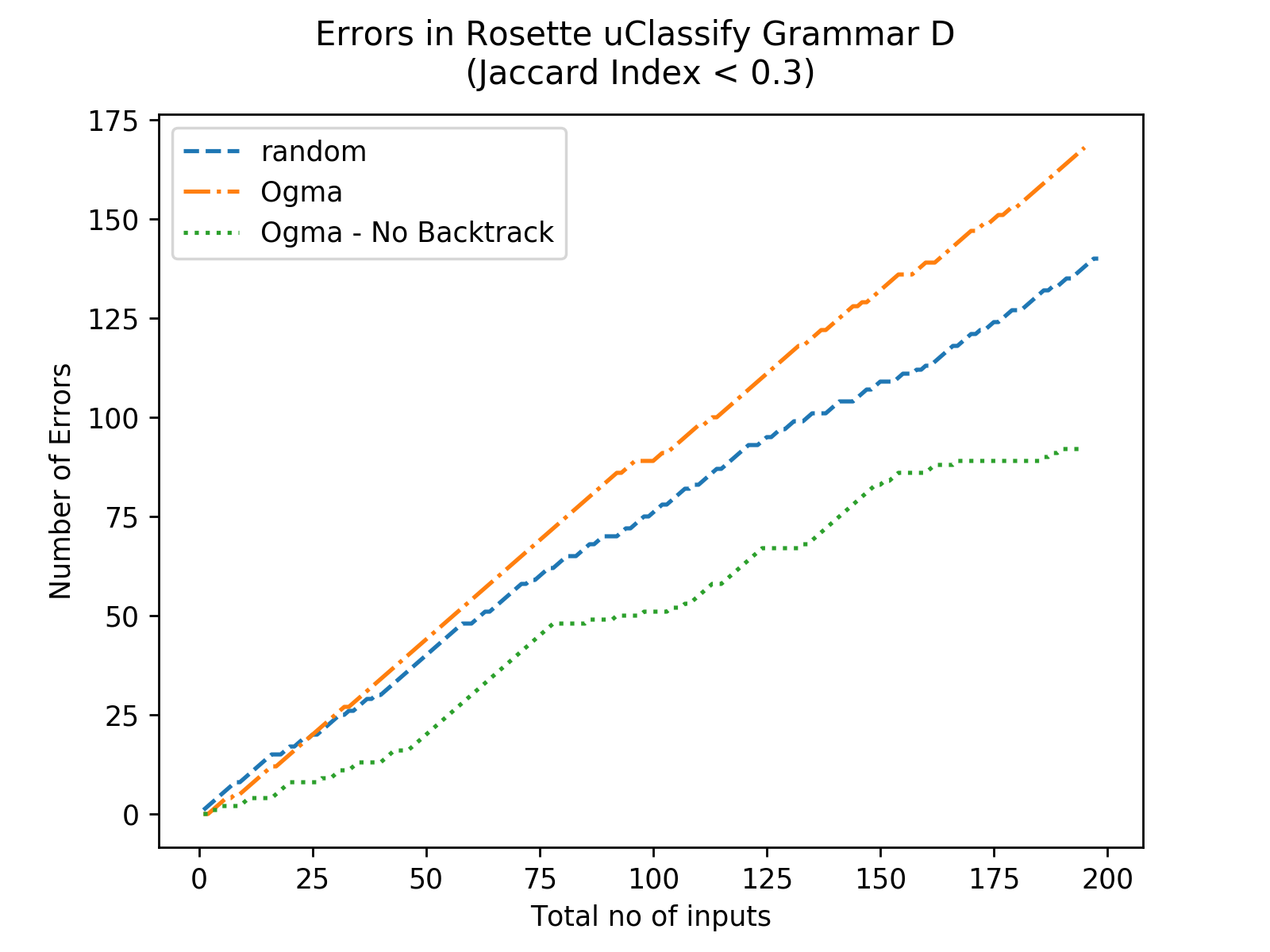}\\

\\
\includegraphics[scale=0.3]{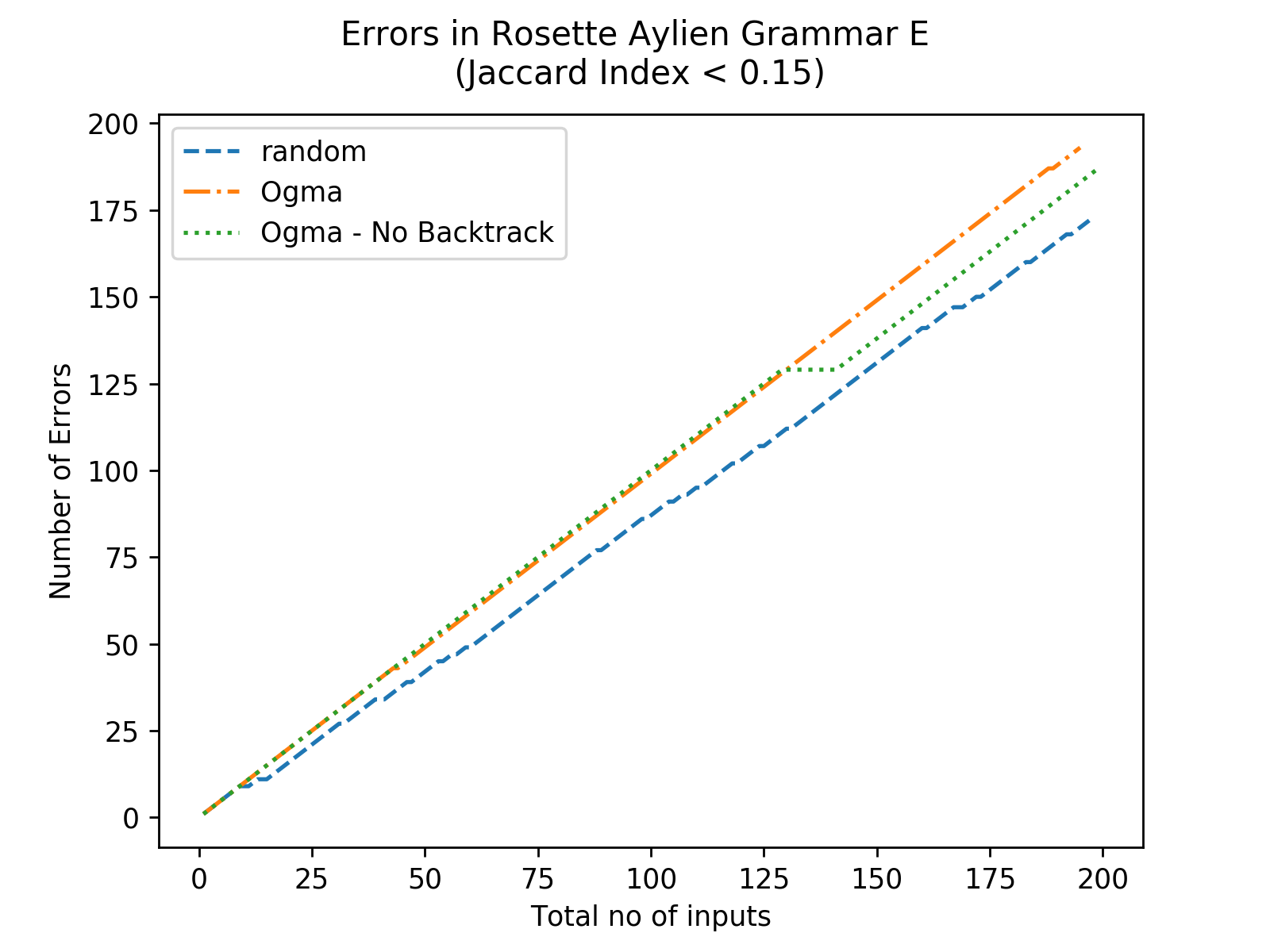} & 
\includegraphics[scale=0.3]{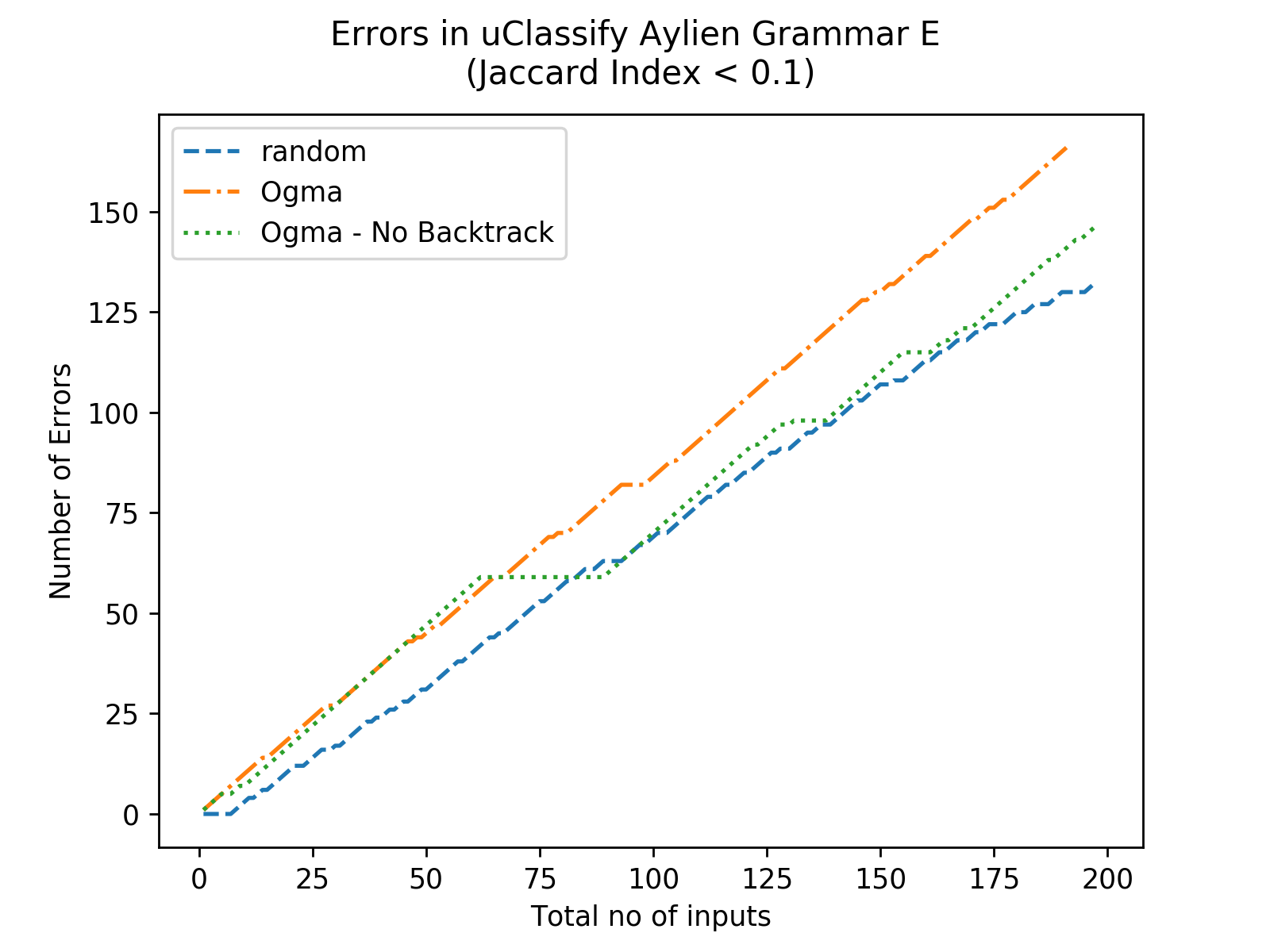} & 
\includegraphics[scale=0.3]{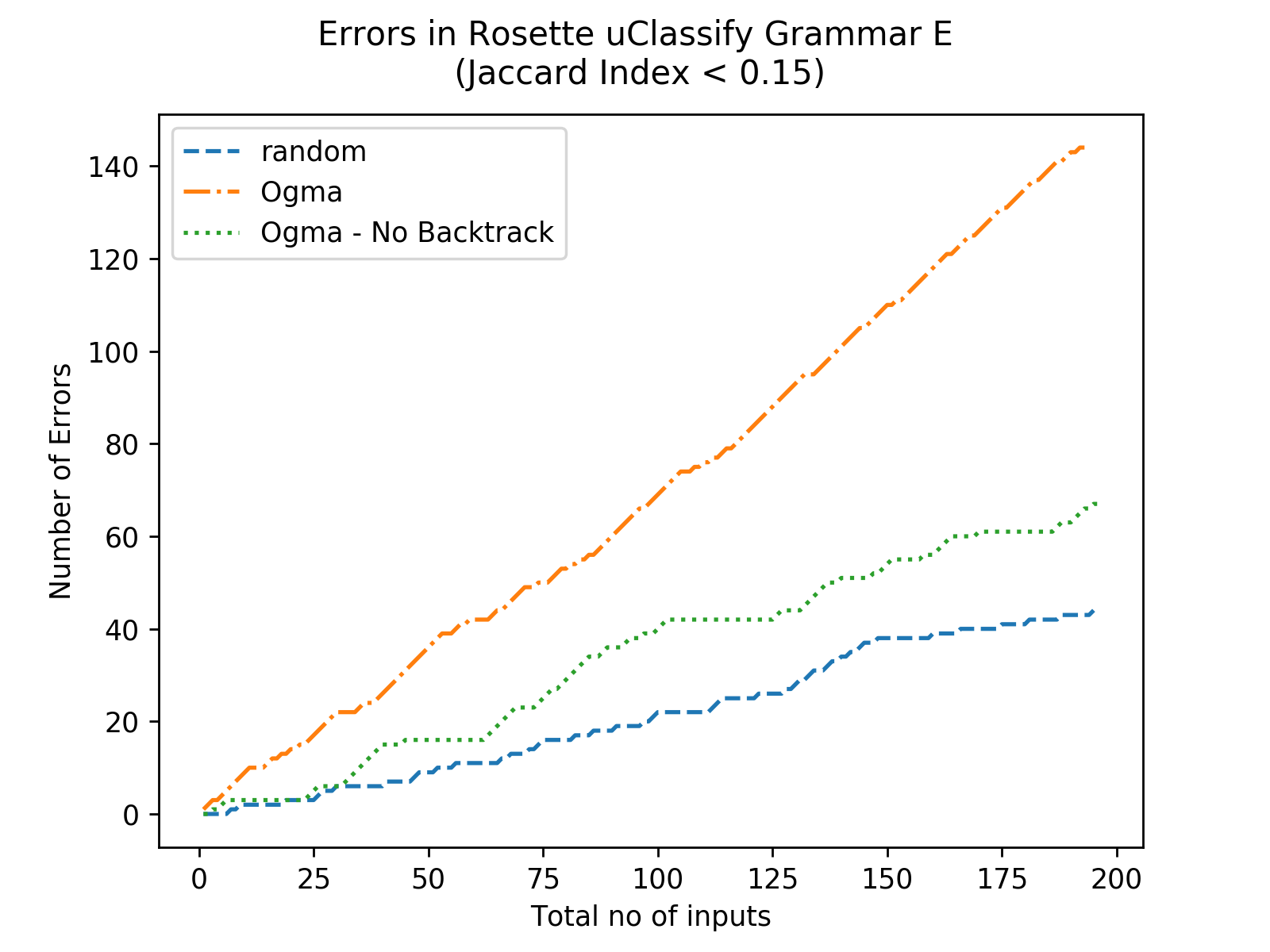}\\

\\
\includegraphics[scale=0.3]{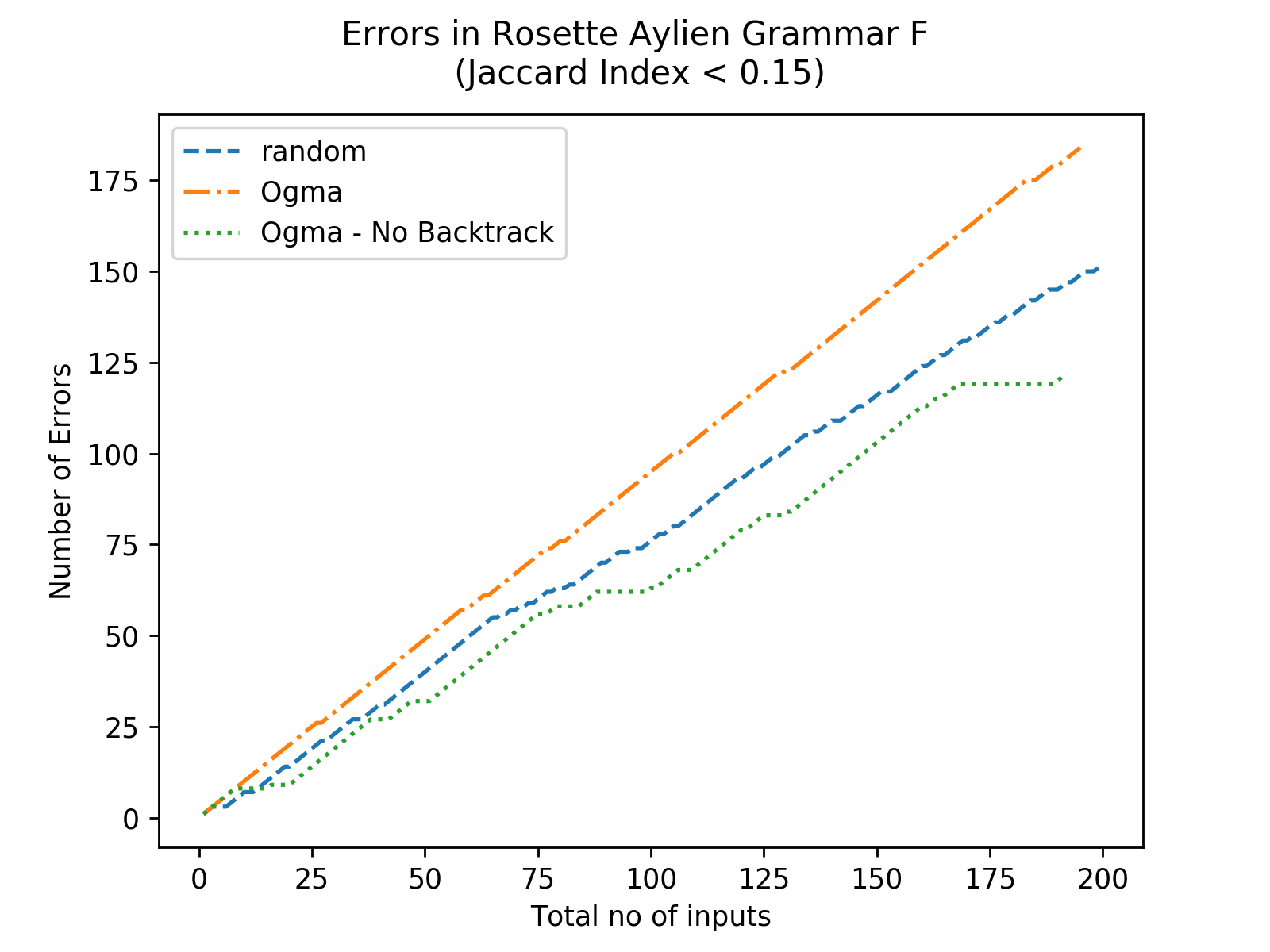} & 
\includegraphics[scale=0.3]{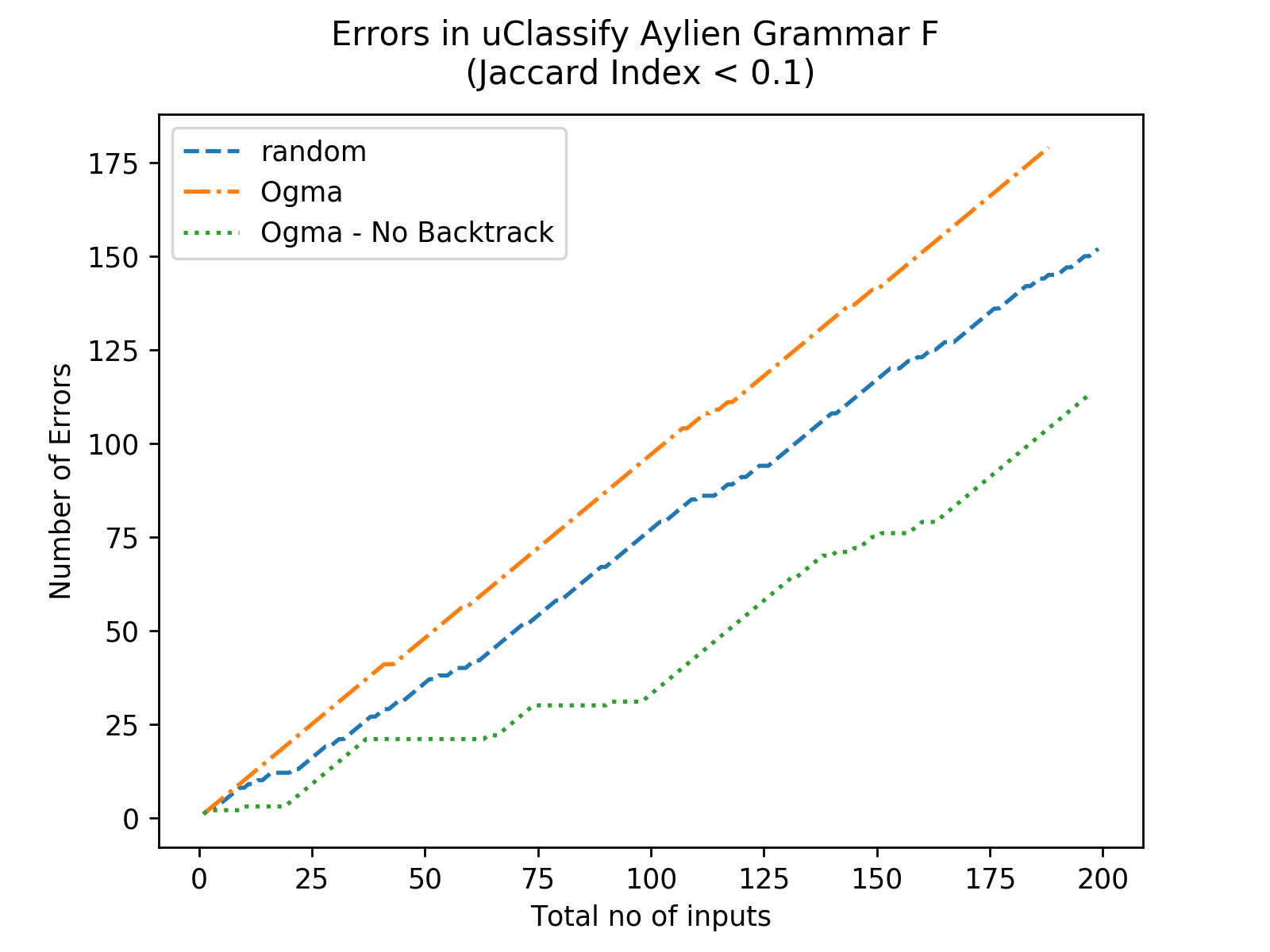} & 
\includegraphics[scale=0.3]{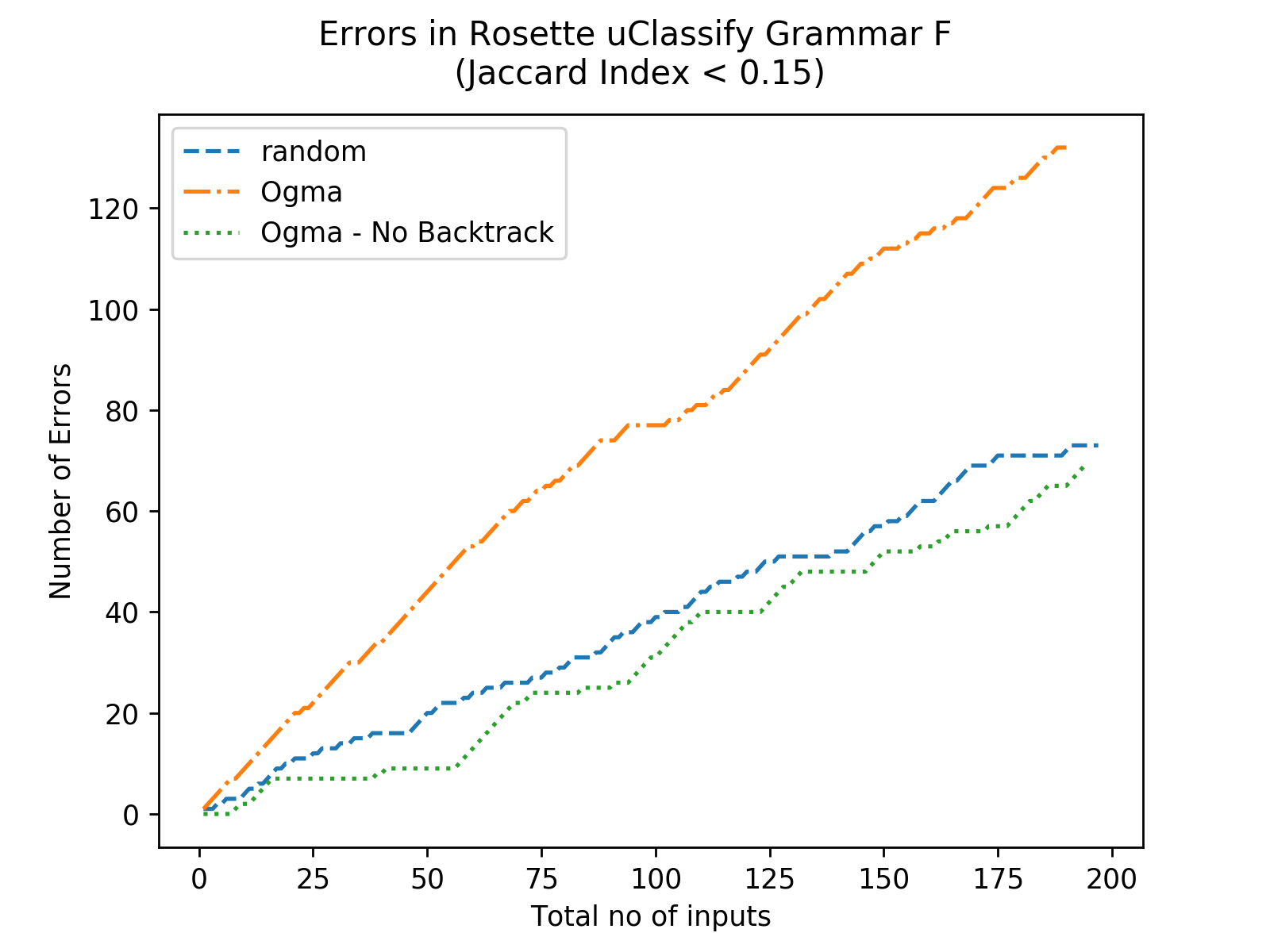}\\

\\
\end{tabular}
\end{center}
\caption{Results with initial input being error inducing}
\label{fig:error-start}
\end{figure*}

\begin{figure*}[t]
\begin{center}
\begin{tabular}{ccc}
\includegraphics[scale=0.3]{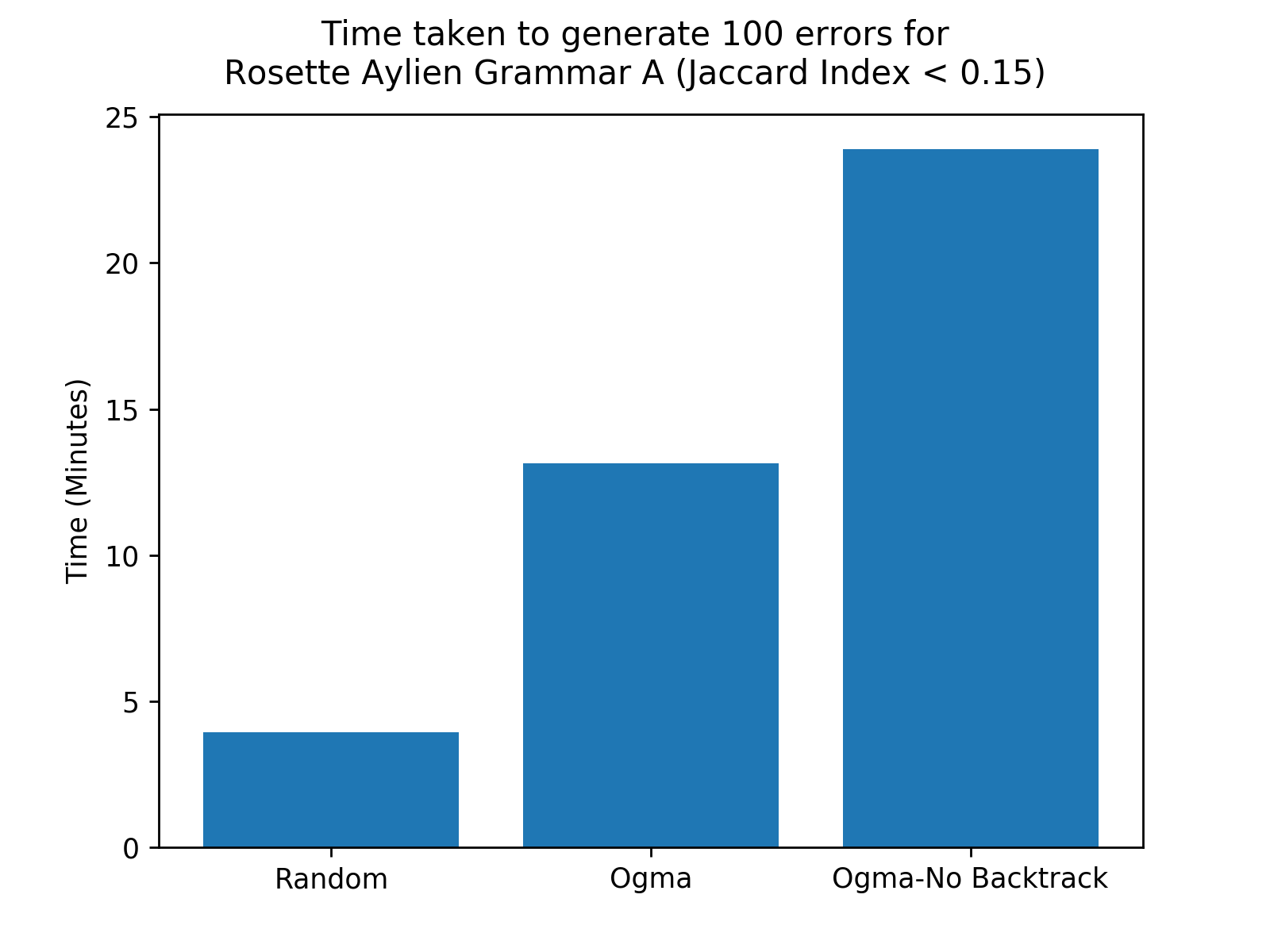} & 
\includegraphics[scale=0.3]{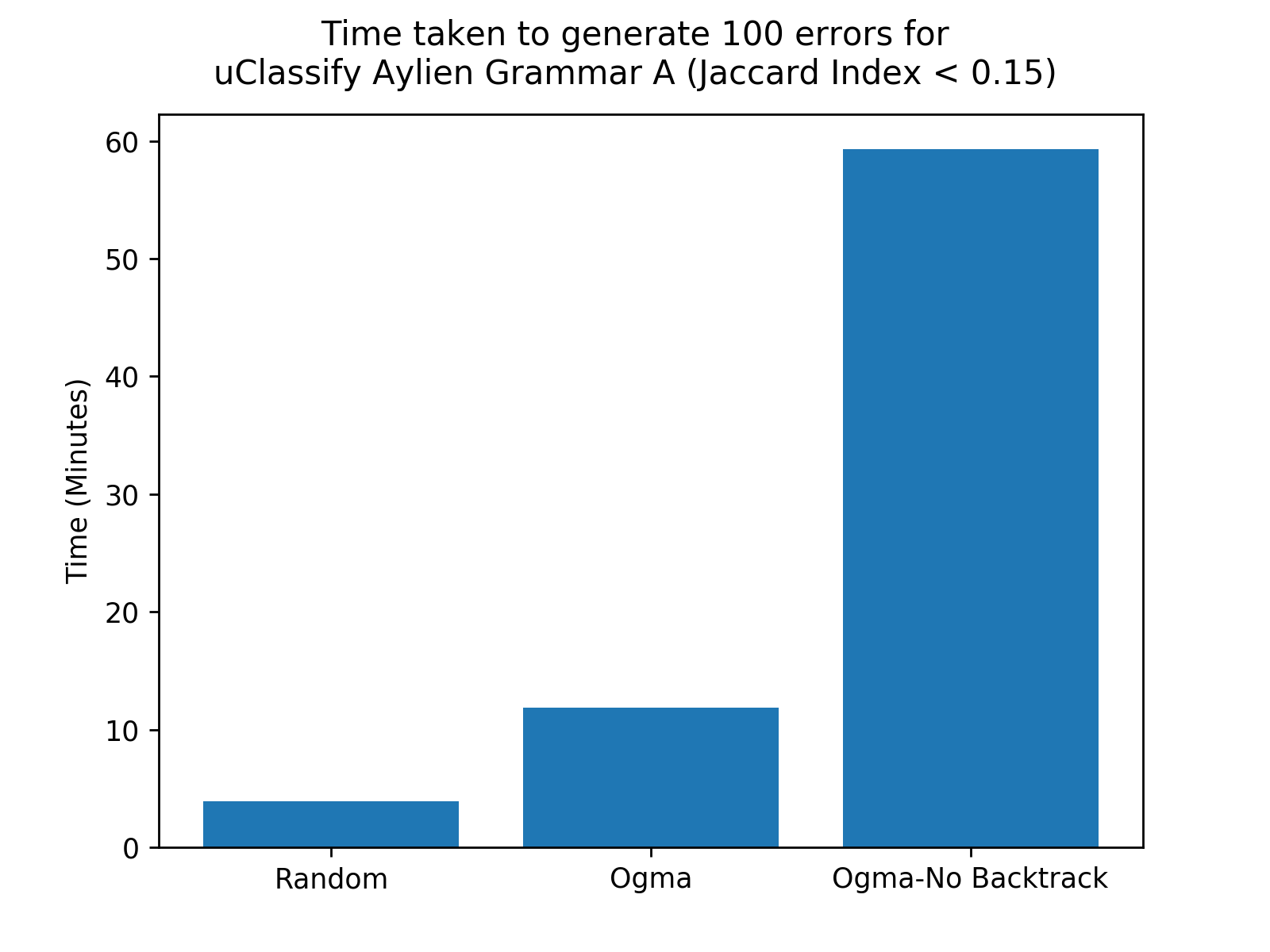} & 
\includegraphics[scale=0.3]{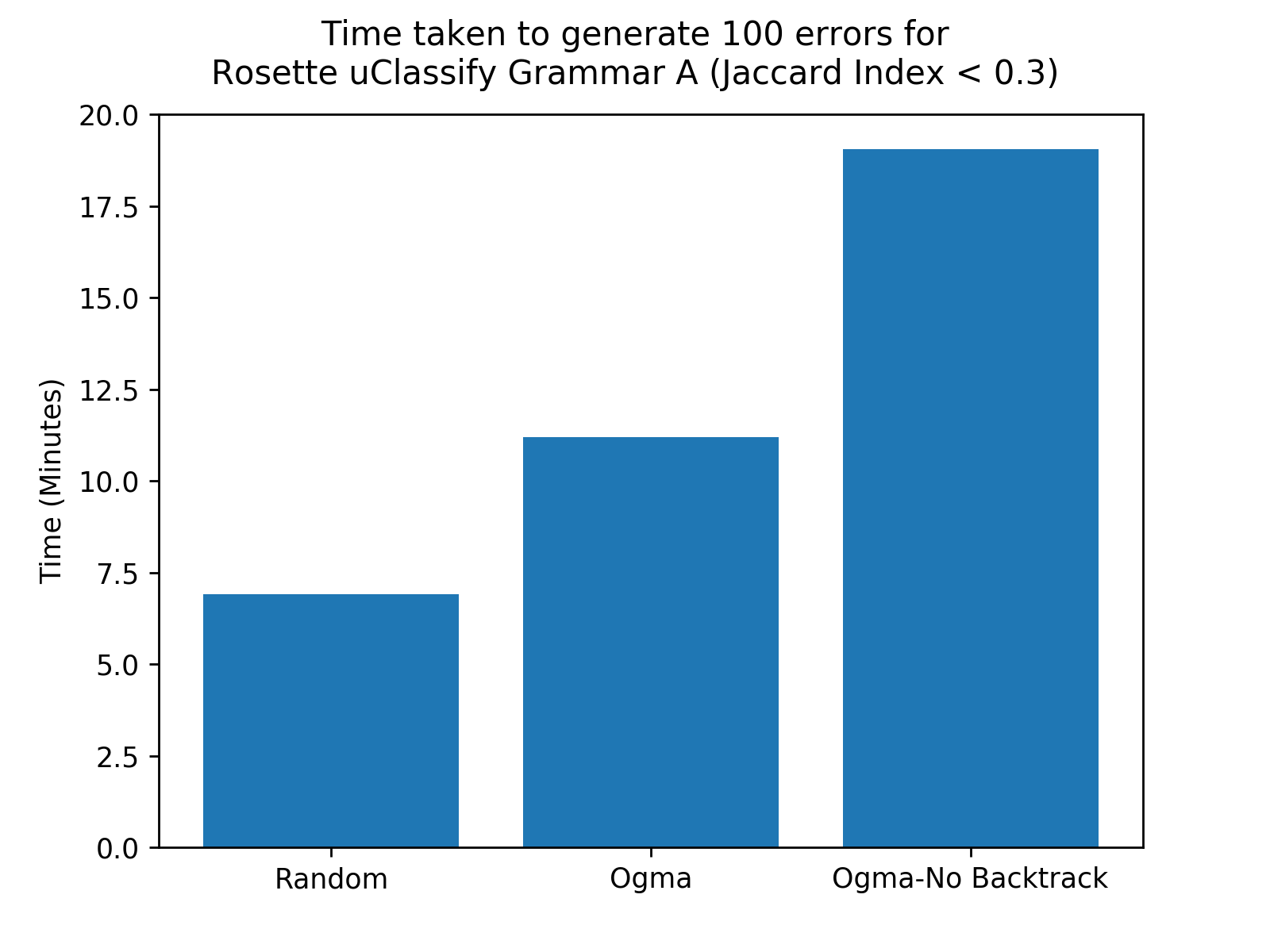}\\

\\
\includegraphics[scale=0.3]{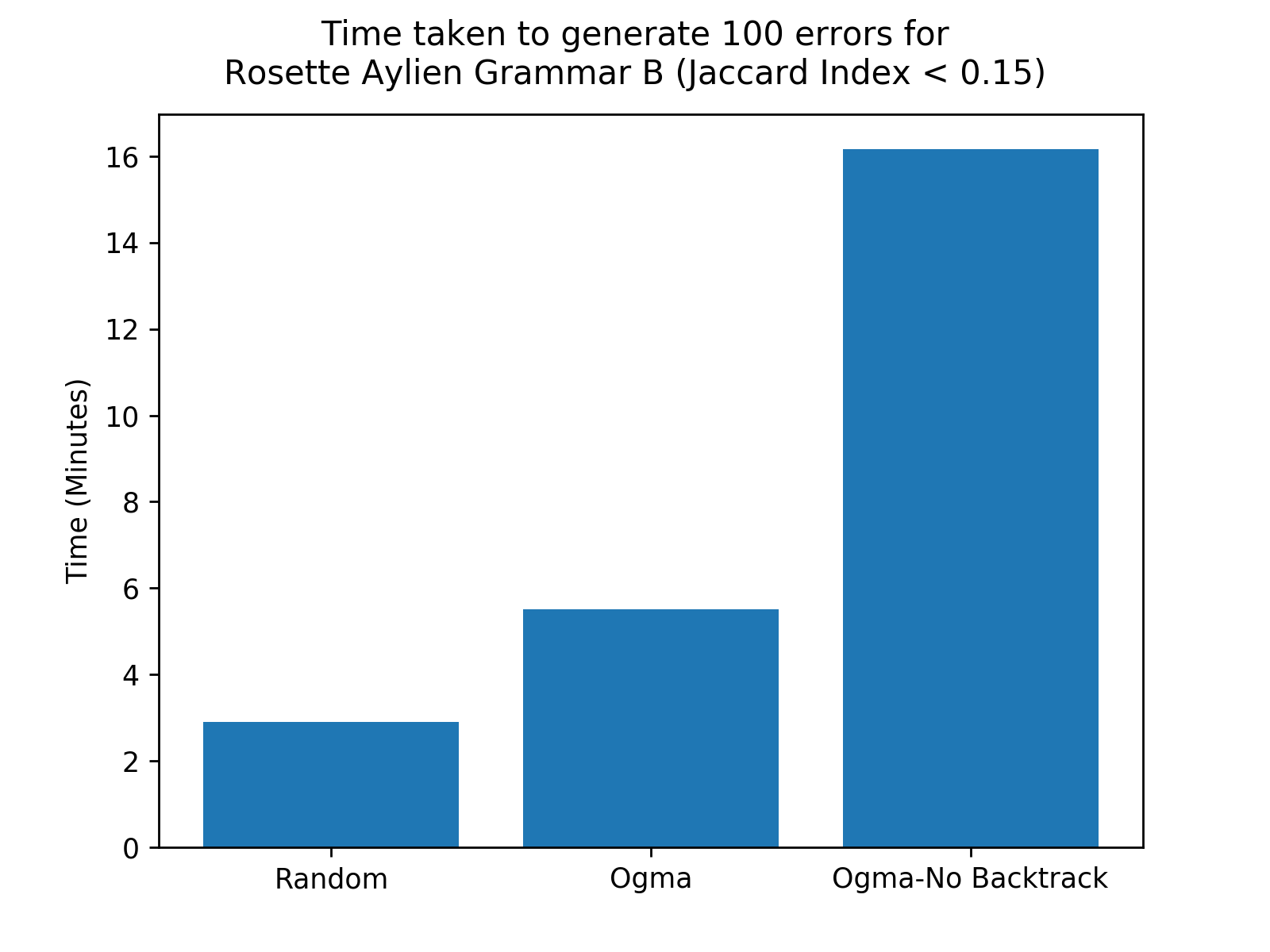} & 
\includegraphics[scale=0.3]{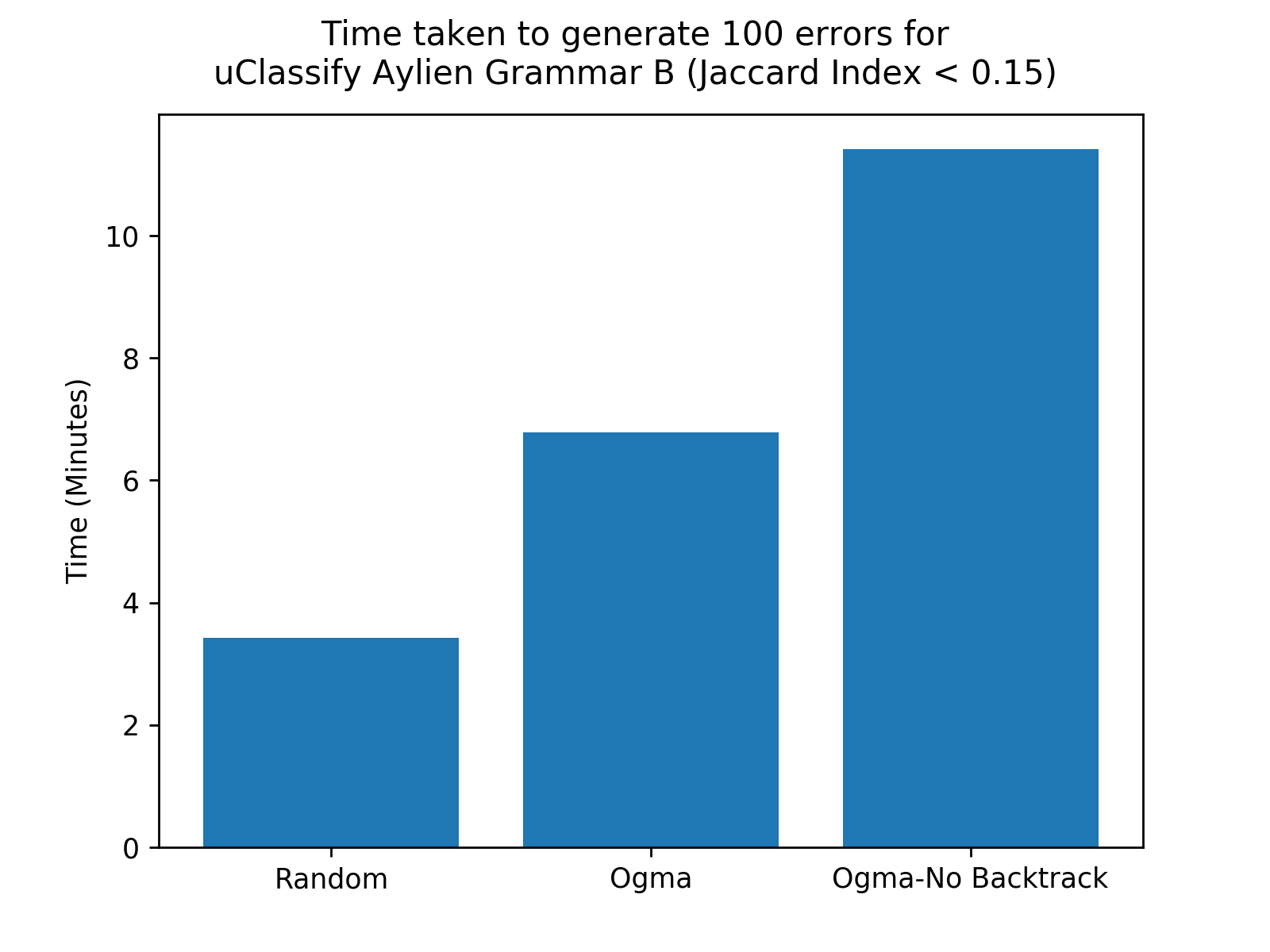} & 
\includegraphics[scale=0.3]{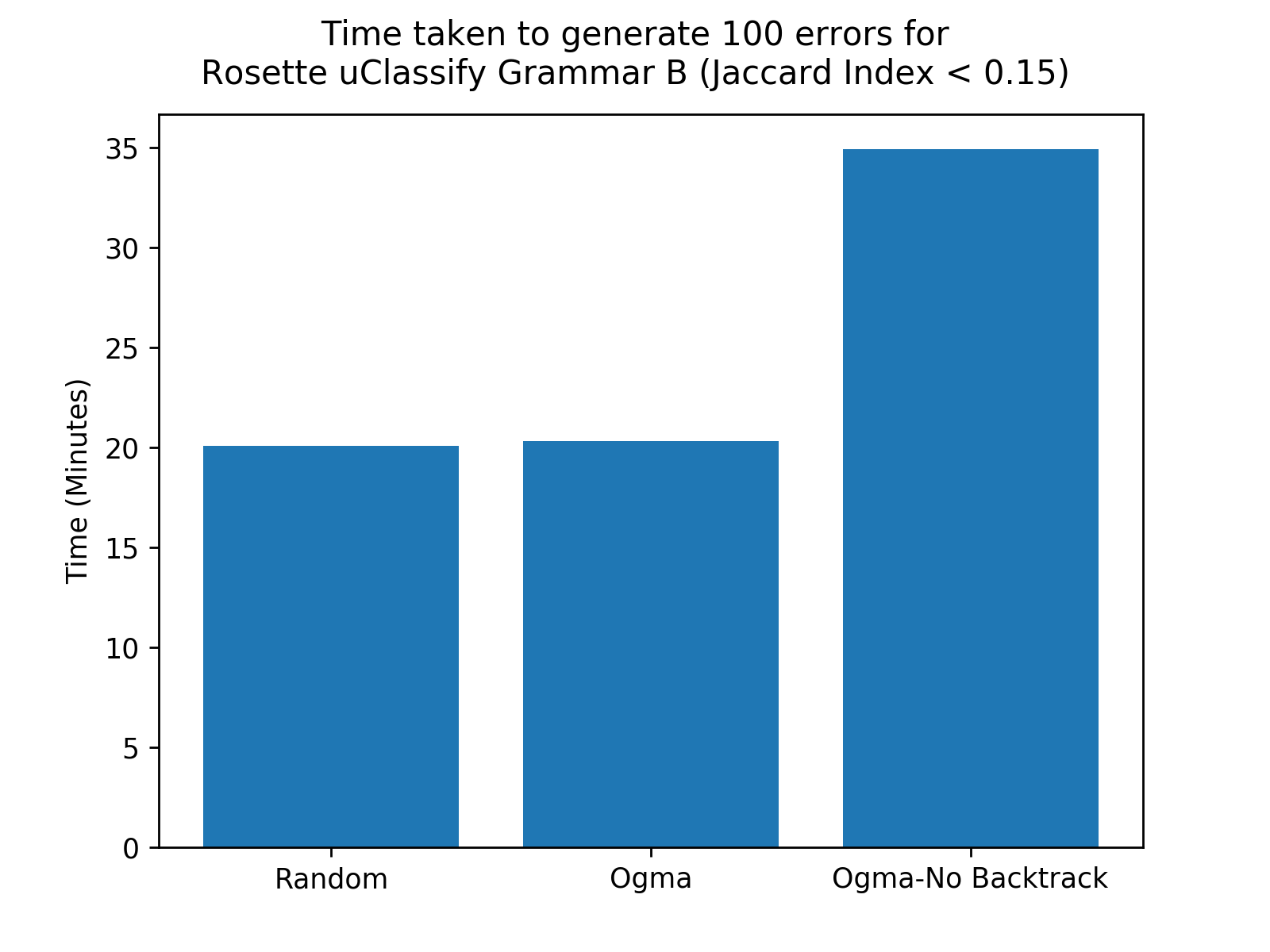}\\

\\
\includegraphics[scale=0.3]{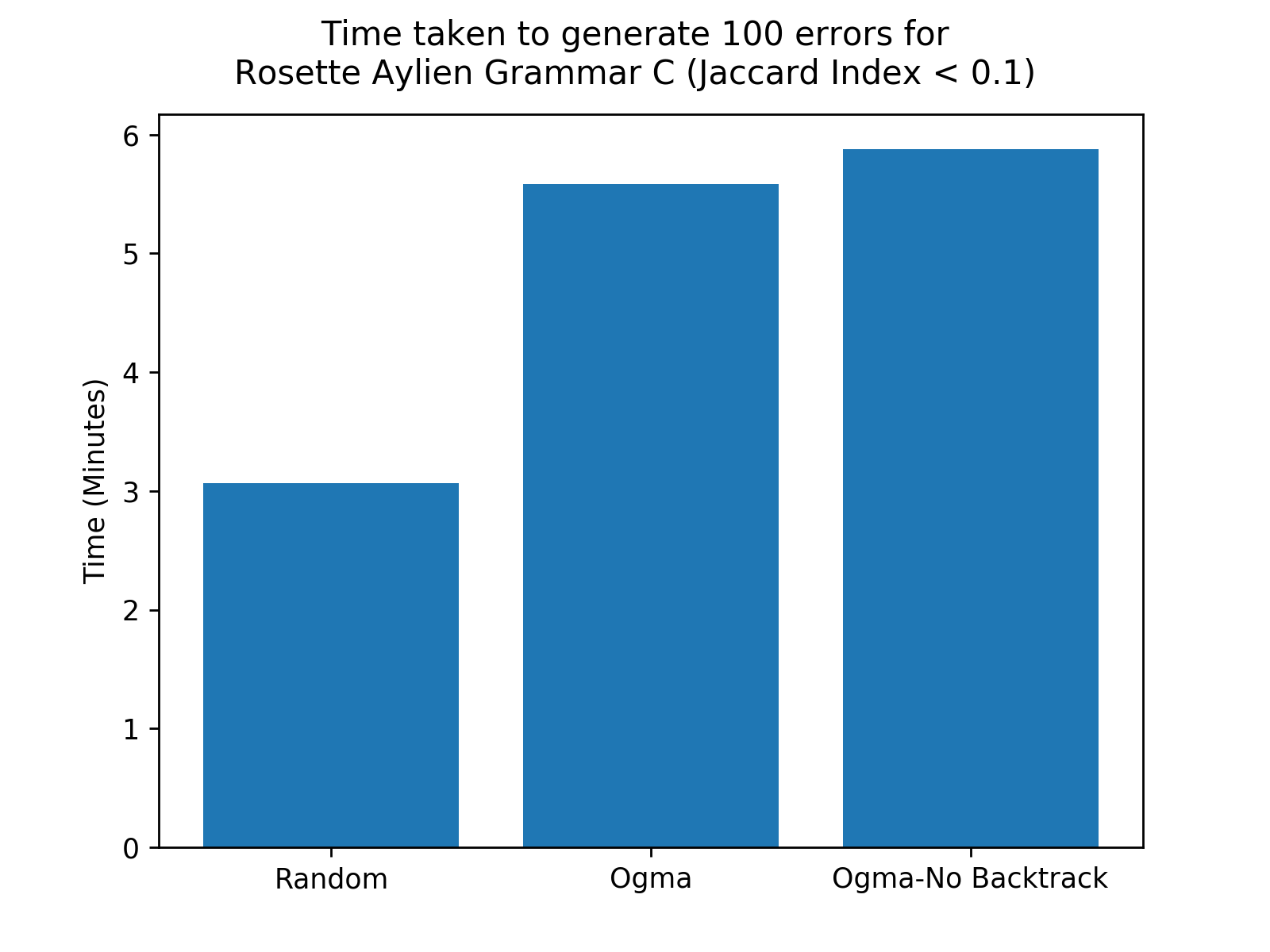} & 
\includegraphics[scale=0.3]{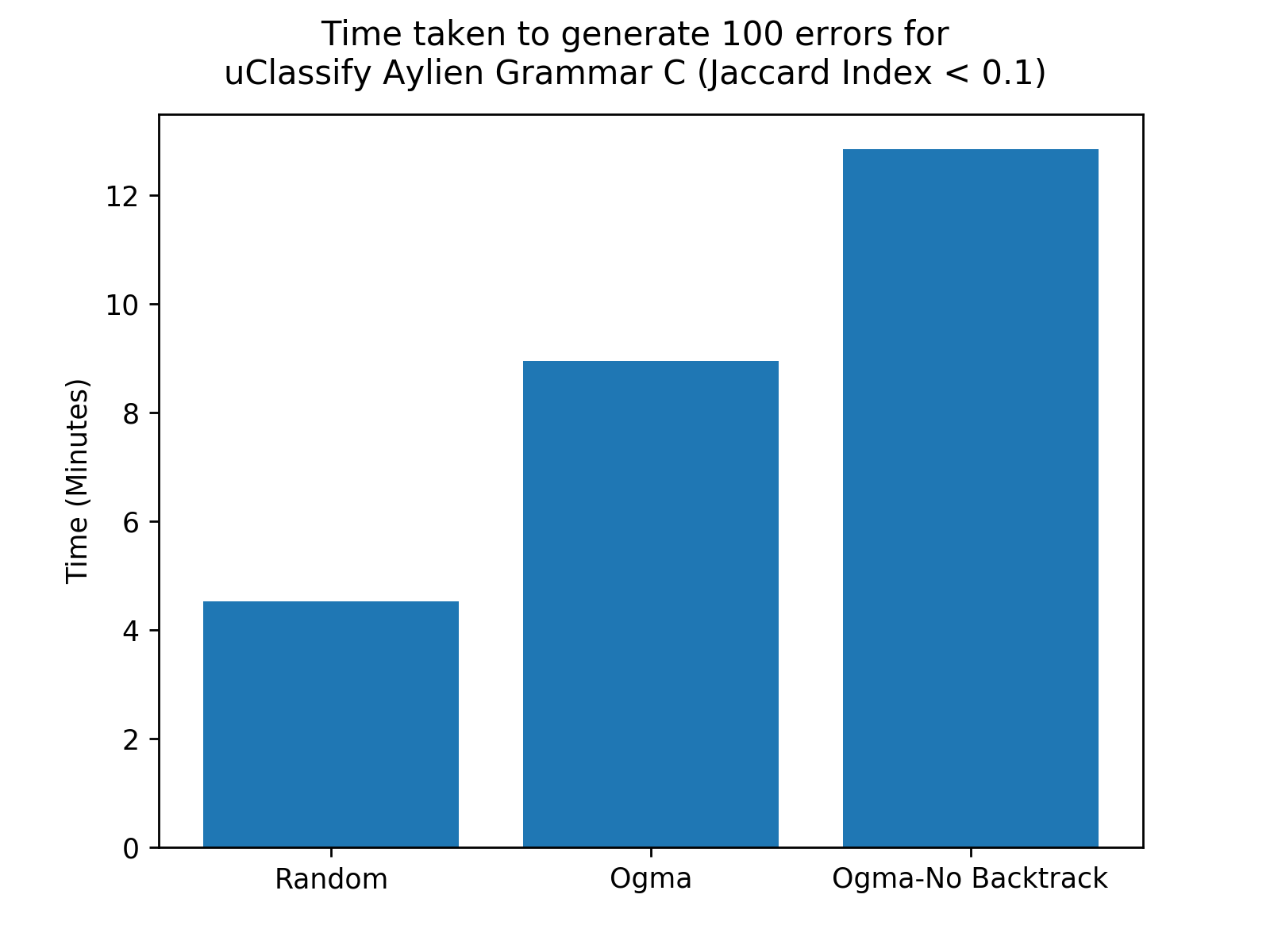} & 
\includegraphics[scale=0.3]{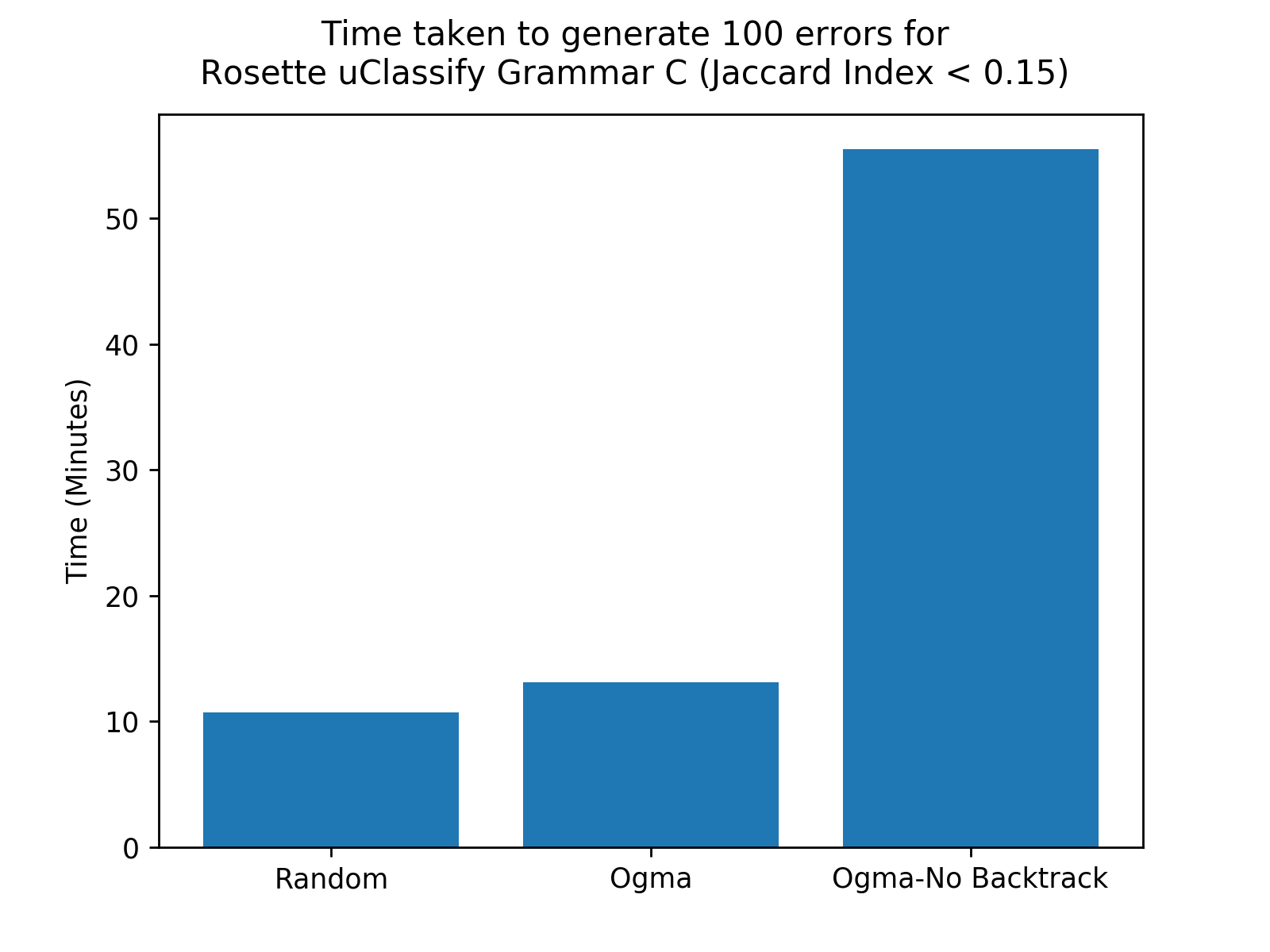}\\

\\
\includegraphics[scale=0.3]{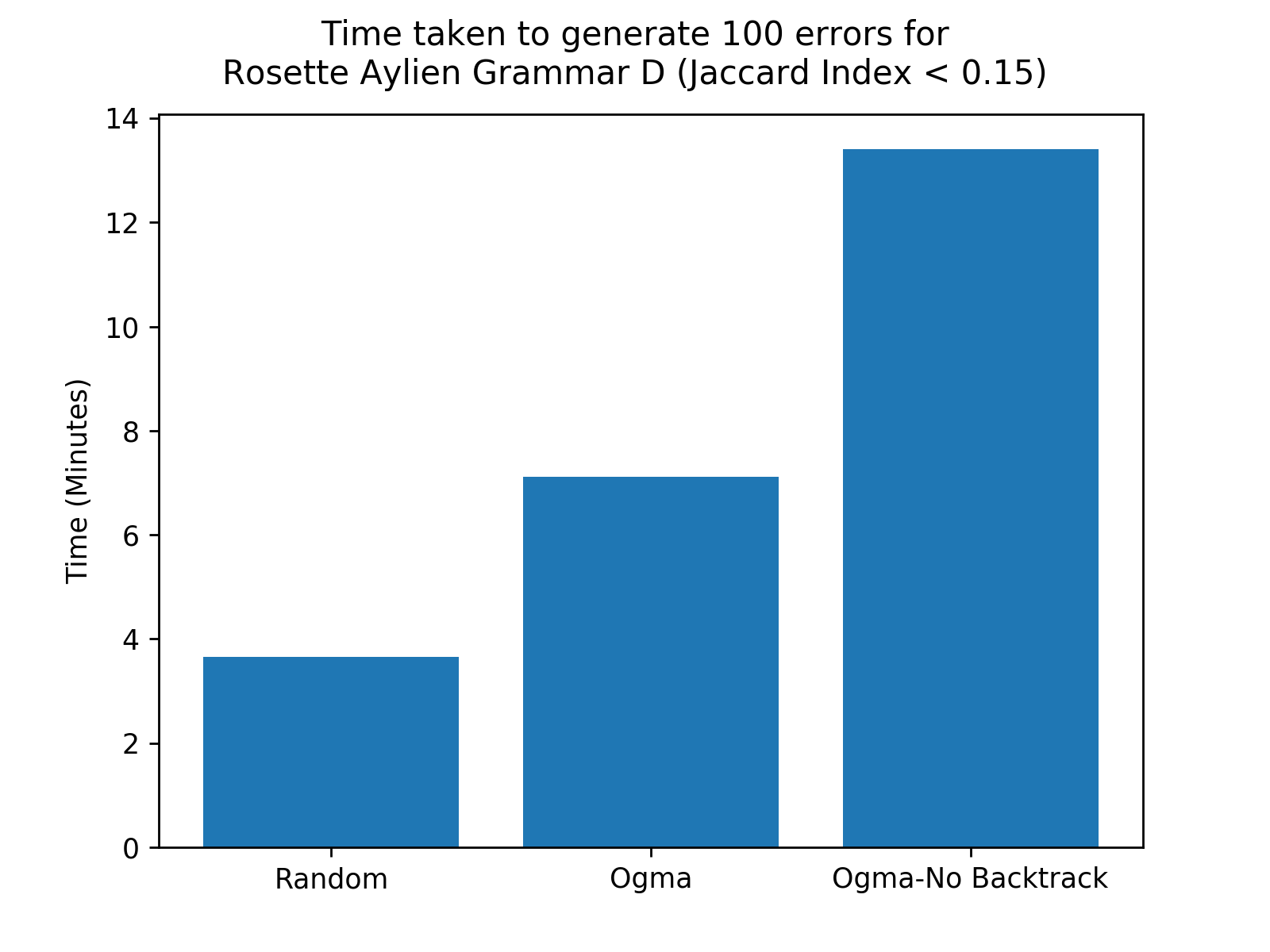} & 
\includegraphics[scale=0.3]{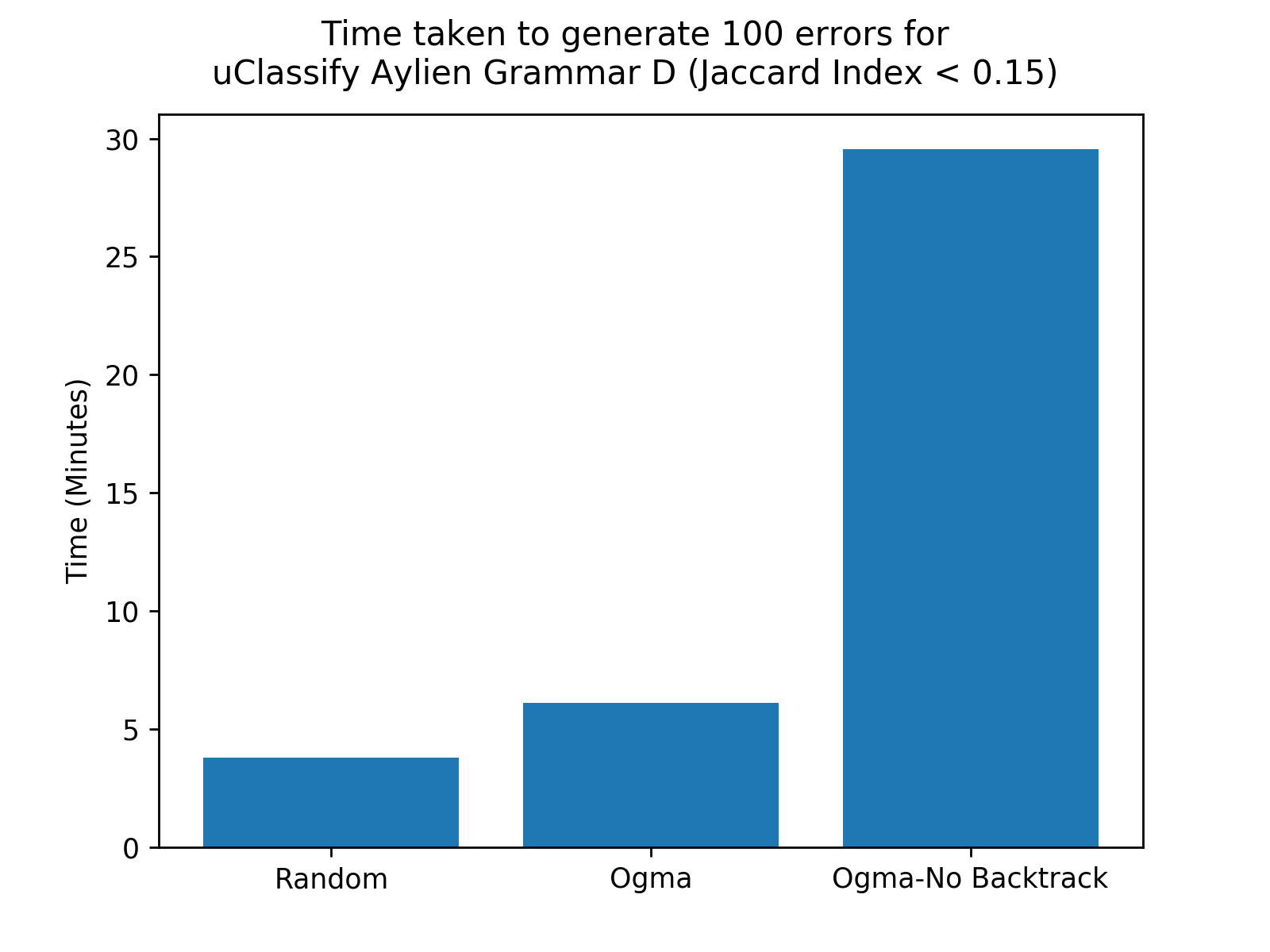} & 
\includegraphics[scale=0.3]{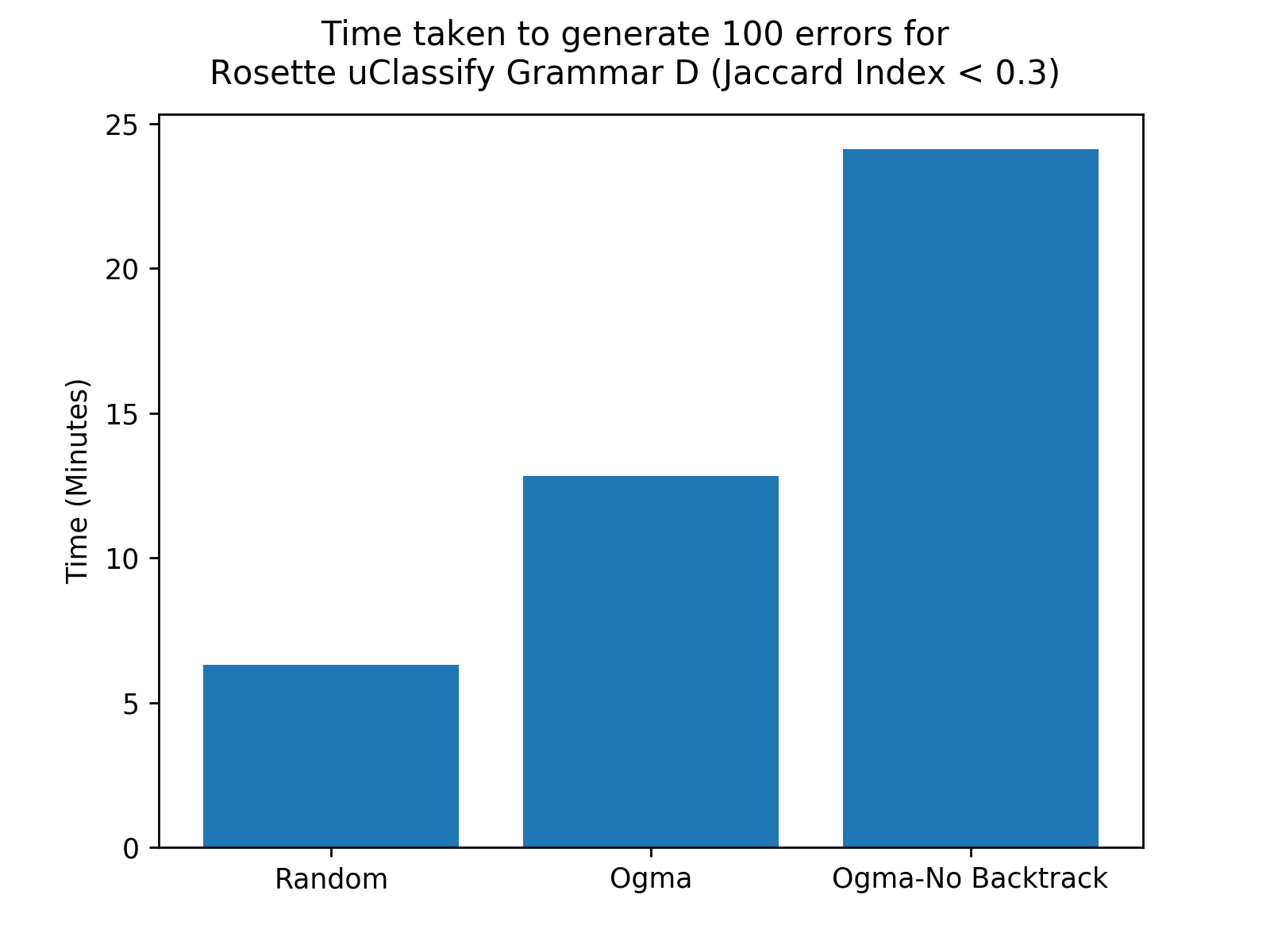}\\

\\
\includegraphics[scale=0.3]{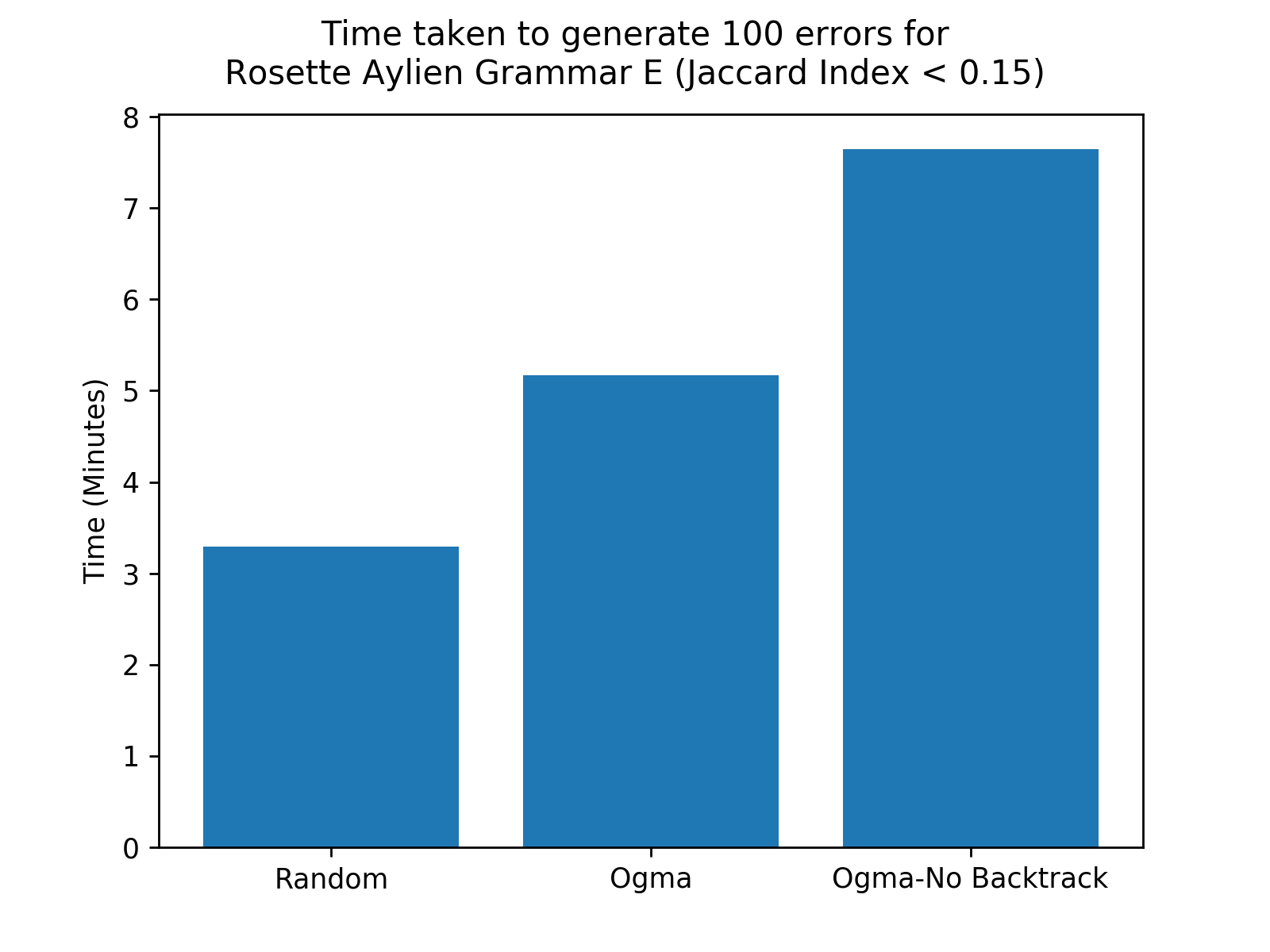} & 
\includegraphics[scale=0.3]{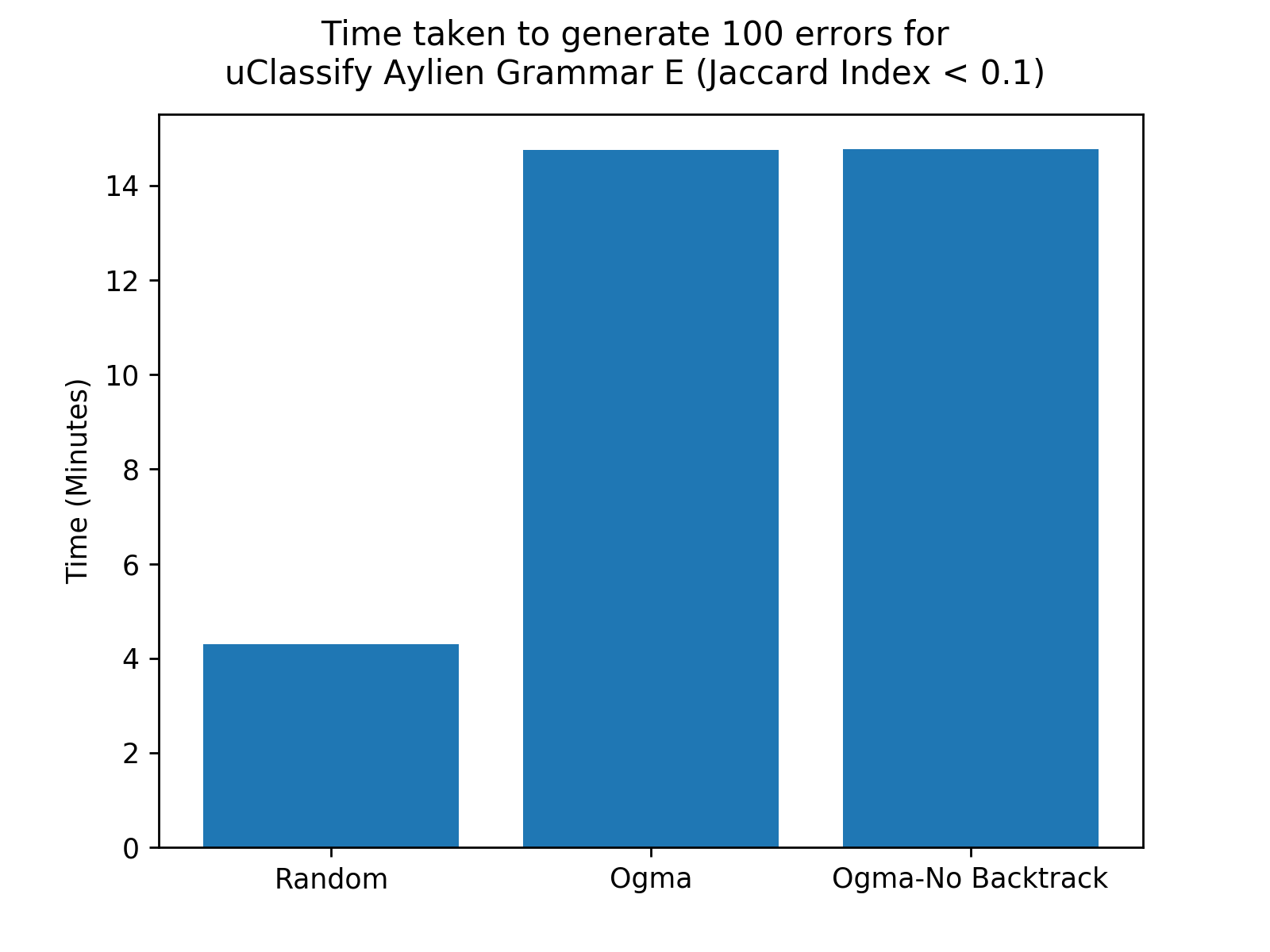} & 
\includegraphics[scale=0.3]{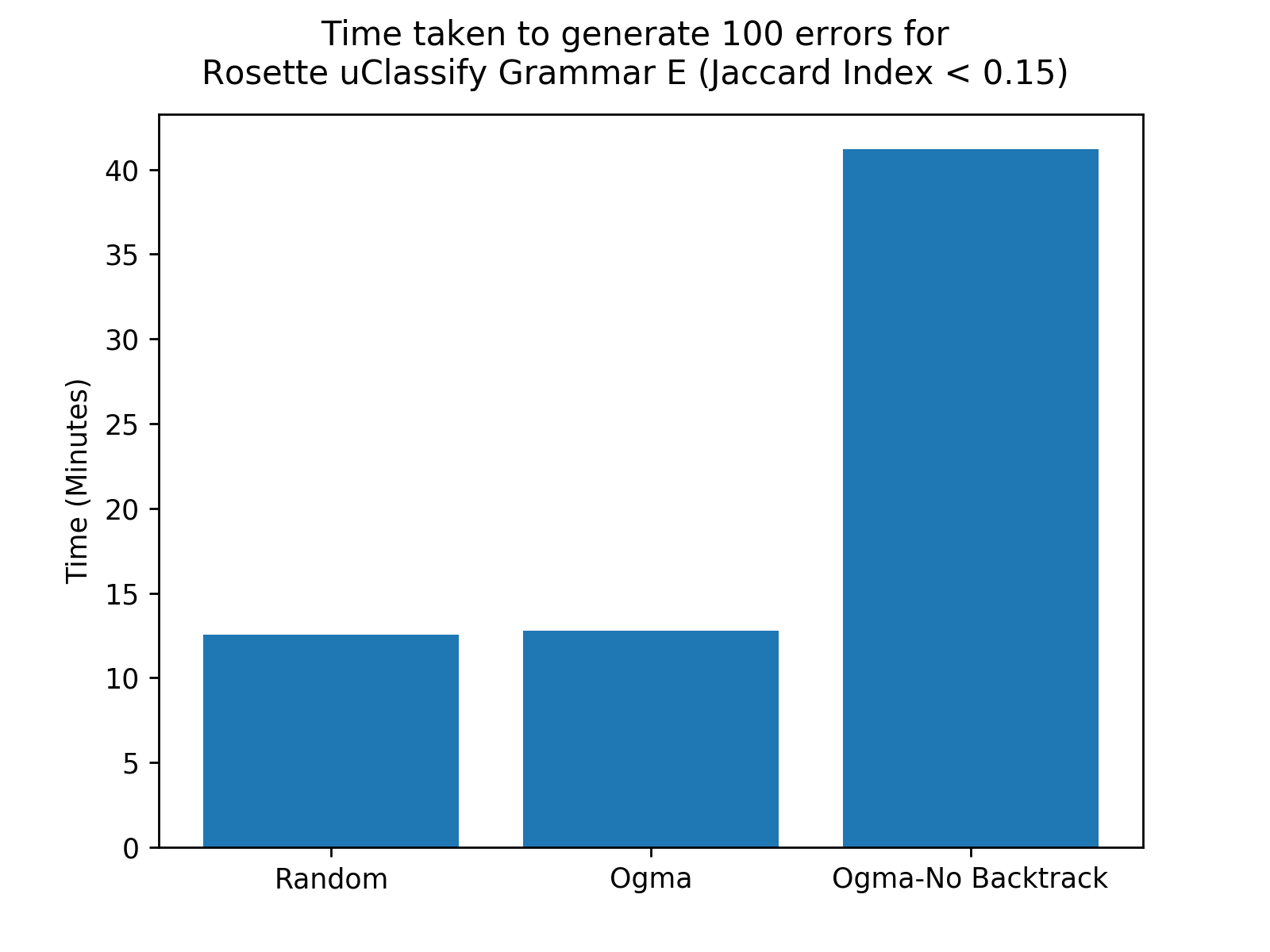}\\

\\
\includegraphics[scale=0.3]{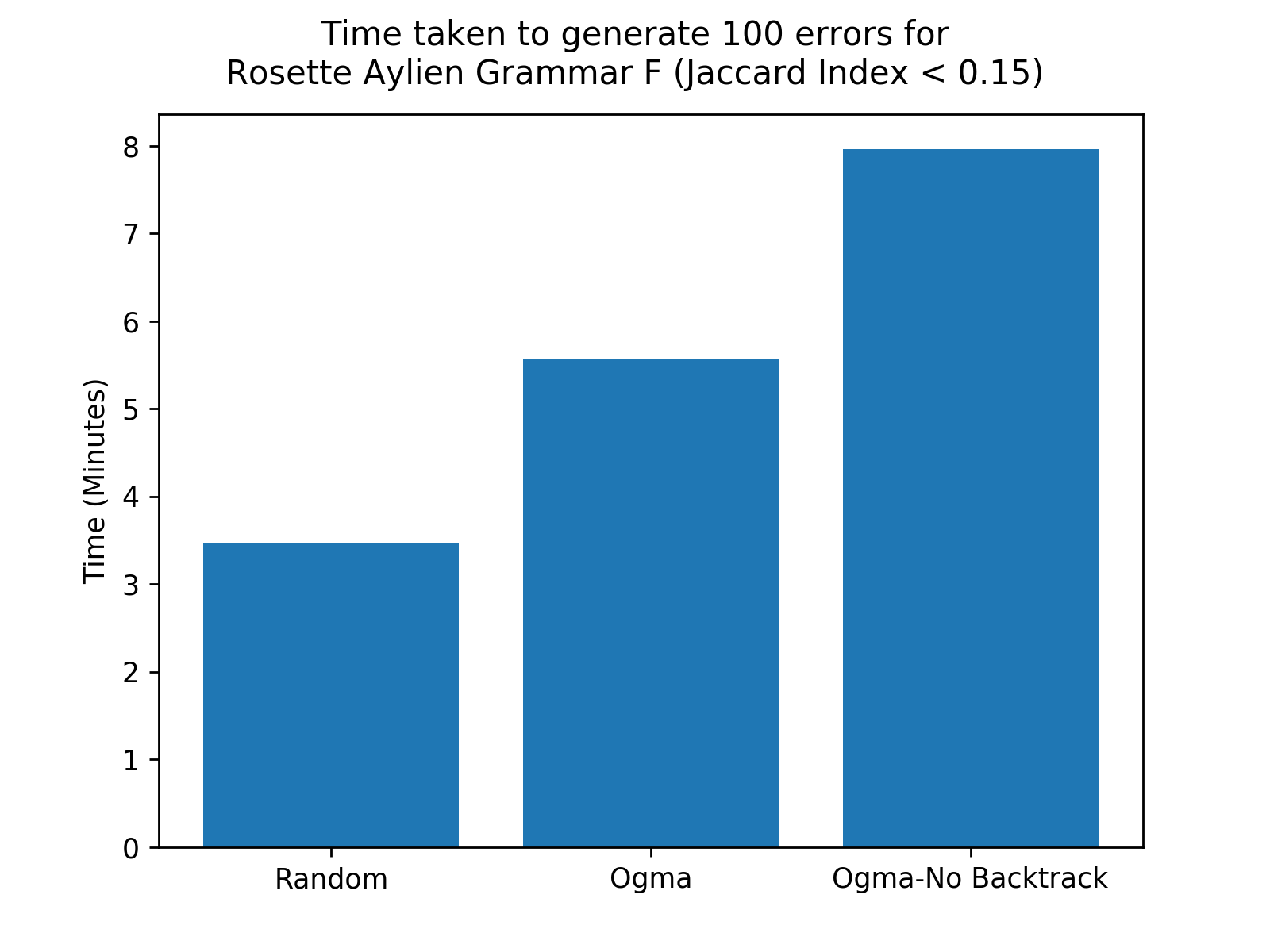} & 
\includegraphics[scale=0.3]{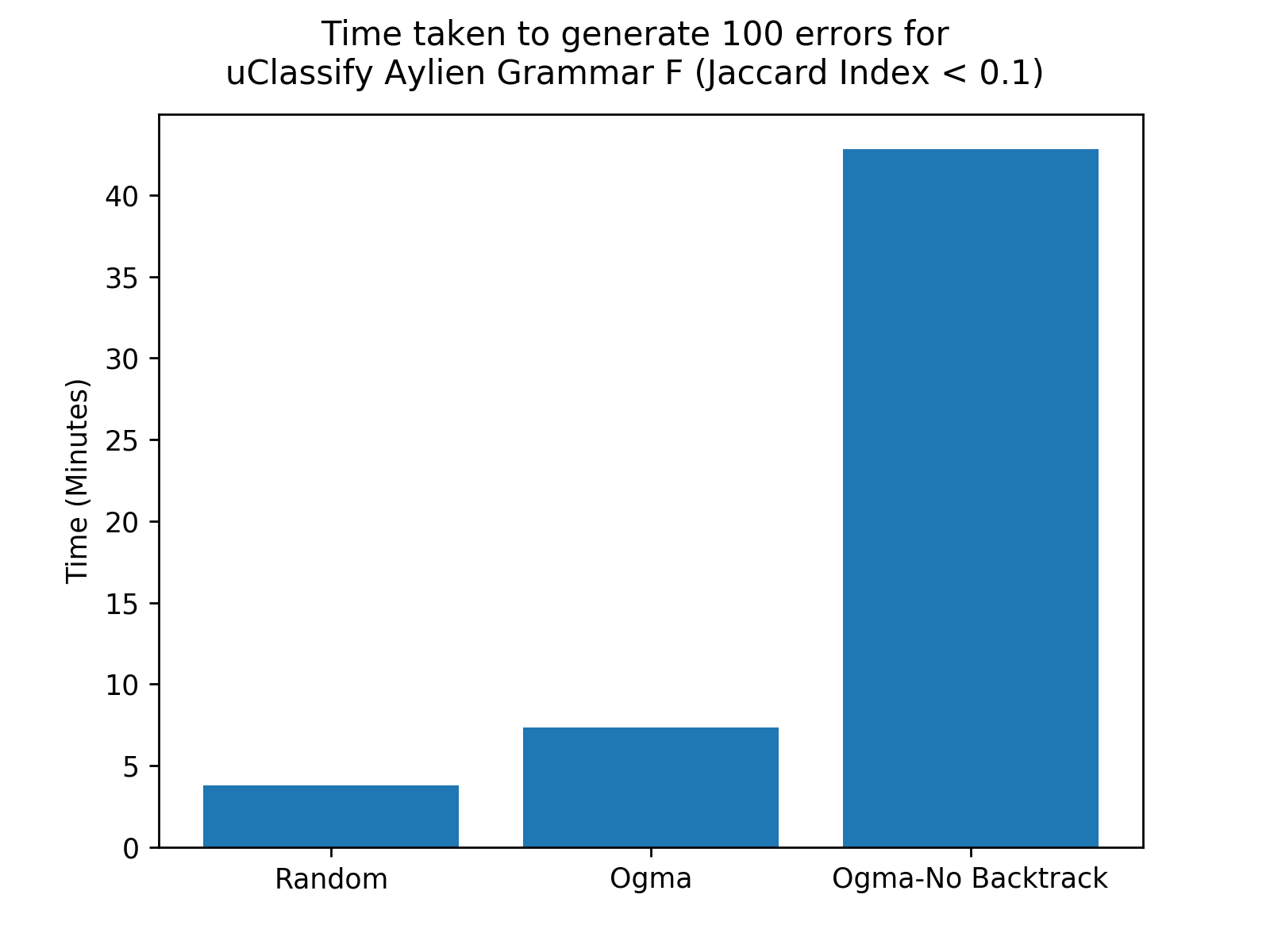} & 
\includegraphics[scale=0.3]{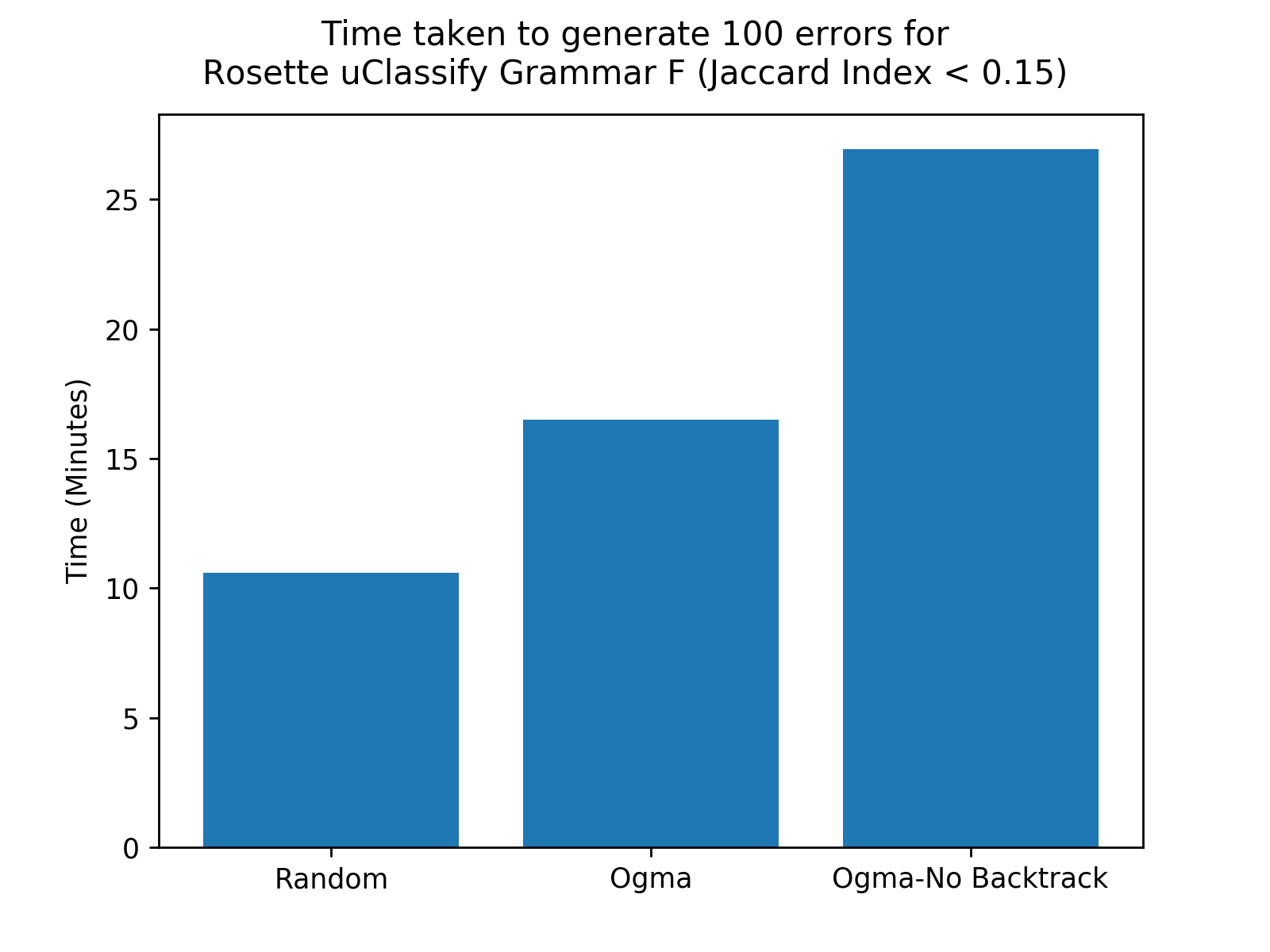}\\

\\
\end{tabular}
\end{center}
\caption{Time taken (in minutes) to complete 2000 iterations}
\label{fig:error-start-time}
\end{figure*}

\begin{figure*}[t]
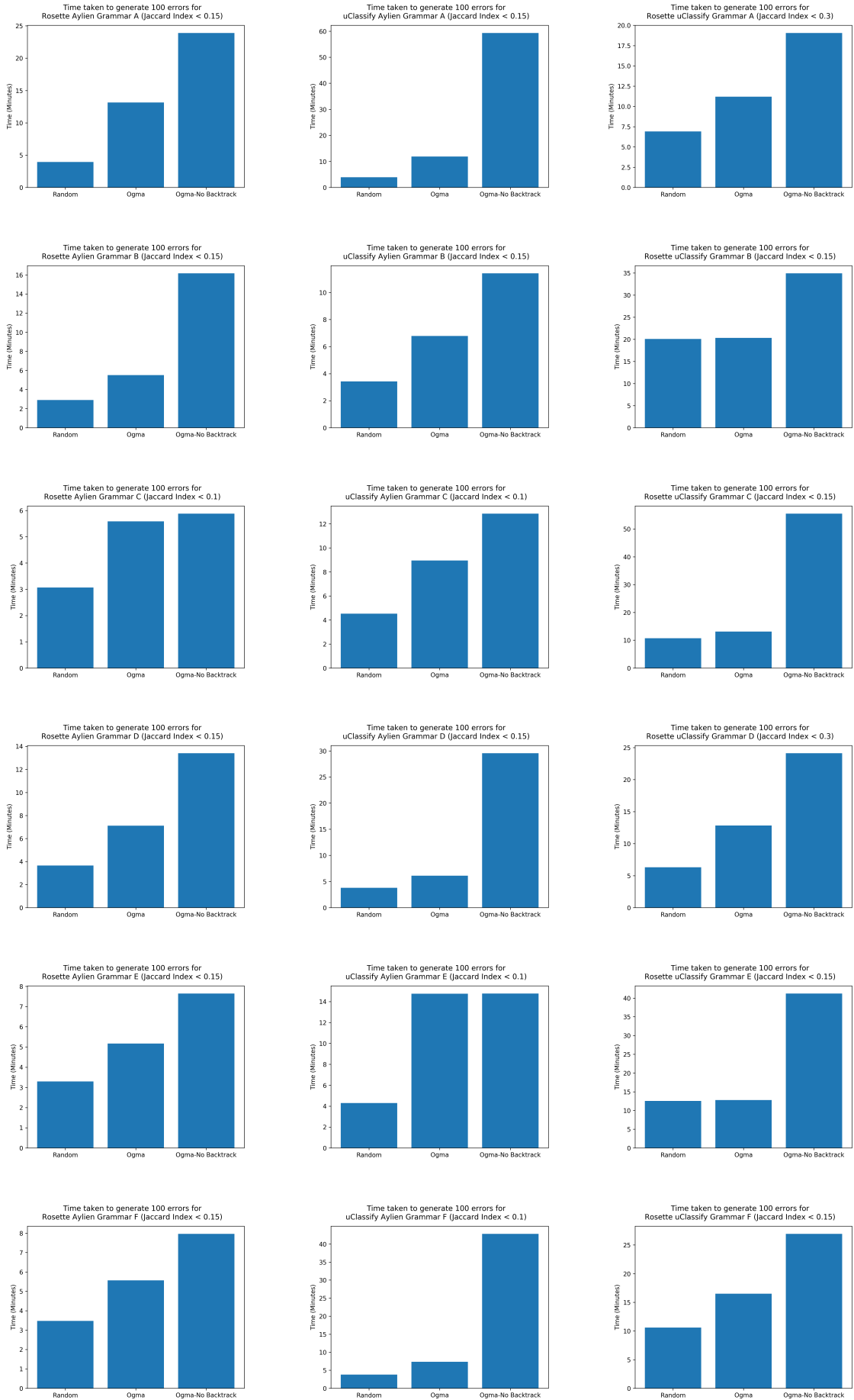

\begin{center}
\begin{tabular}{ccc}
\includegraphics[scale=0.3]{figs/master-non-error-start-time-new-100/Rosette-Aylien-Grammar-A.png} & 
\includegraphics[scale=0.3]{figs/master-non-error-start-time-new-100/uClassify-Aylien-Grammar-A.png} & 
\includegraphics[scale=0.3]{figs/master-non-error-start-time-new-100/Rosette-uClassify-Grammar-A.png}\\

\\
\includegraphics[scale=0.3]{figs/master-non-error-start-time-new-100/Rosette-Aylien-Grammar-B.png} & 
\includegraphics[scale=0.3]{figs/master-non-error-start-time-new-100/uClassify-Aylien-Grammar-B.png} & 
\includegraphics[scale=0.3]{figs/master-non-error-start-time-new-100/Rosette-uClassify-Grammar-B.png}\\

\\
\includegraphics[scale=0.3]{figs/master-non-error-start-time-new-100/Rosette-Aylien-Grammar-C.png} & 
\includegraphics[scale=0.3]{figs/master-non-error-start-time-new-100/uClassify-Aylien-Grammar-C.png} & 
\includegraphics[scale=0.3]{figs/master-non-error-start-time-new-100/Rosette-uClassify-Grammar-C.png}\\

\\
\includegraphics[scale=0.3]{figs/master-non-error-start-time-new-100/Rosette-Aylien-Grammar-D.png} & 
\includegraphics[scale=0.3]{figs/master-non-error-start-time-new-100/uClassify-Aylien-Grammar-D.png} & 
\includegraphics[scale=0.3]{figs/master-non-error-start-time-new-100/Rosette-uClassify-Grammar-D.png}\\

\\
\includegraphics[scale=0.3]{figs/master-non-error-start-time-new-100/Rosette-Aylien-Grammar-E.png} & 
\includegraphics[scale=0.3]{figs/master-non-error-start-time-new-100/uClassify-Aylien-Grammar-E.png} & 
\includegraphics[scale=0.3]{figs/master-non-error-start-time-new-100/Rosette-uClassify-Grammar-E.png}\\

\\
\includegraphics[scale=0.3]{figs/master-non-error-start-time-new-100/Rosette-Aylien-Grammar-F.png} & 
\includegraphics[scale=0.3]{figs/master-non-error-start-time-new-100/uClassify-Aylien-Grammar-F.png} & 
\includegraphics[scale=0.3]{figs/master-non-error-start-time-new-100/Rosette-uClassify-Grammar-F.png}\\

\\
\end{tabular}
\end{center}
\caption{Time taken (in minutes) to reach 100 errors}
\label{fig:error-start-time-100}
\end{figure*}

\begin{figure*}[t]
\begin{center}
\begin{tabular}{ccc}
\includegraphics[scale=0.3]{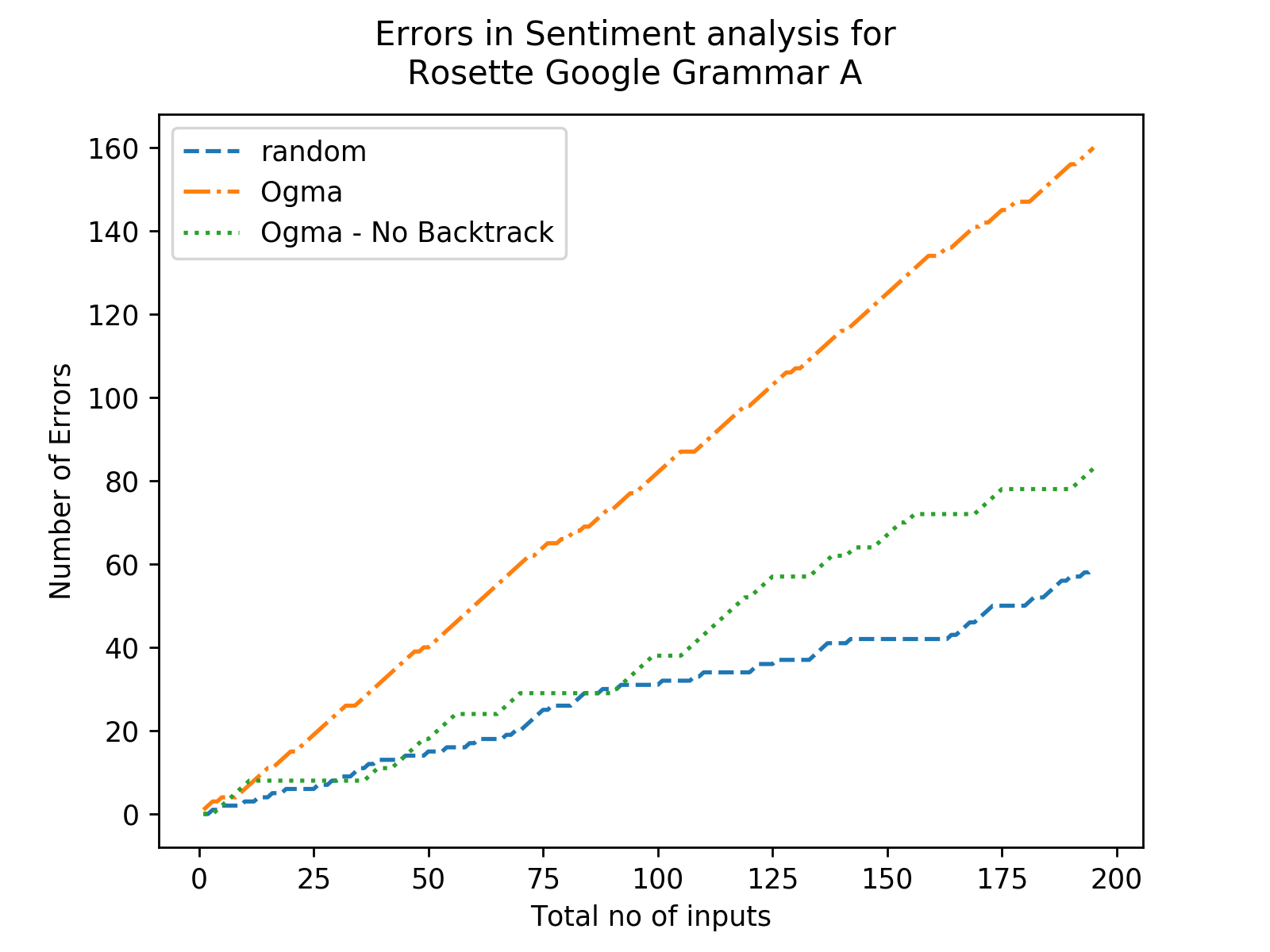} & 
\includegraphics[scale=0.3]{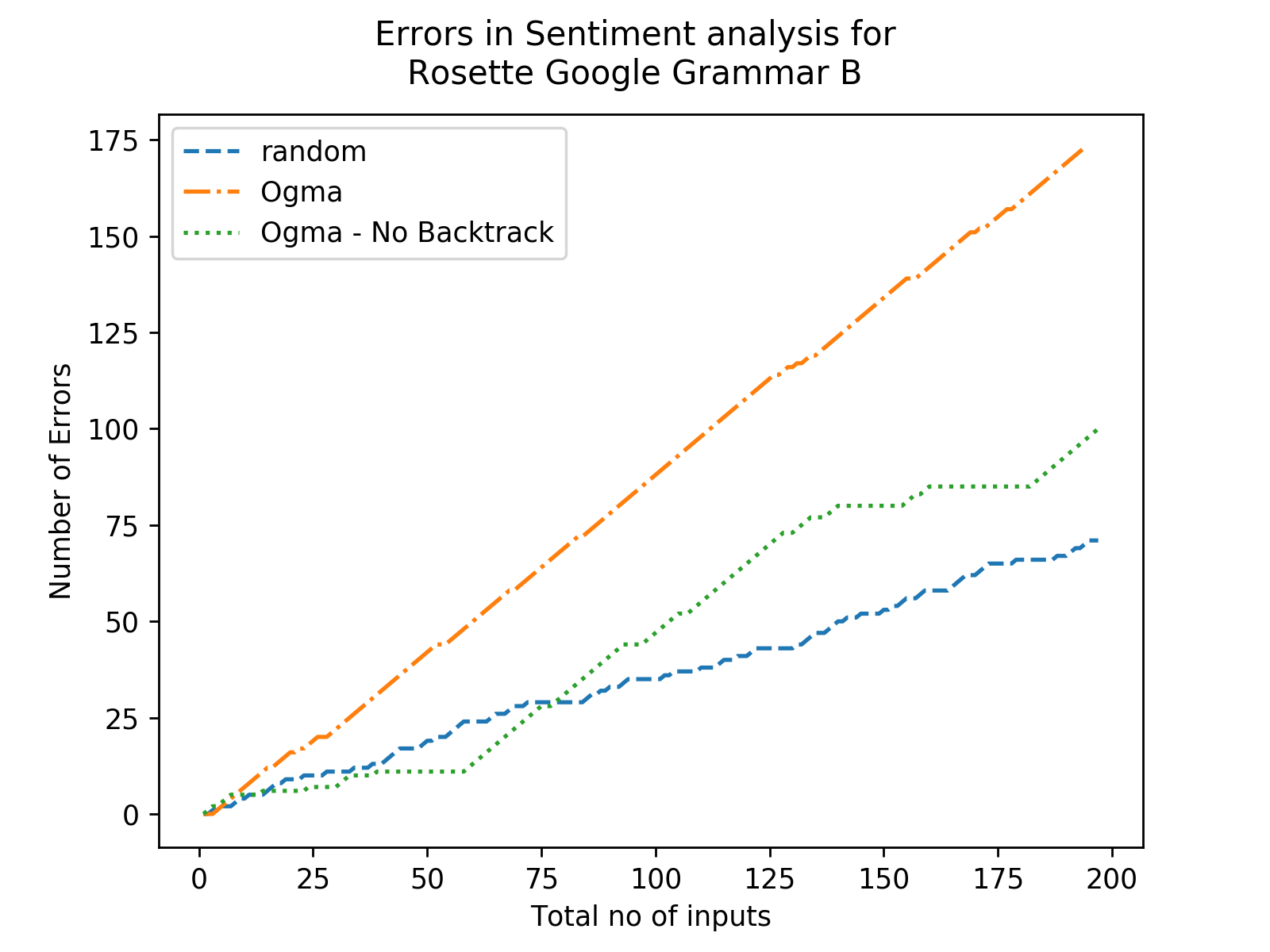} & 
\includegraphics[scale=0.3]{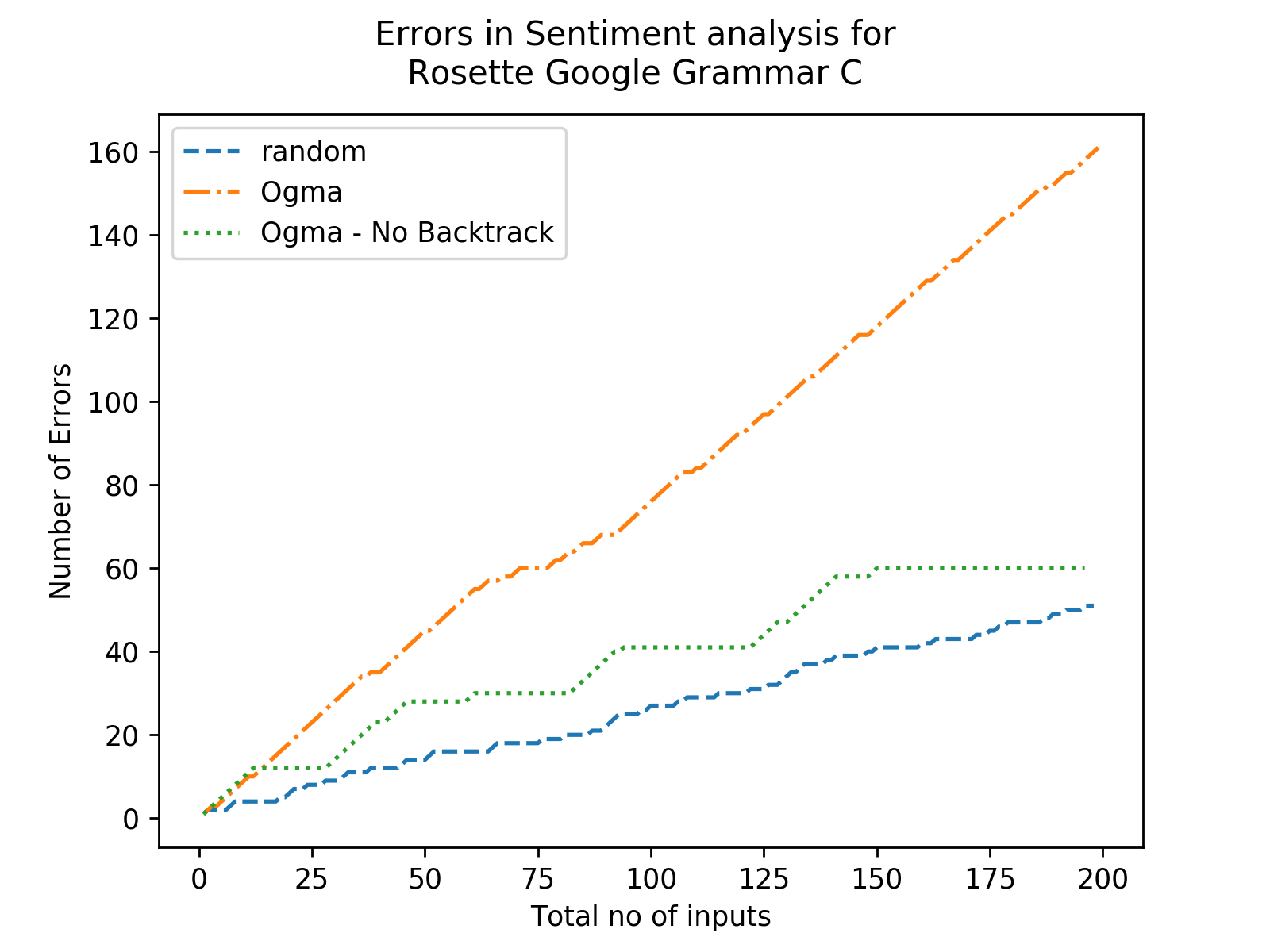}
\\

\includegraphics[scale=0.3]{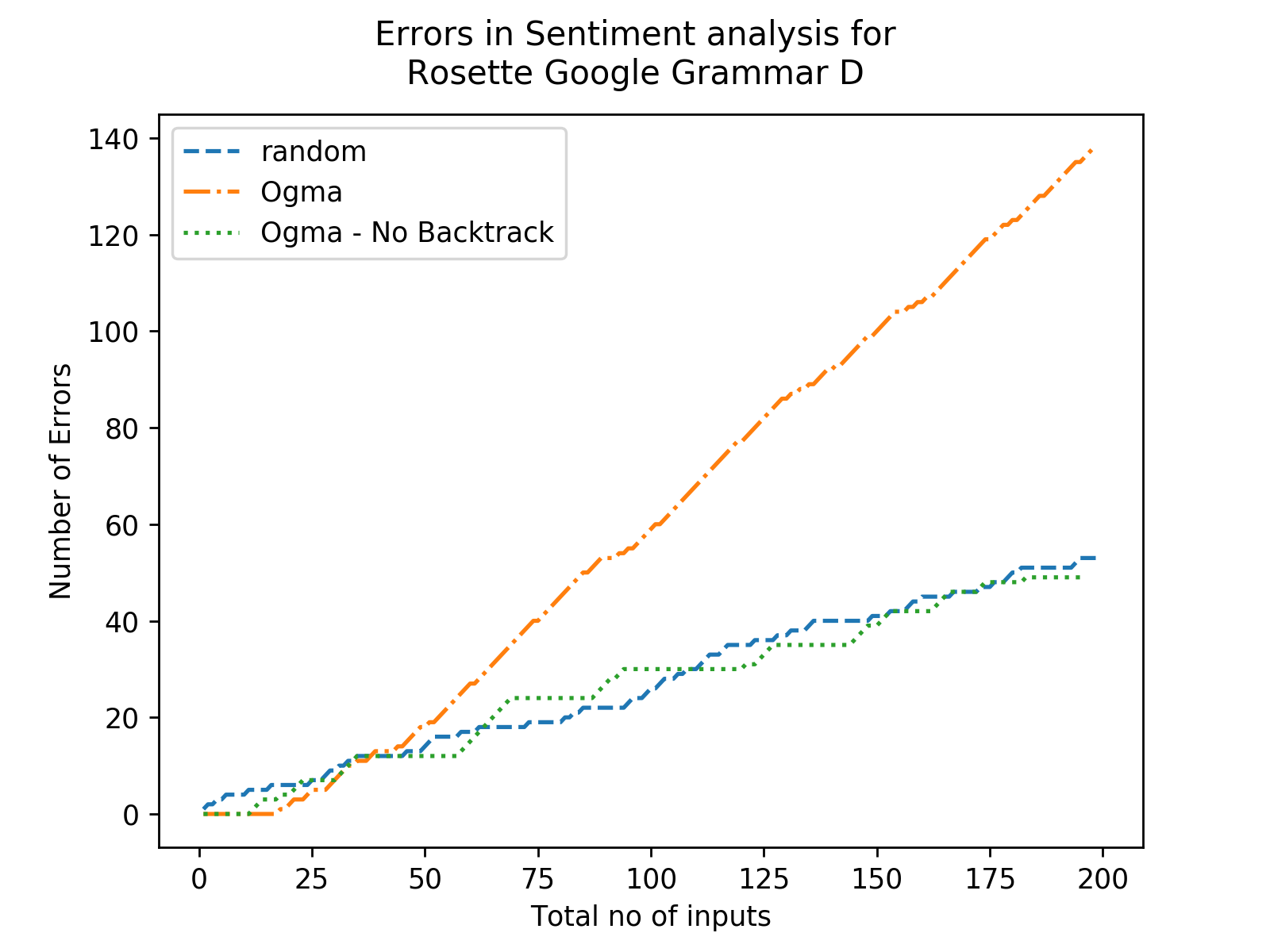} & 
\includegraphics[scale=0.3]{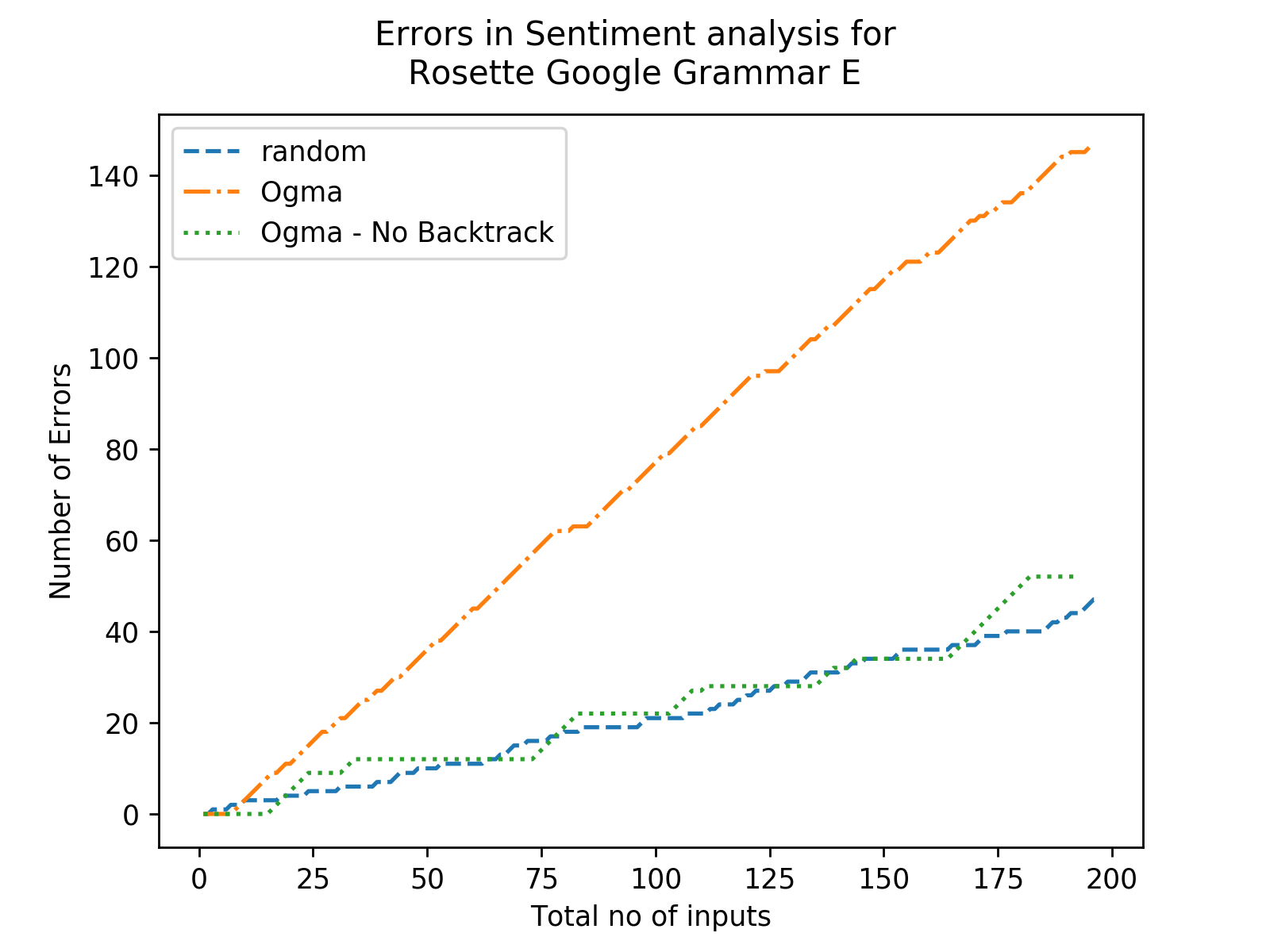} & 
\includegraphics[scale=0.3]{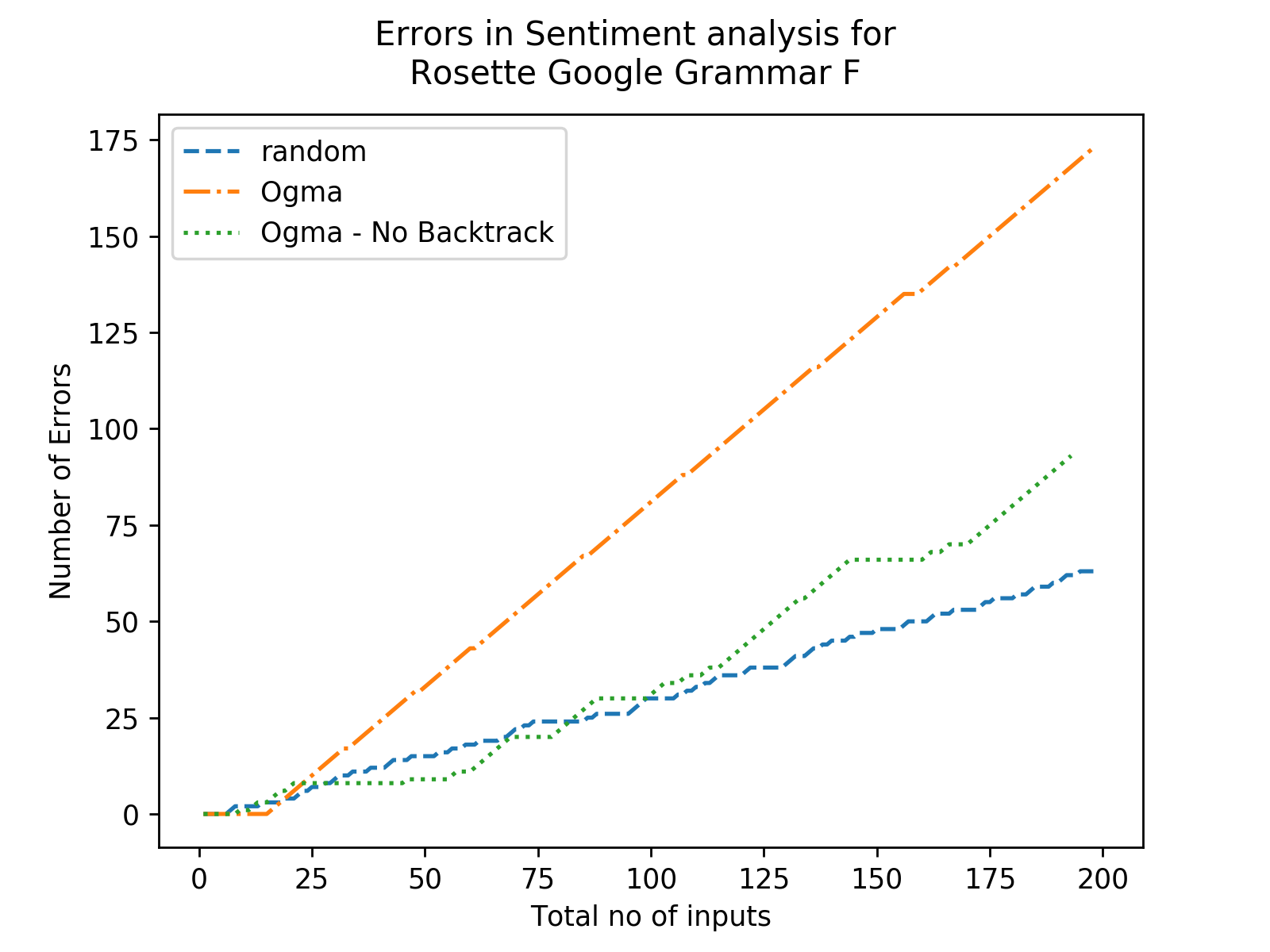}
\\
\end{tabular}
\end{center}
\caption{Errors in Sentiment Analysis for Google and Rosette}
\label{fig:error-sentiment-analysis}
\end{figure*}

\end{document}